%% file: main.tex
\documentclass[a4paper,12pt]{report}

\usepackage[table]{xcolor}

\usepackage{vub} 
\usepackage[english]{babel}

\usepackage{mhchem}

\usepackage{tcolorbox}

\usepackage{graphicx}
\graphicspath{{images/}}

\usepackage{fancyhdr}
\usepackage{caption}
\usepackage{subcaption}
\usepackage{epigraph} 

\usepackage{csquotes}
\usepackage{amssymb}
\usepackage{amsthm}

\usepackage{amsmath}
\DeclareMathOperator*{\argmin}{arg\,min} 

\usepackage{textcomp}
\usepackage{gensymb}
\usepackage{mathtools}
\usepackage{dsfont}
\usepackage{hyperref}
\usepackage{amsmath}
\usepackage{braket}
\usepackage{diagbox}
\usepackage{braket}

\usepackage[sorting=none]{biblatex}
\bibliography{bibliography.bib}

\theoremstyle{definition}

\newtheorem*{definition*}{Definition}

\makeatletter
\newenvironment{chapquote}[2][2em]
 {\setlength{\@tempdima}{#1}
   \def\chapquote@author{#2}
   \parshape 1 \@tempdima \dimexpr\textwidth-2\@tempdima\relax%
   \itshape}
  {\par\normalfont\hfill--\ \chapquote@author\hspace*{\@tempdima}\par\bigskip}
\makeatother

\title{Data-driven design of optical resonators}
\subtitle{Using Artificial Intelligence to gain insight into nanophotonic structures}
\faculty{Faculty of Sciences and Bio-Engineering Sciences \\ Department of Physics} 
\date{June 5, 2020}
\promotors{Prof. Vincent Ginis, Hannah Pinson}
\author{Joeri Lenaerts}

\begin{document}
\maketitle
\tableofcontents
\newpage

\chapter*{Abstract}
\input{chapters/abstract}

\clearpage

\chapter*{Acknowledgements}
\input{chapters/acknowledgements}

\clearpage

\chapter{Introduction: AI for Optics/Physics}
\begin{chapquote}{Plato}
The beginning is the most important part of the work.
\end{chapquote}
\input{chapters/chapter1}

\clearpage

\chapter{Numerical optimization in Nanophotonics}
\begin{chapquote}{Richard Feynman}
There's plenty of room at the bottom.
\end{chapquote}
\input{chapters/chapter2}

\clearpage

\chapter{Deep Learning}
\label{chap 3: deep learning}
\begin{chapquote}{Jean-Paul Sartre}
Life has no meaning a priori… It is up to you to give it a meaning, and value is nothing but the meaning that you choose.
\end{chapquote}
\input{chapters/chapter3}

\clearpage

\chapter{Learning the transmission of a Fabry-Pérot resonator}
\label{chap 4: predicting transmission}
\begin{chapquote}{\textit{Le petit prince}, Antoine de Saint-Exupéry}
All grown-ups were once children... but only few of them remember it. \\
\end{chapquote}
\input{chapters/chapter4}

\clearpage

\chapter{Discovering hidden structure}
\label{chap 5: hidden structure}
\begin{chapquote}{Friedrich Nietzsche}
The real world is much smaller than the imaginary.
\end{chapquote}
\input{chapters/chapter5}

\clearpage

\chapter{Inverse design}
\label{chap 6: inverse design}
\begin{chapquote}{Richard Feynman}
What I cannot create, I do not understand.
\end{chapquote}
\input{chapters/chapter6}

\clearpage

\chapter{Closing remarks}
\input{chapters/closingremarks}

\clearpage

\appendix 

\chapter{Analytical expression Fabry-Pérot}
\label{Appendix: Fabry-Perot}

\input{chapters/appendixA}

\clearpage

\printbibliography

\end{document}

%% file: chapters/abstract.tex
$\\$
The field of Nanophotonics has evolved rapidly over the last years. Optical devices lie at the heart of most of the technology we see around us. When one actually wants to make such an optical device, one encounters the problem that it is very difficult to predict its optical behavior. One needs to compute Maxwell's equations by means of a numerical simulation. If one then asks what the optimal design would be in order to obtain a certain optical behavior, the only way to go further would be to try out all of the possible designs and compute the electromagnetic spectrum they produce by solving Maxwell's equations. When there are many design parameters, this brute force approach quickly becomes too computationally expensive to be useful. We therefore need other methods to create optimal optical devices. 

$\\$
An alternative to the brute force approach is inverse design. In this paradigm, one starts from the desired optical response of a material and then determines the design parameters that are needed to obtain this optical response. There are many algorithms known in the literature that implement this inverse design. Some of the best performing, recent approaches are based on Deep Learning, a subfield of Artificial Intelligence. The central idea is to train a neural network to predict the optical response for given design parameters. Neural networks can make these predictions much faster than numerical simulations that solve Maxwell's equations. Since neural networks are completely differentiable, we can compute gradients of the response with respect to the design parameters. We can use these gradients to update the design parameters and get an optical response closer to the one we want. This allows us to obtain an optimal design much faster compared to the brute force approach. 

$\\$
In my thesis, I use Deep Learning for the inverse design of the Fabry-Pérot resonator. This system can be described fully analytically. It is therefore ideal to study, since we already know what the optimal solution should be. This allows us to analyze the performance of the methods used for inverse design. 

%% file: chapters/acknowledgements.tex
First of all, I would like to thank Prof. Vincent Ginis for giving me the opportunity to work at the interface of Physics and Artificial Intelligence. He provided me with a lot of interesting research ideas to pursue. This made the thesis a research project I really enjoyed. I would also like to thank Hannah Pinson for helpful discussions. Her understanding of Artificial Intelligence gave me a lot of insight into what I was doing.

$\\$
I also want to give a shout-out to my fellow students in Physics. To see them working on their thesis and facing the same challenges as me motivated me a lot to keep going and to keep doing good research. What I liked most were the pleasant thesis discussions over lunch. 

$\\$
At last I want to thank my mother. She has always provided me with everything I needed so I only needed to worry about doing Physics. Mom, thank you for everything you do. It is very much appreciated.

%% file: chapters/chapter1.tex
\noindent The ancient Greek philosopher Plato believed in the world of Forms. The Forms are the ideal versions of things like justice, education or government. According to Plato, the Forms can never be achieved, but it is useful to think about what the ideal Form of something would look like. This allows us to gain insight in how to best approximate the Forms. We can apply these ideas to Optics.

$\\$
What would an ideal optical device look like? We could use it to control and bend light in any way we want. When the light is refracted, we could specify exactly how much of the light is reflected and how much is transmitted. We could also control light based on its color, or more technical on the wavelength of the light. We could make a device that splits light into its different components and that bends them in any direction we want. This would be the ideal optical device.

$\\$
Just like the Forms of Plato, the ideal optical device can never be achieved. It does inspire us however to create devices that come quite close. One method to obtain these optical devices with stunning capabilities is inverse design. This paradigm takes the ideal behaviour of the optical device as a starting point, just as Plato would have done. Then it tries to construct an optical device that best approximates this ideal behaviour. This thesis thus continues the two thousand year old idea of Plato to take the ideal Form as a starting point for progress.

\section{Context}

Inverse design offers a set of computational methods to create optical devices. To configure these devices, we need to specify material parameters and parameters describing the geometry. We call these parameters collectively the design parameters. The methods of inverse design compute the optimal parameters of the optical device in order to obtain a desired optical behaviour. In recent years, many computational methods for inverse design have been improved by a subfield of AI called Machine Learning. This is what we further investigate in this thesis.

$\\$
The main goal of Machine Learning is to create models that can learn from data. This learning takes the form of 3 standard problems: supervised learning, unsupervised learning and reinforcement learning. These three paradigms each have their own goals. Supervised and unsupervised learning are the ones that have been most useful in Physics to date. These are also the ones that I worked with in my thesis. We now further explain those paradigms based on \cite{Carleo2019}.

$\\$
In the case of \textbf{supervised learning}, the data is a set of inputs $\textbf{x}$ together with its labels $\textbf{y}$. Both $\textbf{x}$ and $\textbf{y}$ are given as vectors. For image recognition for example, the input $\bf{x}$ contains the intensity values of the pixels of the image, while the label $\bf{y}$ describes what is shown in the image. The goal of supervised learning is to find a function $f: \bf{x} \mapsto \bf{y}$ that is able to predict the label of the image and is also able to generalize to new, unseen images. This function is approximated by a model with parameters $\bf{w}$. Training the model amounts to finding the optimal parameters $\bf{w}$ for the model $f_{\bf{w}}$. In order to train these parameters, we need a $\textbf{loss function}$. This function $L$ describes how well the predictions of the model agree with the true labels. We write this as

\begin{equation}
    \nonumber
    L ( f_{\bf{w}}(\bf{x}), \bf{y}).
\end{equation}
The optimal parameters $\bf{w}$ for the model minimize this loss function. The form of the loss function depends on the task we are dealing with. There are two classes of problems we want to solve, classification and regression. In classification, the output labels $\bf{y}$ are discrete, like in the example of image classification. The loss tells us whether or not the labels match. In regression, the output $\bf{y}$ is a continuous variable. An example is the prediction of the price of a house based on its properties. A common loss function for regression is the $\textbf{mean squared error (MSE)}$ given by

\begin{equation}
    \nonumber
    MSE = \vert f_{\bf{w}}(\bf{x}) - \bf{y} \vert ^2.
\end{equation}

$\\$
The goal of $\textbf{unsupervised learning}$ is different. Here, we do not have labels $\bf{y}$, but only data $\bf{x}$. We want to find the underlying structure of the data. Two typical examples are clustering and generative modeling. When we are clustering data, we want to know something about the similarity of different data points. In the case of images, this means putting images with similar content in the same cluster. On the other hand, generative modelling tries to model the underlying probability distribution $p(\bf{x})$ from which the data is sampled. To this end, latent variables $\bf{z}$ are introduced. These latent variables are the generating factors that determine the structure of the data. For images, they would characterize what is shown in the image. For simple shapes, latent variables could be the size, the rotation and the position of the shape. This information describes the image with much less data than specifying all pixels. When we try compress high dimensional data $\bf{x}$ to a low dimensional representation, we speak of $\textbf{dimensionality reduction}$. This type of generative modelling is actually very similar to the way we do Physics. We also compress the world around us into a set of observables governed by equations. This set of observables and equations provides us with everything we need to understand the world around us. 

$\\$
$\textbf{Reinforcement learning}$ trains agents that interact with their environment. These agents can for example learn to play classic Atari games. Learning in this paradigm consists of finding the action that will lead to the greatest reward. This reward can be in the short term or more commonly in the long term, like winning the game. A major breakthrough was the training of an agent that beat the world champion in the board game of Go. A great challenge was that there is only a single reward at the end of the game, winning or losing. Victory depends on all of the dozens of moves that were taken during the game. The challenge was overcome using the same algorithm as for Atari games, which is called $\textbf{Deep Q Learning}$. Another advance that I consider one of the most impressive in reinforcement learning is the creation of OpenAI Five \cite{openai2019dota}. This AI system learned to play the online multiplayer game of Dota, which is a stunningly complex game. A player needs to know which weapons to buy, which attacks to upgrade on level up and the strengths and weaknesses of the hero she is playing with. Next to that, there is a nearly infinite amount of movement each player can make. Despite these difficulties, an agent was trained to beat the world's leading team of Dota. This was a great leap forward in the field of AI. In spite of these great achievements, the use of reinforcement learning for Physics has been limited so far. We will therefore not discuss this paradigm any further in this thesis. 

\section{Motivation}

Machine Learning has already found a lot of interesting applications in Physics. A first example is the field of Statistical and Condensed Matter Physics. An important task in these fields is the identification of phases and phase transitions. In \cite{Carrasquilla2017}, a neural network was trained to discover these phases of matter. The authors show the validity of their approach on the Ising model. 

$\\$
Another application of AI to Physics arises in Astronomy. There is an enormous amount of astronomical observations available. To form conclusions based on these observations is a long and quite tedious task to do by manual inspection of millions of images. Therefore, Machine Learning can aid in the processing of images taken by telescopes. Machine Learning was also used in the IceCube collaboration to detect signals of neutrino's. A neural network was trained to detect these neutrino's in the ice of Antarctica \cite{icecube}. An overview of Machine Learning in Astronomy can be found in \cite{baron2019machine}.

$\\$
A last application we would like to mention is in particle collider experiments. In large collider experiments like LHC, there are of the order O($10^8$) sensors recording data. This data is recorded for millions of events every second. In order to select the interesting data, high level features of the low level data are constructed. Afterwards, statistical analysis is performed on these high level features of the collision. Much can be gained however by looking at the raw data. Deep Learning can provide a method to also take these low level features into account. The Deep Learning approach was already successfully applied to the classification of events as signal or background processes in a supersymmetric particle search. A review of Deep Learning for LHC Physics can be found in \cite{Guest2018}.

$\\$
These examples show that Machine Learning is already applied in a wide variety of fields in Physics. It offers us a set of new tools do to research. This complements the analytical derivations and numerical simulations on which Physical theories are built today. In this thesis, I investigate how these tools can be applied to Optics.

\section{Thesis outline}

The use of Machine Learning in Physics is now well motivated. In this thesis, we apply Machine Learning to perform inverse design of optical devices in Nanophotonics. The optical device we choose to investigate is the Fabry-Pérot resonator. This system is analytically well understood, which provides us with a way of assessing the performance of multiple inverse design methods. The thesis is structured as follows.

\paragraph{Chapter 2}In chapter 2, we give an introduction to current computational methods for inverse design in Nanophotonics. The research in using computational methods to design optical devices started in the late 90s \cite{spuhler1998}, \cite{cox1999}. In the following years, the methods were refined and saw a steady increase in performance. In the last few years, Machine Learning has been introduced for inverse design, with great success. 

\paragraph{Chapter 3}This chapter gives an introduction to a subfield of Machine Learning called Deep Learning. The work discussed in this thesis is all situated in this subfield of Deep Learning. It makes use of neural networks that are inspired by the working of the brain. Deep Learning lies at the basis of the great AI progress in image processing, speech recognition and even self-driving cars that we have seen in the last years.

\paragraph{Chapter 4}After the literature study in chapter 2 and 3, we present our results. In this first part, we train a neural network in a supervised setting to predict the transmission of the Fabry-Pérot resonator. 

\paragraph{Chapter 5} We continue by analyzing the Fabry-Pérot resonator in an unsupervised setting. From the analytical expression of the transmission spectrum, we know that it is fully determined by two physically interesting parameters. We investigate whether these parameters can be retrieved from the transmission by unsupervised learning.

\paragraph{Chapter 6} Finally, we build upon the work of the two previous chapters to perform inverse design.

%% file: chapters/chapter2.tex
\noindent The field of Nanophotonics has allowed us to create devices that can manipulate light in incredible ways. Waveguides make it possible to bend light in any way we want, beam splitters make it possible to select light based on its wavelength, bandgap engineering has made solar power possible and in IT, optical technology has enabled the communication between computers that we like to call the internet. The list of useful technologies is long and is only expected to grow longer. 

$\\$
In order to design most of the technologies mentioned above, the standard approach is based on intuition and brute-force calculations. To design an optical device, you start from the standard templates available in the optical literature. These are quite simple designs with a high degree of symmetry or other useful properties. To optimize this structure, you try out some of these different templates, make an educated guess for their parameters and evaluate their optical response. After testing a few designs, you have created a new optical structure. This strategy has proven to be very successful.

$\\$
Nonetheless, there are two limitations to this approach. The first limitation is that it takes a lot of time and skill to design these structures. You need a broad knowledge of all the available templates and some domain knowledge to know which design is the most appropriate for a given task. Secondly, there is no way to check if the final structure is really optimal. It might be that there are other designs with a much better performance. 

\begin{figure}[h]
    \centering
    \includegraphics[width=0.4\textwidth]{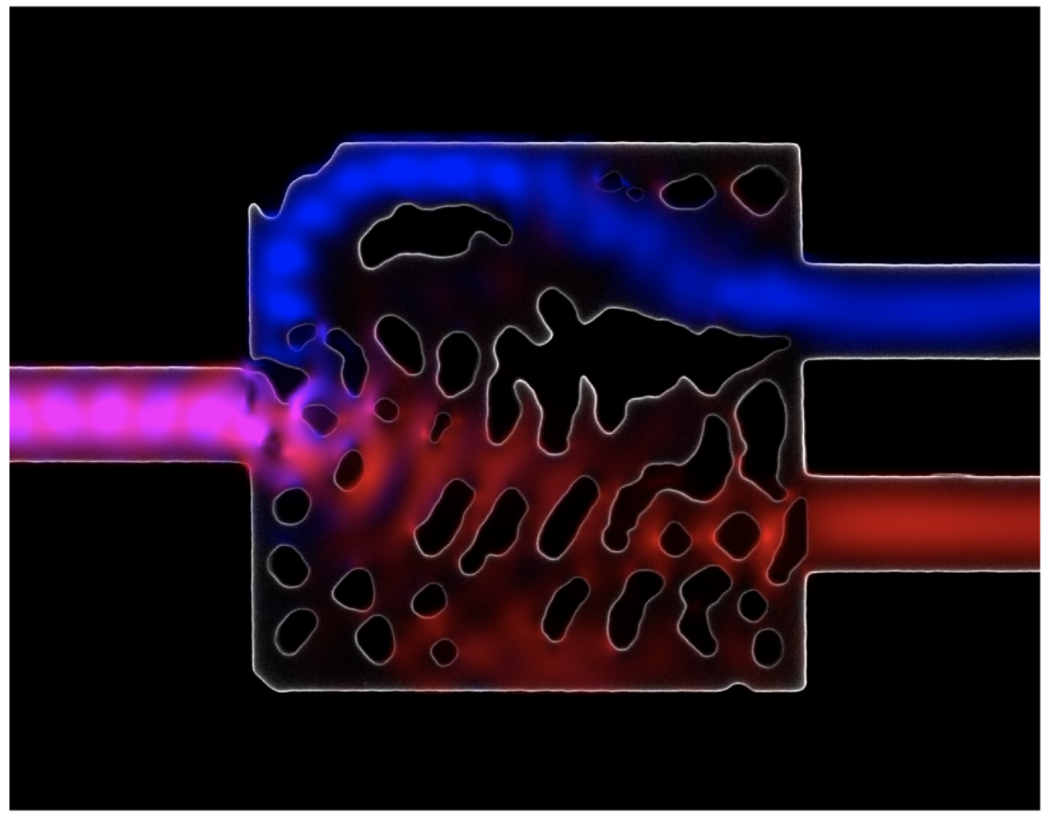}
    \caption{Optical structure separating light of wavelength 1300 nm and 1550 nm. Image taken from \cite{Piggott2015}.}
    \label{fig: nontrivial design}
\end{figure}

$\\$
In order to go beyond these limitations, researchers turned to inverse design. This paradigm starts from a desired optical response and then tries to design a structure that closely matches this desired output. This idea can overcome the two limitations of intuition-based design. First, it allows us to design much more complex structures. For structures with hundreds or thousands of tunable parameters, it is nearly impossible to get an intuition for the way in which each of these parameters influence the outcome. We are however capable of handling a lot of parameters computationally. An added benefit is that very non-trivial designs can be created. The design in figure \ref{fig: nontrivial design} for example, is a structure that humans would never intuitively come up with. 

$\\$
Secondly, in inverse design structures are usually optimized over all possible values for the design parameters. This means that we have information on all of the possible designs. If we can successfully perform inverse design on a structure, it will be the most optimal design we can make within the specified set of parameters. The ability of inverse design to go beyond the limitations of intuition-based design is why inverse design has gained a lot of attention in recent years. 

\noindent In the last few years, Machine Learning entered inverse design. Machine Learning approaches are very promising since they can reduce computation time and scale up to a lot of parameters. Moreover, in some cases they lead to better performing devices compared to non-Machine Learning inverse design.  

$\\$
In this chapter, we first give an overview of the development of inverse design for Nanophotonics. This field has seen steady progress over the last 20 years, with many interesting results along the way. After this historical overview, we continue to explain how Machine Learning and more specifically Deep Learning has become the new state of the art in the field. 

\section{Historical overview}

The general idea of inverse design has been around for hundreds of years. One of the first inverse design problems is the brachistochrone problem posed and solved by Johann Bernoulli at the end of the 17th century. The problem amounts to designing the curve that minimizes the time it takes for a particle to go from point A to point B under the influence of gravity. The problem can be solved by calculus of variations, minimizing the time as a functional of the curve. Another interesting inverse design problem is the principle of least action of Maupertuis. Here, the equations of motions of a mechanical system are found by optimizing the action of this system. These examples show that historically, the idea of inverse design has proven to be very useful in Physics. 

$\\$
Two early applications of inverse design in Nanophotonics can be found in the late 90s in the work of Spühler et al. \cite{spuhler1998} and Cox and Dobson \cite{cox1999}. Their approach was quite different from one another. Spühler et al. used an evolutionary algorithm to design a photonic device that couples a telecom fibre to a wave guide. Such evolutionary algorithms are an approach to AI based on the idea of natural selection. A generation of solutions to the optimization problem is created in the first step. Then in every iteration, the best solutions in the generation are retained. These solutions then undergo mutation and reproduction, causing random variation in the population. This variation can lead to better solutions. After a few iterations, the solutions in a generation converge to an optimal solution. 

$\\$
The other paper of Cox and Dobson used a gradient-based algorithm. These algorithms perform gradient descent, which always converges to a local optimum. The optical structure it was applied to is a 2D periodic structure of two materials. By altering the distribution of the two materials in a unit cell, the bandgap of the material was enlarged. 

$\\$
After the initial success of inverse design, researchers wanted to perform inverse design on problems with increased complexity and a larger number of parameters. This lead to a need for a better theoretical formulation of the problem. There are two formulations that are still very popular today, the \textbf{level set method} and \textbf{density topology optimization}. Both methods provide a way to describe the material parameters of an optical structure such as the permittivity $\epsilon$. The material parameter is specified on a discrete design space. In 1 dimension, this design space contains line elements, in 2 dimensions it is made up of pixels and in 3 dimensions it is made up of voxels. For every element of the design space, we can choose what material it is made of. The goal of inverse design is to find the optimal partitioning leading to the desired optical response. 

$\\$
Such a partitioning can be described by level sets. Consider that we have the choice between material A and material B and that the design space consists of elements $\textbf{x} \in D$. We can then describe the partitioning with a scalar function $\Phi(\bf{x})$ on the design space. The level sets of this function determine whether an element should be made of material A or B, we have

\begin{equation}
    \noindent
    \begin{aligned}
    D_A = \{\mathbf{x} \ \vert \ \Phi (\mathbf{x}) < 0\}, \\
    D_B = \{\mathbf{x} \ \vert \ \Phi(\mathbf{x}) > 0\}.
    \end{aligned}
\end{equation}

\noindent The function $\Phi(\bf{x})$ can be optimized by equations of motions. The level-set method thus provides a way of describing a binary-valued geometry in terms of a single function of the grid $\Phi(\bf{x})$ that we can optimize.

$\\$
Another way to formalize the inverse design problem is by density topology optimization. Let us again consider a binary-valued geometry and assume that we are interested in the permittivity $\epsilon$. Material A has a permittivity $\epsilon_A$ and material B has a permittivity $\epsilon_B$. For each cell in our discrete geometry, we can write the permittivity as a linear combination of these permittivities, given by

\begin{equation}
    \epsilon(\mathbf{x}) = \epsilon_A + \lambda(\mathbf{x}) \epsilon_B,
\end{equation}
where the parameter $\lambda(\mathbf{x}) \in [0, 1]$. In the end, the actual design is only able to take the discrete variables $\epsilon_A$ and $\epsilon_B$. We can however only compute gradients if the design parameters are continuous. The trick to make $\epsilon(\mathbf{x})$ continuous is what allows gradient-based optimization algorithms to work. 

$\\$
A very popular method in density topology optimization is the \textbf{adjoint method}. This method provides a way to compute gradients of a loss function with respect to the design parameters. The design parameters lead to an electric field and both the design and the electric field influence the final performance. The adjoint method allows us to compute only the dependence on the design parameters. This method was recently used to design tunable metasurfaces \cite{chung2019tunable}. The behaviour of the metasurface can be tuned by turning a voltage running through the structure on or off. The metasurface deflects light in a different direction based on the voltage in the structure.

\begin{figure}[h!]
     \centering
     \begin{subfigure}[b]{0.45\textwidth}
         \centering
         \includegraphics[width=\textwidth]{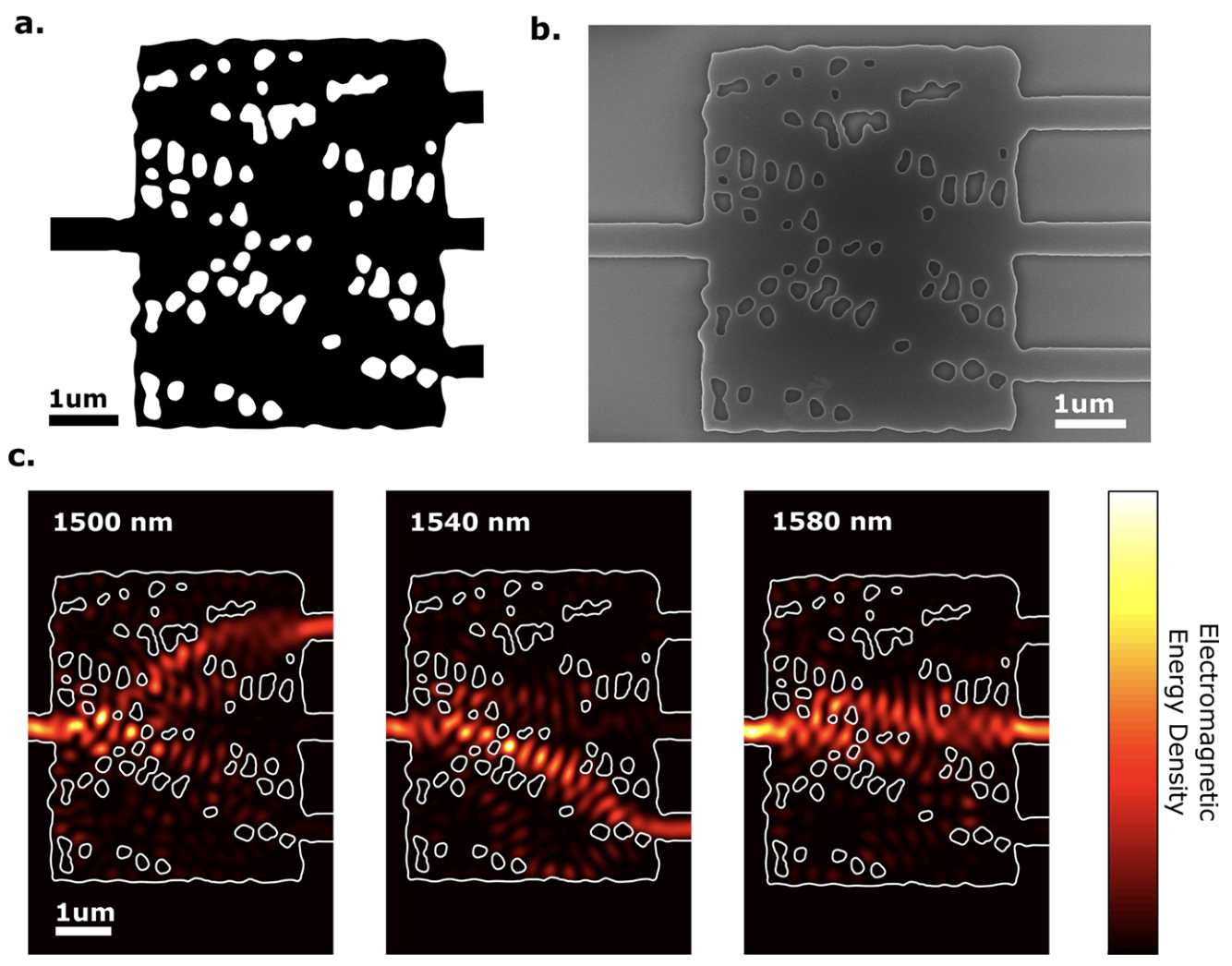}
         \caption{Black represents silicon (Si) and white represents silica ($\ce{SiO2}$).}
         \label{fig: demultiplexer3}
     \end{subfigure}
     \hfill
     \begin{subfigure}[b]{0.45\textwidth}
         \centering
         \includegraphics[width=\textwidth]{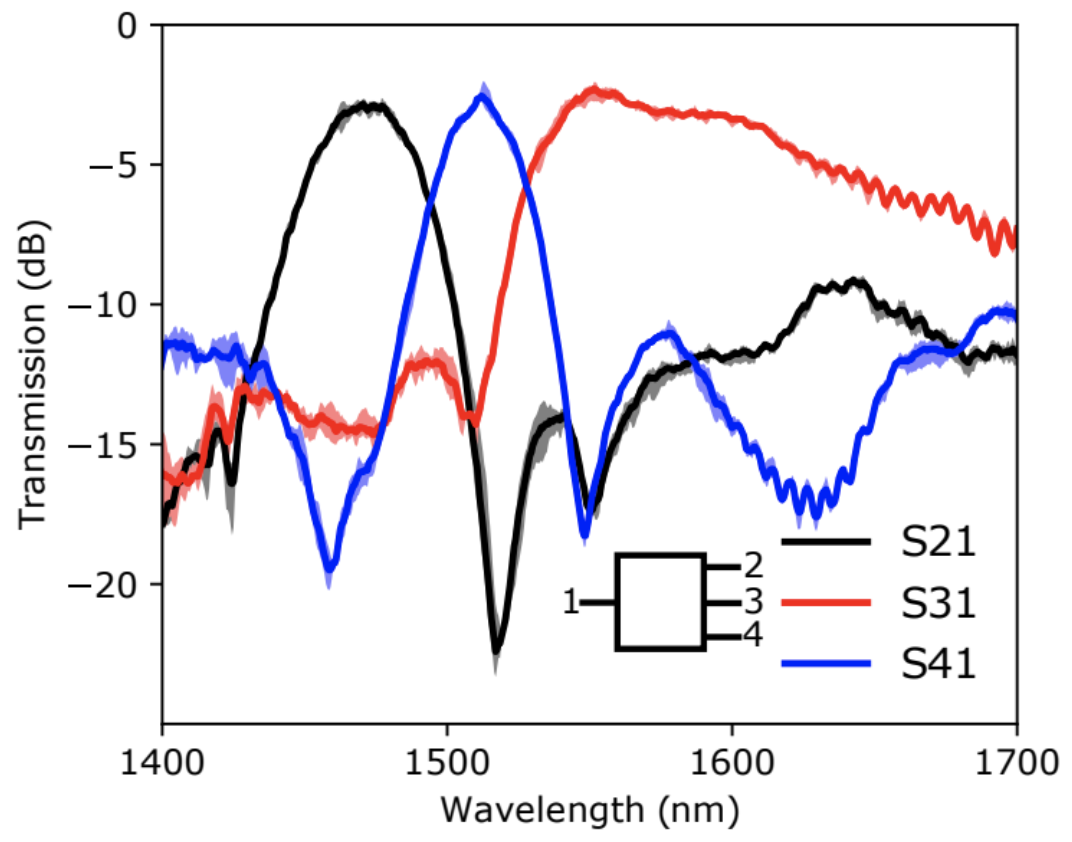}
         \caption{Transmission of the demultiplexer.}
         \label{fig: demultiplexer transmission}
     \end{subfigure}
        \caption{The creation of an optical device separating light of wavelengths 1500 nm, 1540 nm and 1580 nm. Images taken from \cite{Su2018}.}
        \label{fig: demultiplexer}
\end{figure}

$\\$
Density topology optimization and the adjoint method were also used to design a demultiplexer that separates light of wavelengths 1300 nm and 1550 nm \cite{Piggott2015}. This demultiplexer was already shown in figure \ref{fig: nontrivial design}. The permittivity varied continuously between the permittivity of air, silicon and $\ce{SiO2}$. Three years layer, the same research group created a demultiplexer to separate 1500 nm, 1540 nm and 1580 nm \cite{Su2018}, shown in figure \ref{fig: demultiplexer}. The inverse design problem was in both cases described by density topology optimization. 

$\\$
\noindent The historical overview that we gave is based on the great review by Molesky et al. \cite{Molesky2018}. The interested reader is referred to this review for more information on the development of inverse design in Nanophotonics.

\clearpage

\section{Deep Learning based inverse design}

The methods mentioned above have proven their success, but there are some drawbacks. First of all, the optical devices we are able to design are getting increasingly complex. Current methods do not scale well with the increasing number of parameters required for these complex structures. Secondly, in order to do an iteration of gradient descent, a full Maxwell equation simulation has to be run. This takes quite a while for just one simulation. No matter how advanced methods become, as long as inverse design depends on these simulations, the computation time will be high.

$\\$
Deep Learning can provide a way to go beyond these limitations. On the first point, the research on Deep Learning suggests that they are fully capable to scale up to any number of parameters. In the computer vision literature, neural networks work with images of several megapixels, which are millions of parameters. Nonetheless, results in computer vision are stunning. This leads us to believe that Deep Learning techniques are also able to scale with increasingly complex nanophotonic structures. Secondly, Deep Learning techniques are able to drastically reduce computing time. The training of the network takes some time, but once it is trained, it can make predictions in several milliseconds. This drastically speeds up the time it takes to do an iteration of gradient descent. This allows us to search a larger region of the parameter space and to create an even better performing design.

\begin{figure}[h!]
    \centering
    \includegraphics[width=0.8\textwidth]{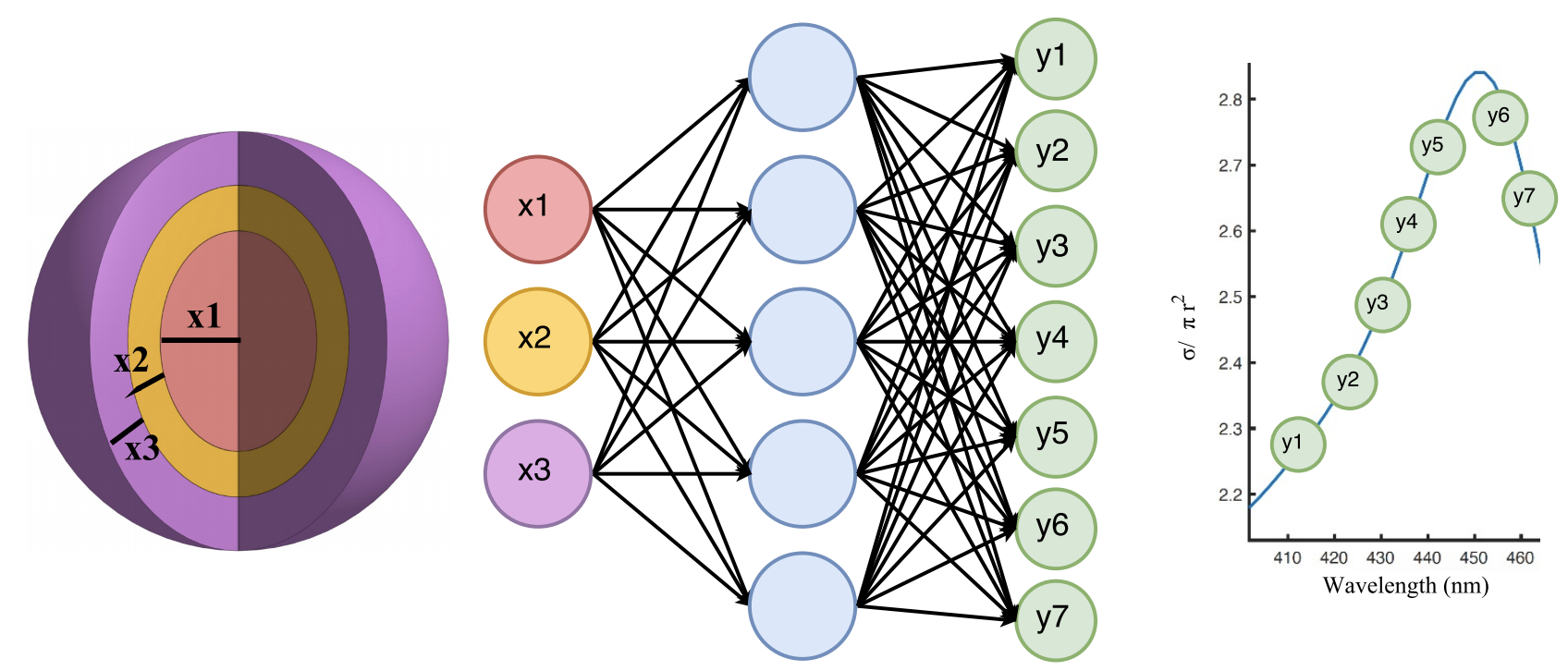}
    \caption{Nanoparticle design with a neural network. The network is trained to map the thicknesses of different shells to the scattering cross section in function of the wavelength of the scattered light. Image from Peurifoy et al. \cite{Peurifoy2017}}
    \label{fig: peurifoy}
\end{figure}

\noindent The paper that inspired a lot of the work in this thesis was written by Peurifoy et al. in 2018 \cite{Peurifoy2017}. The authors applied Deep Learning to the inverse design of nanoparticles, see figure \ref{fig: peurifoy}. These are layered particles where the thickness of each layer determines their behaviour. A neural network mapped the thicknesses of each layer to the scattering cross section as a function of the wavelength. Since this network is fully differentiable, it could be used to perform gradient descent on the design parameters. We follow a similar approach in this thesis. So et al. improved the inverse design of nanoparticles in 2019 \cite{so2019}. The authors of this paper adapted the loss function of the neural network to learn the thickness of each layer as well as the material it is made of. 

$\\$
Besides gradient-based optimization, one can also train a second neural network to predict the design. This network then maps an optical response directly to the design parameters. A challenge is that this is not a one-to-one mapping. For a given response, many different designs are valid solutions. This is something difficult to learn for a neural net. This fundamental problem of non-uniqueness was overcome using a tandem network by Liu et al. \cite{Liu2017}. The idea of the tandem is to couple a first network to predict a response to a network that uses this response to predict the original design. The two networks are then trained simultaneously. The authors applied their method to the transmission spectrum of multilayer structures of alternating layers of $\ce{SiO2}$ and $\ce{Si3N4}$.  

\begin{figure}[h!]
    \centering
    \includegraphics[width=0.7\textwidth]{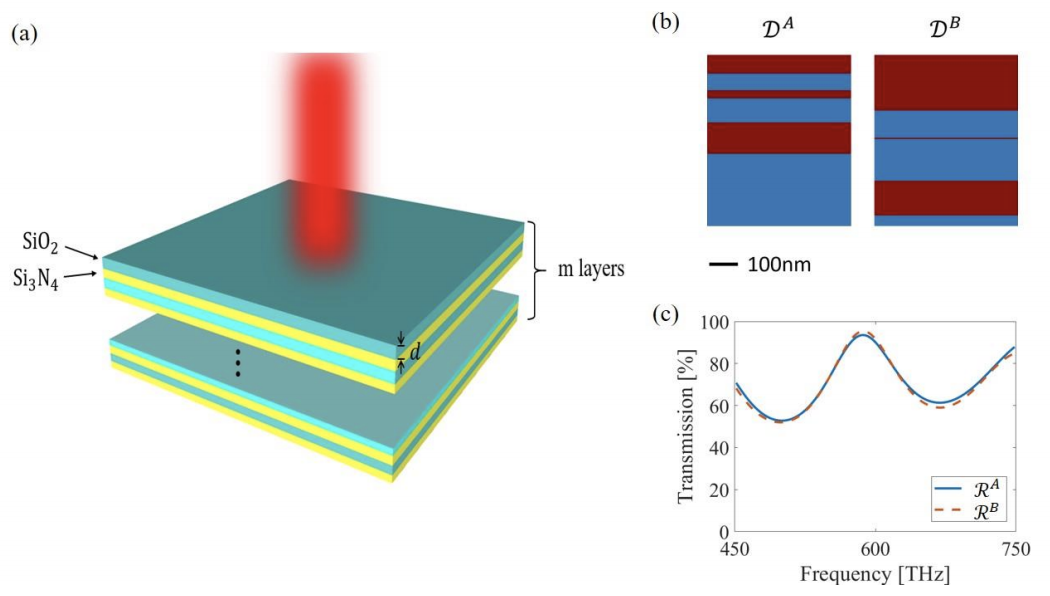}
    \caption{Inverse design of the transmission of a multilayer structure by Liu et al. \cite{Liu2017}. a) The design specified by the thicknesses of $m$ layers. b) Two example designs A and B with 6 layers. c) Transmission of the 6-layer designs in b.}
    \label{fig: tandem Liu}
\end{figure}

\begin{figure}[h!]
    \centering
    \includegraphics[width=0.5\textwidth]{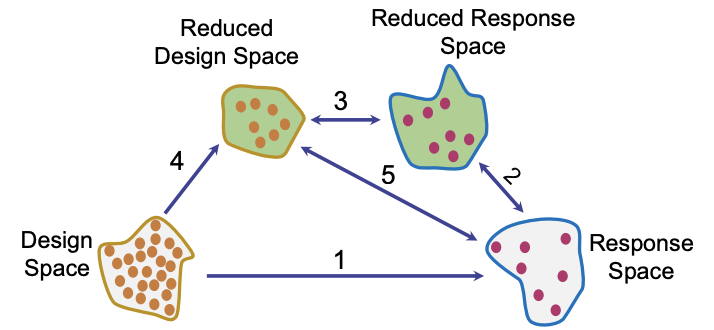}
    \caption{Dimensionality reduction on both the design space and the response space. Image from Kiarashinejad et al. \cite{Kiarashinejad2020}.}
    \label{fig: kiarashinejad}
\end{figure}

\noindent A method to gain more insight into inverse design was proposed by Kiarashinejad et al. \cite{Kiarashinejad2020}. They used an autoencoder to reduce the dimension of the design space and the response space, shown in figure \ref{fig: kiarashinejad}. They applied their method to design the reflectivity of a metasurface. The dimensionality reduction allowed them to overcome the non-uniqueness of optimal designs. 

\begin{figure}[!h]
    \centering
    \includegraphics[width=0.9\textwidth]{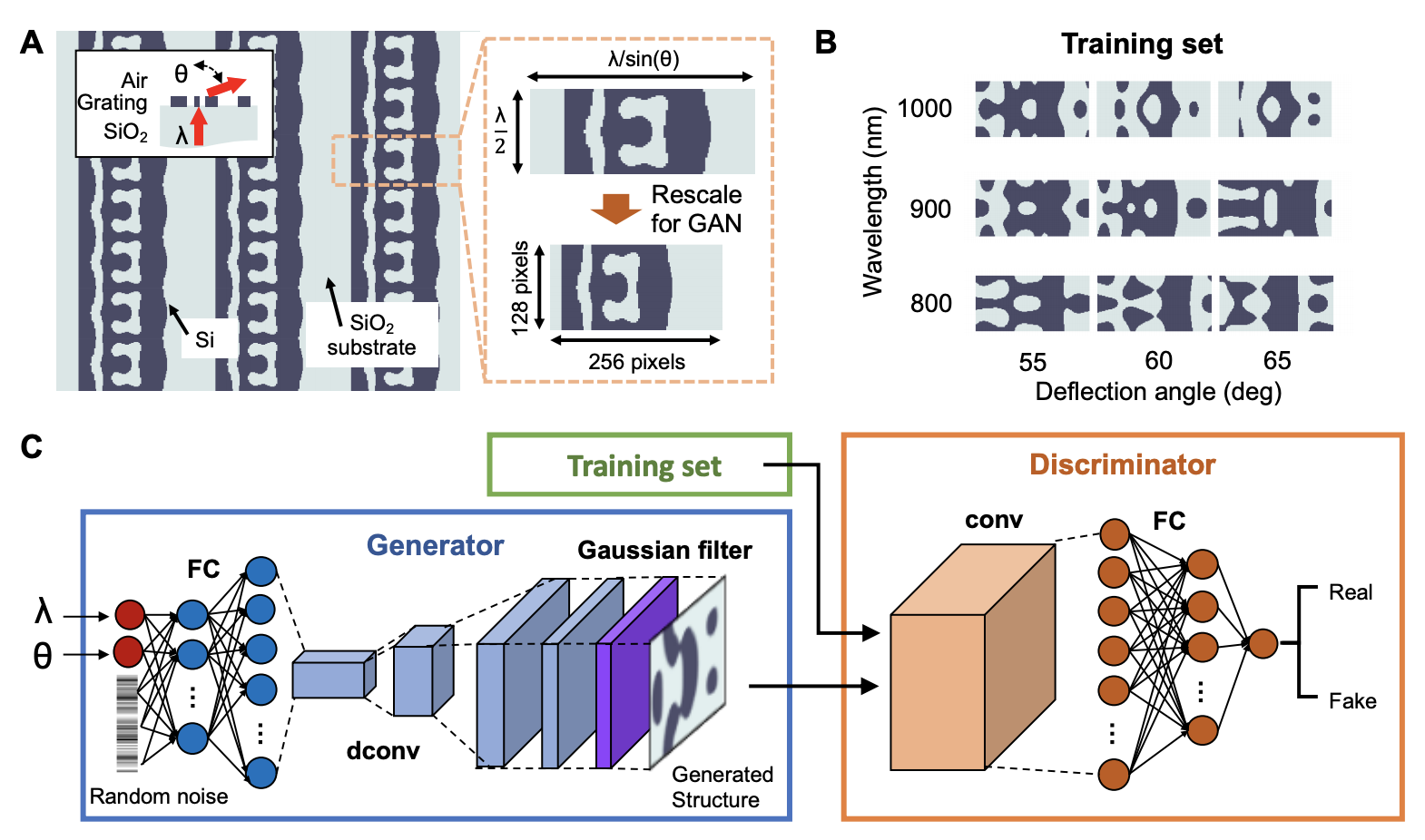}
    \caption{Inverse design of a metagrating with a GAN \cite{Jiang}. a) Problem set-up: light of wavelength $\lambda$ is deflected at an angle $\theta$. b) Training set containing topology-optimized metagratings. c) Network architecture.}
    \label{fig: GAN}
\end{figure}

$\\$
A somewhat different kind of artificial network called the \textbf{Generative Adversarial Network (GAN)} has also been used for inverse design by Jiang et al. \cite{Jiang}. The generator network of the GAN generates an image of the optical structure based on a random noise vector. The generator is trained simultaneously with a discriminator. The goal of the discriminator is to determine whether an image came from the training set or from the generator. The GAN was used to create an optical material to deflect light at a certain angle, see figure \ref{fig: GAN}. The network was trained with designs made by topology optimization. The training sets contained metagratings that deflect light with a high efficiency. This shows that GANs can also be used for inverse design in Nanophotonics.

%% file: chapters/chapter3.tex
\noindent Deep Learning is a subfield of Machine Learning that uses deep neural networks to learn from data. These artificial neural networks are inspired by the way neurons process information in the brain. The activation of one neuron leads to the activation of another and so forth. Similarly, information in an artificial neural network is processed by a succession of activated neurons. One of the greatest differences between Deep Learning and Machine Learning is its ability to learn from raw data. In Machine Learning, there is usually a large part of the work devoted to finding suitable features. This is completely avoided in Deep Learning. The disadvantage is that Deep Learning models are more of a black box than other Machine Learning models. This makes it difficult to understand what is going on under the hood.

$\\$
Researchers started to think about neural networks for the first time in the fifties. One of the pioneers was Frank Rosenblatt who made a neural network with one layer which he called the perceptron in 1958 \cite{Rosenblatt1958}. It is interesting to note that he was a psychologist, inspired by the working of the brain. It shows that Deep Learning has been interdisciplinary from the very beginning. After some initial success, it was the inability to find an efficient training algorithm that hindered the development of neural networks. This changed in 1986 with the conception of the backpropagation algorithm by Rumelhart, Hinton and Williams \cite{Rumelhart1986}. This algorithm consists of a very clever trick called automatic differentiation. It allows us to compute the gradient of a loss function with respect to thousands or even millions of parameters, and this in just several milliseconds. This laid the foundations for Deep Learning as we know it today.

$\\$
The moment where Deep Learning really started to boom was at the inception of Alexnet in 2012 \cite{alexnet2012}. This was the first time a neural network with many layers was efficiently trained on a large data set. The network was applied to the ImageNet dataset of 1.2 million real life images. The task is to classify these images into 1000 categories. A common metric to assess the performance on this data set is the top-5 error. This metric indicates how often the 5 most likely categories predicted by the network match the correct labels. Alexnet achieved a top-5 error of 17\%, which was a tremendous leap foward compared to earlier Machine Learning methods achieving a top-5 error of 26.2\%. This was the start of the massive popularity of Deep Learning we see today.

$\\$
In the following subsections, we explain Deep Learning for both supervised and unsupervised learning. We focus on the neural networks used in this work. For a more elaborate review, interested readers are referred to a great introduction to Machine Learning for physicists \cite{Mehta_2019}. More general explanations of Deep Learning are found in the books by Bishop \cite{bishop} or Goodfellow, Bengio and Courville \cite{GoodBengCour16}.

\section{Supervised learning}

\begin{figure}
    \centering
    \includegraphics[width=0.6\textwidth]{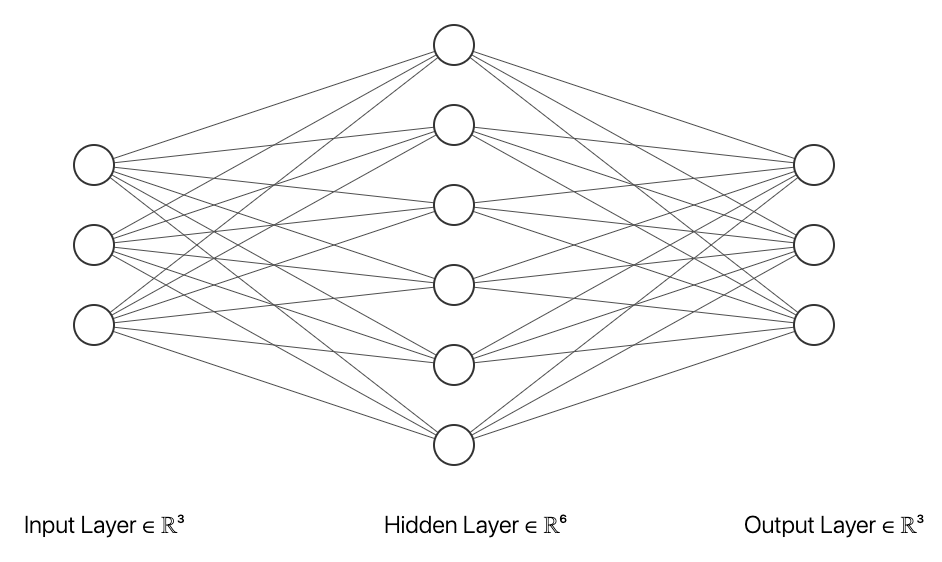}
    \caption{Architecture of a neural network. An input vector gets mapped through one or more hidden layers. Each of these hidden layers contains a vector which we call the activation vector. The activation vector of the final hidden layer is eventually transformed into an output vector.}
    \label{fig: neural network}
\end{figure}

In the context of supervised learning, neural networks provide a nonlinear mapping from an input vector $\bf{x}$ to an output vector $\bf{y}$. These nonlinear functions are parameterized by a layered architecture, resembling the working of neurons in the brain. Each layer consists of a number of nodes. In figure \ref{fig: neural network}, the input layer represents the vector $\bf{x}$ and the output layer represents the output vector $\bf{y}$. In between are a set of hidden layers. The vectors in these layers are called activation vectors. In the analogy to the brain, a large value for a node represents the firing of a neuron. To get from one layer to the next, we need to perform two operations: a linear transformation and a nonlinear activation function. For the linear transformation in the first layer we get 

\begin{equation}
    \nonumber
    \bf{z_1} = \bf{W_{in, 1}} \cdot \bf{x} + \bf{b_1}.
\end{equation}
We call the vector $\bf{z_1}$ the preactivation. $\bf{W_{in, 1}}$ is called the weight matrix and has dimension $n$ x $m$, where $n$ is the number of nodes in the first hidden layer and $m$ is the length of the input vector. The vector $\bf{b_1}$ of dimension $n$ is called the bias vector. During training, the parameters $\bf{W_{in, 1}}$ and $\bf{b_1}$ are optimized to get an optimal neural network. Often they are collectively referred to as the weights of the neural network. After the linear transformation, a nonlinear activation function $\sigma$ is applied so we get

\begin{equation}
    \nonumber
    \bf{a_1} = \sigma(\bf{z_1}). 
\end{equation}
The vector $\bf{a_1}$ is the activation vector of the first hidden layer. For a 1-layer neural network, we can then compute the output $\bf{y}$ as 

\begin{equation}
    \nonumber
    \bf{y} = \bf{W_{2}} \cdot \bf{a_1} + \bf{b_{2}}
\end{equation}

\noindent Even with only one layer, a neural network already has quite remarkable properties. The \textbf{universal approximation theorem} states that this architecture is able to approximate any function to arbitrary precision. A visual proof is given in chapter 4 of the online book by Michael Nielsen \cite{proofunivapprox}. The proof shows that the parameters of the neural net can correspond to the height and width of a step function. It then follows that a neural network can approximate any combination of step functions. Since a combination of many step functions can approximate any function, it follows that a 1-layer network is also capable of making this approximation.  

$\\$
Theoretically, the universal approximation theorem tells us that we should look no further than one layer neural networks. They are able to learn any function and are therefore equipped to tackle any problem we can imagine. In practice however, one layer would need a huge amount of nodes to tackle even the most basic problems. Moreover, it turns out to be quite hard to train the parameters of this network in an efficient way. A solution is to stack multiple layers. This means that we take the activation $\bf{a_1}$ of the first layer as the input to a second layer. We then get for the activation of the second layer

\begin{equation}
    \nonumber
    \bf{a_2} = \sigma(\bf{z_2}) = \sigma(\bf{W_{1, 2}} \cdot \bf{a_1}+ \bf{b_2}).
\end{equation}
The weights $\bf{W_{1, 2}}$ and biases $\bf{b_2}$ are different and completely independent from the weights and biases in the first layer. The nonlinear activation function $\sigma$ can in principle also be different in each layer. It is however common to choose the same activation function for every layer. In the following, we therefore consider the function $\sigma$ to be the same across the network. Let us now discuss some activation functions we can choose for our neural network.

\subsection*{Activation functions}

The activation function has a very important role in the neural network. Without the activation function, the layers would represent one linear transformation after another. This would result in a total transformation that would again be linear, such that the network would only be able to learn linear functions. It is therefore crucial to include a nonlinear activation function in every layer to be able to approximate more general functions. 

\begin{figure}
    \centering
    \includegraphics[width=0.9\textwidth]{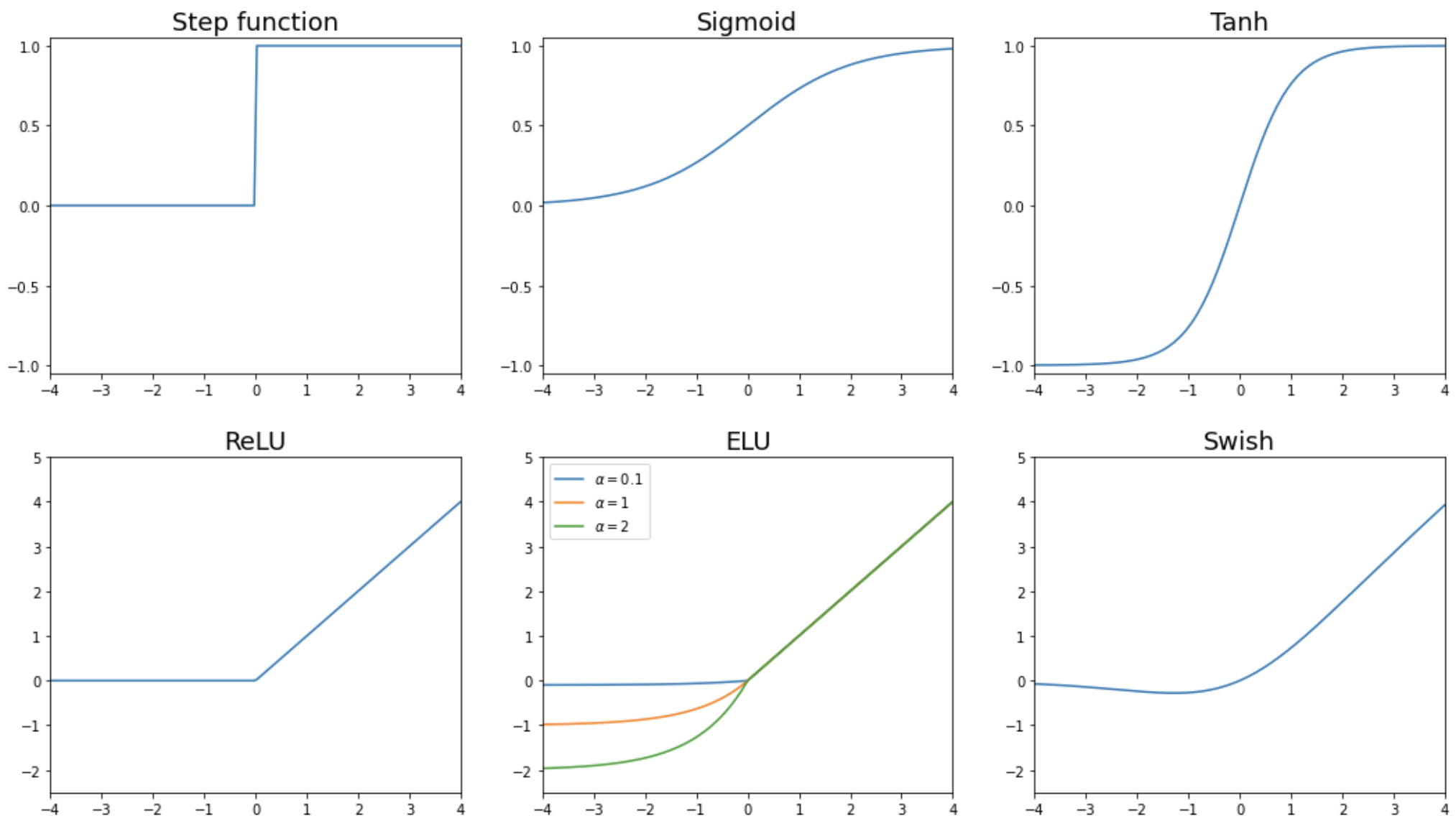}
    \caption{The most popular activation functions for neural networks. The step function was historically the first activation function that was used. Later it was replaced by sigmoid and tanh, because they are more continuous and therefore it is easier to compute gradients. For deep neural networks, an activation function from the ReLU family is preferred, since these functions solve the vanishing gradient problem.}
    \label{fig: activation functions}
\end{figure}

$\\$
Historically, the first activation function was the step function. This led to problems when computing gradients, since the step function has a discontinuity at 0. To solve this problem, the step function was replaced by similar but more smooth functions like sigmoid and tanh. These functions are shown in figure \ref{fig: activation functions}. The sigmoid function is given by

\begin{equation}
    \nonumber
    \sigma_{sigmoid}(x) = \frac{1}{1 + e^{-x}}.
\end{equation}

\noindent These activation functions worked very well in the early days when networks had only a few layers and led to applications like the digit recognition of postal codes by Yann LeCun et al. in 1990 \cite{lecun1990}. Their network only had 4 hidden layers. When more and more layers were added to the network however, these activation functions started to fail. The main reason being that they saturate for large values. These large values are mapped to a plateau where the derivative of the function is close to zero. As this happens in subsequent layers, the gradients become smaller and smaller. This is called the \textbf{vanishing gradient problem}. If this occurs, the gradient in the early layers becomes so small that no steps are taken and nothing is learned. 

$\\$
In order to overcome this problem, new activation functions were proposed that do not saturate for large values. The most popular of them is the \textbf{Rectified Linear Unit (ReLU)}, for which we have

\begin{equation}
    \nonumber
    \sigma_{ReLU}(x) = \left\{ 
    \begin{matrix}0 & \mbox{for } 0 \leq x, \\ 
    x & \mbox{for } 0 > x.
    \end{matrix}\right.
\end{equation}

\noindent The derivative of this function is 0 for negative values and 1 for positive values. Large values are therefore multiplied by 1 when propagating through this activation function, such that the gradient does not vanish. This allows the training of very deep neural networks with many layers, hence the name Deep Learning. Alexnet mentioned earlier was one of the deepest at the time of its inception in 2012, containing 8 hidden layers. 3 years later in 2015, researchers at Microsoft created a much more advanced network for image recognition called ResNet \cite{He2016}. This network has a stunning number of 152 hidden layers. The network was also tested on the Imagenet dataset and obtained a top-5 error of just 3.57\%. This is a big improvement compared to Alexnet which achieved a top-5 error of 17\%. This supports the empirical knowledge that deeper networks with a larger number of layers perform better.

$\\$
A variation of ReLU is the \textbf{Exponential Linear Unit (ELU)}. Unlike ReLU, the ELU does not have zero activation for negative values. Instead, it has activations decreasing exponentially for lower values. The function is given by 

\begin{equation}
    \nonumber
    \sigma_{ELU}(x) = \left\{ 
    \begin{matrix}\alpha (e^x - 1) & \mbox{for } 0 \leq x, \\ 
    x & \mbox{for } 0 > x.
    \end{matrix}\right.
\end{equation}

\noindent The ELU activation introduces a value $\alpha$. We are free to choose a value for $\alpha$ that is best suited to the problem. The ELU is only one of the variations of ReLU. A lot of other variations exist in the literature \cite{nwankpa2018activation}.

$\\$
The ReLU activation has become the current standard for training deep neural networks. Nonetheless, a large scale study to find the best activation function was performed in 2017 by Google Brain \cite{ramach2017}. Reinforcement learning was used to search a large space of possible activation functions. The search in this function space was guided by training a neural network with possible optimal activation functions and assessing the performance of this network. Let us think about this for a moment. This is AI discovering how to improve AI. We believe this to be quite the conceptual breakthrough. The activation function that was found to be the best was the \textbf{Swish}, given by

\begin{equation}
    \nonumber
    \begin{aligned}
        \sigma_{swish}(x) &= x \cdot \sigma_{sigmoid}(x), \\
        &= \frac{x}{1 + e^{-x}}.
    \end{aligned}
\end{equation}

\subsection*{Learning dynamics}

So far, we have described the architecture of a neural network. We saw that it consists of layers applying a transformation on the output of the previous layer. The transformation in each layer is given by

\begin{equation}
    \nonumber
    \bf{a_{i+1}} = \sigma(\bf{z_{i+1}}) = \sigma(\bf{W_{i, i+1}} \cdot \bf{a_i} + \bf{b_{i+1}}).
\end{equation}
When we set up a neural network, the size of the input vector $\bf{x}$ and the output vector $\bf{y}$ are determined by the problem. It is then up to us to determine the number of hidden layers and the number of nodes in each layer. There has been a lot of research on the optimal number of layers and nodes for a given problem, but so far all results are empirical. They are not backed up by a formal theoretical understanding. Creating the perfect architecture for the problem at hand remains thus more of an art than a science.

$\\$
After picking the numbers of layers and nodes, we fix the activation function $\sigma$ in each layer. The activation functions could be different in each layer, but it is more common to choose the same activation for all layers in the network. When the network has a large number of layers, it is best to pick an activation function that does not saturate to a constant value for large inputs, like ReLU or Swish. Once we have specified the architecture of the neural network, we can start training. 

$\\$
We can train a neural network by feeding it input data $\bf{x}$ together with the corresponding label $\bf{y}$. The network will then adapt its \textbf{trainable parameters} in order to fit the network to the data. These trainable parameters are the weights $\bf{W}$ and biases $\bf{b}$ of the linear transformations in each layer. These parameters are updated by means of gradient descent. For quite a long time, it was infeasible to compute the gradients of thousands of parameters for thousands of data samples in every iteration. This changed when the backpropagation algorithm was invented. This algorithm massively speeds up the computation of gradients and this makes the training of a neural network much faster. 

$\\$
Next to developments on the software side, there is also a development on the hardware side that sped up the training of neural networks. In 2012, Alexnet was the first network to be trained on a \textbf{Graphics Processing Unit (GPU)}. This is a processing unit found in most laptops and computers today to render high quality images for video games or movie streaming. What makes this processing unit so special, is that its hardware is designed for fast matrix operations. Therefore, neural networks today are nearly always trained on a GPU. For reference, the training of a neural network with 6 layers on 50000 training samples $\bf{x}$ and $\bf{y}$ takes only 10-15 minutes on a modern GPU. 

$\\$
Using the backpropagation algorithm, the parameters of the network are trained by gradient descent. First, a loss function is constructed that describes how well the neural network performs. We can write the neural network as a function $f_{\bf{w}}$, indexed by its trainable parameters $\bf{w}$. The predictions of the network can then be written as $f_{\bf{w}}(\bf{x})$. The loss is a function $L ( f_{\bf{w}}(\bf{x}), \bf{y})$ of these predictions and the correct labels. A very popular loss function is the \textbf{mean squared error (MSE)} given by

\begin{equation}
    \nonumber
    L = (\hat{\bf{y}}- \bf{y})^2
\end{equation}

\noindent This loss is then minimized by updating the trainable parameters $\bf{w}$. To do this, gradients of the loss are computed with respect to the parameters $\bf{w}$. The parameters are then updated in the opposite direction of the gradient. The step size of this update is determined by the $\textbf{learning rate}$ $\eta$. For the update in iteration $i$ we have

\begin{equation}
    \nonumber
    \mathbf{w_{i+1}} = \mathbf{w_i} - \eta \nabla_{\mathbf{w_i}} L 
\end{equation}

\noindent After a few iterations, the parameters $\bf{w}$ converge to minimize the loss $L$. To perform the parameter update above, we need to compute the gradient of the loss on the training data. However, when the data contains thousands of data samples, it is not feasible to load all of this data into RAM memory which typically has only 8GB-32GB of memory. An improved gradient descent algorithm is therefore $\textbf{Stochastic Gradient Descent}$. This algorithm divides the data randomly into smaller sets of equal size called $\textbf{batches}$. In every iteration, the gradient is computed on one of these batches and a parameter update is made. Every parameter update is thus based on only part of the training data. In this way, stochastic gradient descent allows us to use a lot of data for training. 

$\\$
The learning rate $\eta$ determining the step size can be a constant or can be adapted in every iteration. There are many algorithms to adapt the learning rate in every iteration and they are collectively called optimization algorithms. A great review on the most popular gradient descent optimization algorithms for Deep Learning is given by Ruder \cite{Ruder2016}. The optimization algorithm we use throughout this thesis is the Adam optimization algorithm proposed by Kingma and Ba \cite{adam2015}. 

\section{Unsupervised learning}

The algorithms of supervised learning we have seen so far are based on pattern recognition. They are able to learn the patterns in the input data $\bf{x}$ and then use these patterns to predict an output $\bf{y}$. This approach has led to remarkable progress in image processing, speech recognition and natural language processing. While these achievements are very exciting, pattern recognition is only one aspect of the artificial intelligence we would like to create. An important other part of intelligence is reasoning. This is one of the goals of unsupervised learning.

$\\$
When a neural network trained by supervised learning sees an image or an equation, it is not able to think about it. It uses the patterns it sees in the image in order to obtain a certain output, but it does not understand what the data means. If we want artificial intelligence to also reason about what it sees, a crucial step is the ability to form representations of the world around us. As a human we do this all the time. In Physics for example, we can represent objects as point particles governed by the laws of classical mechanics. Finding this representation of the world is what caused Newton to come up with his laws of mechanics. The key insight he had, was that the Sun, the Moon and an apple could all be represented as a point particle with a certain mass, governed by the same laws. This insight allowed Newton to make predictions about the world that were never seen before. This insight created Physics as we know it today. 

$\\$
The benefits of finding a good representation explain why there is a growing need to develop neural networks that are not only able to recognize patterns, but are also able to create meaningful representations. We want these representations to have two properties. We want an optimal representation to be both as small as possible and to retain all useful information. 

$\\$
The notion of useful information is of course related to the problem we consider. When thinking about classical gravity for example, we only need to know the mass of each body to do our calculations. If we are working with classical electromagnetism on the other hand, we also need to know the charges of the particles. It is in fact quite a challenge to get only that part of the data that is needed to solve a problem, see for example the average Physics student trying to decipher an exam question. There are many approaches to achieve meaningful representations. We will discuss a method that is very popular in Deep Learning called an \textbf{autoencoder}.

\subsection*{Autoencoder}

The autoencoder was originally proposed in the context of representation learning by Hinton and Salakhutdinov in 2006 \cite{autoenc2006}. An autoencoder tries to form a representation by pushing the input through a bottleneck and then asking the network to make an accurate reconstruction of the input. To make a good reconstruction, the autoencoder needs a good representation in its bottleneck. This is the latent vector $\bf{z}$. By making this latent vector a lot smaller than the input, the data needs to be compressed but still retain the useful information to reconstruct the input. 

\begin{figure}
    \centering
    \includegraphics[width=0.9\textwidth]{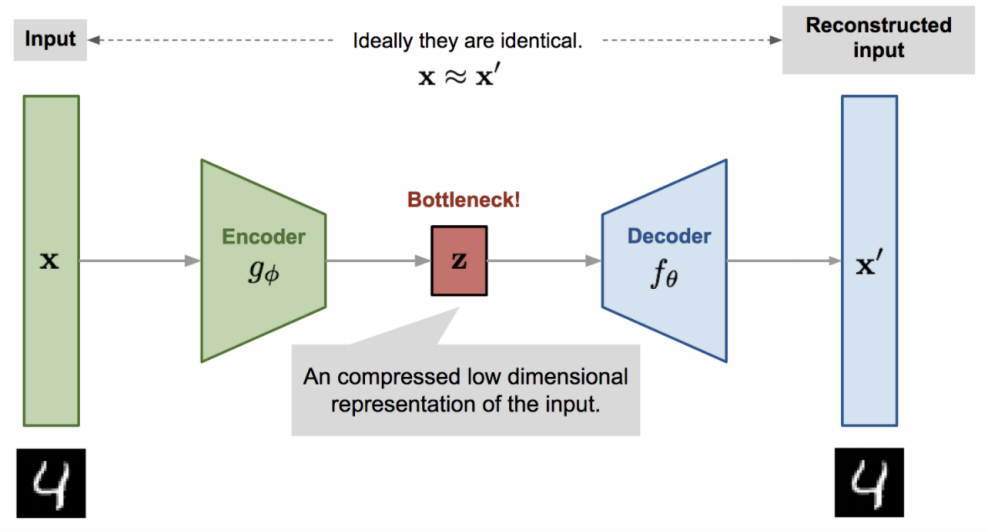}
    \caption{An autoencoder compresses its input data into a smaller bottleneck. In this case, the encoder maps an image of a handwritten digit to a latent vector $\bf{z}$. This latent vector needs to contain as much information as possible, such that the decoder can make an accurate reconstruction of the digit image. Image taken from \cite{weng2018VAE}.}
    \label{fig: autoencoder}
\end{figure}

$\\$
We call the neural network to go from the input to the latent vector the $\textbf{encoder}$ and the network that goes from this latent vector to a reconstruction of the input the $\textbf{decoder}$. This is shown in figure \ref{fig: autoencoder}. An autoencoder can be trained by any loss function that compares the reconstructed data with the original data. A common loss function is the mean squared error (MSE), given by

\begin{equation}
    \nonumber
    MSE = (\mathbf{x} - \mathbf{x'})^2.
\end{equation}

\noindent In this equation, $\bf{x'}$ is the reconstruction proposed by the decoder. Autoencoders are very useful for denoising data. In this regard, a noisy image is fed into the autoencoder and a noiseless image is expected as output. This is possible because the noise will be random and therefore contain no learnable pattern. The autoencoder will only retain the structure it can find in the image, and this structure is exactly the original image. 

\noindent Recently, the autoencoder was adapted to a quantum version that could denoise data in quantum entanglement \cite{Bondarenko2020}. This shows that autoencoders have proven their use already in Physics. The interested reader who would like to know more about autoencoders in general is referred to a great review in \cite{weng2018VAE}. 

\begin{figure}
    \centering
    \includegraphics[width=0.5\textwidth]{{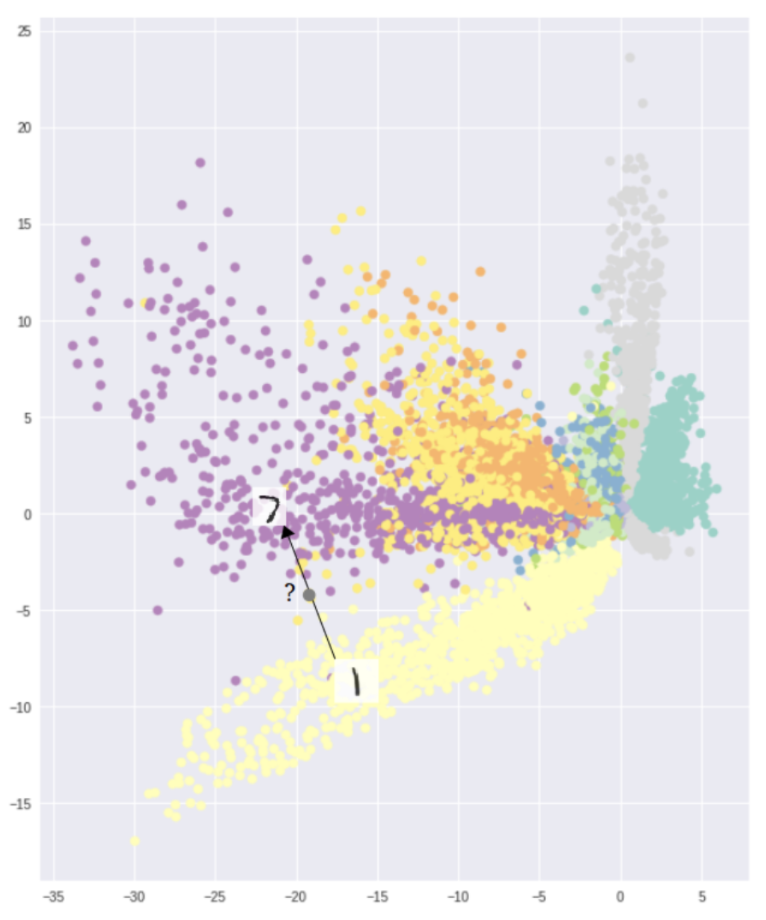}}
    \caption{This image shows the two-dimensional latent space of the MNIST dataset obtained by an autoencoder. Each point represents the encoding in the latent space of a handwritten digit. The color of the points indicates the number of the handwritten digit. Different clusters are formed for each of the numbers. We also observe that it is not possible to interpolate between a 1 and a 7, since the latent space in the area between those two cluster leads to reconstructions that do not look like numbers. Image taken from \cite{aevsvae}.}
    \label{fig: problem autoencoder}
\end{figure}

$\\$
One problem with autoencoders, however, is that they are not able to form continuous latent representations. This can be see in figure \ref{fig: problem autoencoder}. An autoencoder was used on the MNIST data set. This dataset contains images of handwritten numbers from 0 to 9. The images are represented as two dimensional latent vectors. The data points are colored depending on the number they represent. We observe that distinct clusters are formed in the latent space. There is no interpolation possible between data samples in different clusters. For example, we can not go from a 7 to a 1 in a continuous way. Moreover, it also means that the values in between clusters do not have a meaningful reconstruction. This is not very desirable if we think about what this means for representations in Physics. 

$\\$
When we think about gravity for example, the latent representation should correspond to mass. From a physical perspective, the behaviour of an object in a gravitational field should be continuously dependent on its mass. It does not make much sense to have a model that is able to predict the gravity of apples with a mass of 0.1 kg, of the earth with a mass of $6 \cdot 10^{24}$ kg, but is unable to say what would happen to the African bush elephant with a mass of 6000 kg. This was one of the motivations to create a new architecture called the \textbf{Variational Autoencoder (VAE)}. After its inception in 2014 by Kingma and Welling \cite{kingma2013}, this model has become very popular for generative modelling in recent years.

\subsection*{Variational Autoencoder (VAE)}

Variational autoencoders (VAEs) are very suited to solve this problem, because their latent spaces are continuous by design. The main difference with an autoencoder is the following. In the autoencoder, both the encoder and the decoder are deterministic mappings $\bf{x} \mapsto \bf{z}$ and $\bf{z} \mapsto \bf{x'}$. The idea of the VAE is to make the encoder and the decoder probabilistic. For the encoder, this means that instead of mapping to a fixed $\bf{z}$, the input $\bf{x}$ is mapped to a conditional probability $p(\bf{z} \vert \bf{x})$ over the latent variables. The decoder also becomes probabilistic. A latent vector $\bf{z}$ is sampled from $p(\bf{z} \vert \bf{x})$, and is then mapped to a conditional probability $p(\bf{x} \vert \bf{z})$ over the input space. It is this transition from deterministic to probabilistic encoders and decoders that will create a continuous latent space.

\begin{figure}
    \centering
    \includegraphics[width=0.7\textwidth]{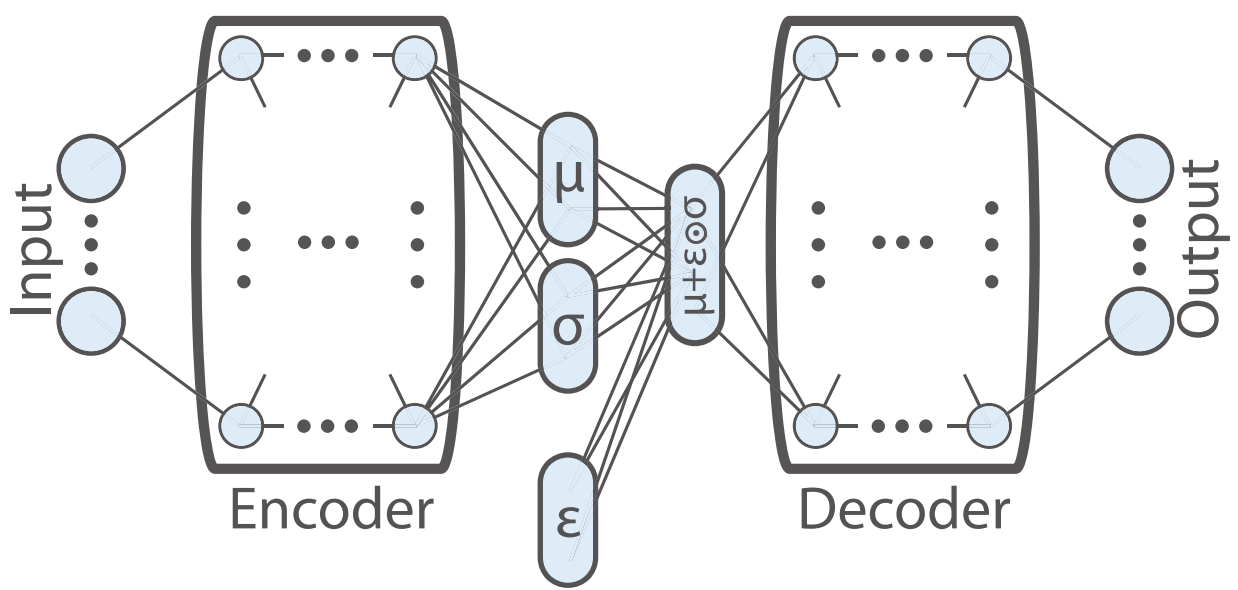}
    \caption{Architecture of the variational autoencoder (VAE). The decoder maps the input to a gaussian distribution $p(\bf{z} \vert \bf{x})$, parameterized by $\mu$ and $\sigma$. A latent vector is then sampled from this distribution using the reparameterization trick and further decoded into a reconstruction. Image taken from \cite{Iten2020}.}
    \label{fig: vae}
\end{figure}

$\\$
In order to learn the probability distribution $p(\bf{z} \vert \bf{x})$, we need to restrict it to a fixed family of functions. A common family of functions are multivariate gaussians. This means that every dimension of the latent vector $\bf{z}$ corresponds to a normal distribution $\mathcal{N} (\mu(\bf{x}), \sigma(\bf{x}))$, where the mean and standard deviation are functions of the input $\bf{x}$. This leads us to the architecture depicted in figure \ref{fig: vae}.

$\\$
For the decoder to make a reconstruction, we need to sample a latent vector $\bf{z}$ from this probability distribution $p(\bf{z} \vert \bf{x})$. Here we run into problems, because the sampling operation is not differentiable. This means that we can not compute gradients in the part before the sampling operation. This problem was solved by a clever idea called the \textbf{reparameterization trick.} Instead of sampling directly, a random node is added to the network. This random node samples a random vector $\epsilon$, where each component of the vector is drawn from a normal distribution $\mathcal{N} (0, 1)$. The latent vector $\bf{z}$ is then given by

\begin{equation}
    \nonumber
    \bf{z} = \bf{\mu} + \epsilon \cdot \bf{\sigma}.
\end{equation}

\noindent The reparameterization trick makes sure that the full architecture is differentiable. We can now explain how the change of a deterministic to a probabilistic architecture makes the latent representation continuous. There are two reasons for this. The first is that instead of mapping each input to a fixed point, inputs are mapped to a distribution in latent space. This can cause the encoding distributions of different inputs to overlap, which makes the latent space more continuous. 

$\\$
The second reason the latent space is continuous, is that the distributions $p(\bf{z} \vert \bf{x})$ are restricted to be close to unit gaussians $\mathcal{N} (0, 1)$. This makes sure that the network only uses a latent dimension if this dimension would contain useful information. Otherwise the dimension is empty and equal to the unit gaussian. Even when a latent dimension is activated, the distribution $p(\bf{z} \vert \bf{x})$ is still encouraged to be close to the unit gaussian. 

\begin{figure}
    \centering
    \includegraphics[width=0.4\textwidth]{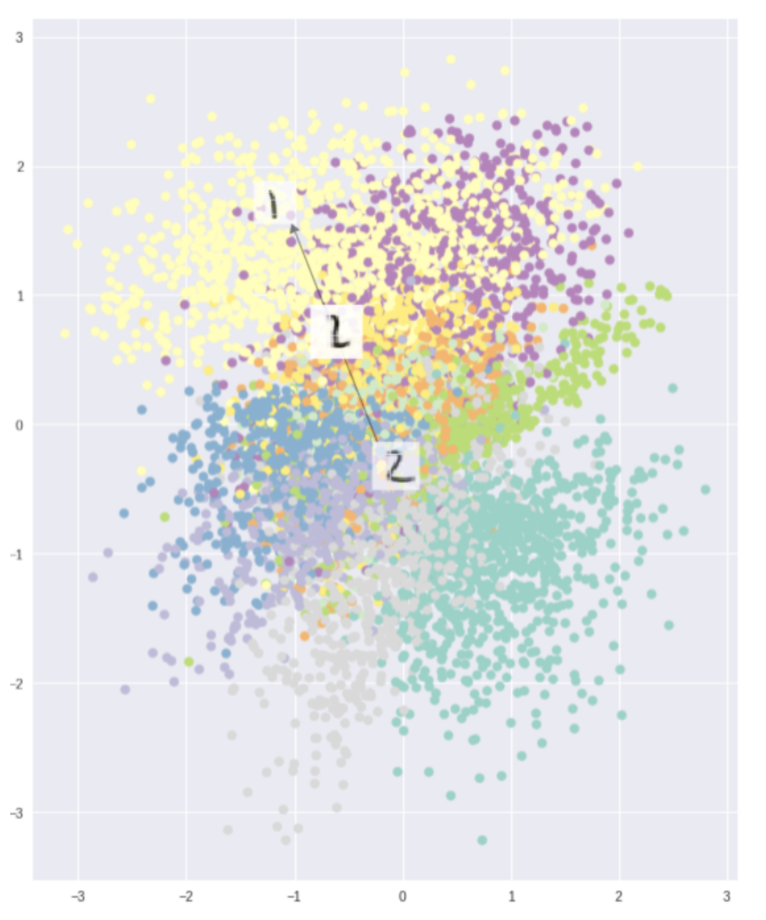}
    \caption{The latent space of the MNIST dataset obtained by a variational autoencoder (VAE). The latent representation is more continuous because the distributions in latent space are motivated to be close to unit gaussians $\mathcal{N} (0, 1)$. This continuous latent space allows us to meaningfully interpolate between digit images in the dataset. Image taken from \cite{aevsvae}.}
    \label{fig: solution vae}
\end{figure}

$\\$
The result is that most encodings in latent space are still centered around the origin. Gaining information about the inputs breaks spherical symmetry and creates different clusters in latent space, as in the case of the autoencoder. The restriction that the distribution $p(\bf{z} \vert \bf{x})$ should be close to unit gaussians will however cause a centering around the origin of the latent space. This trade-off leads to different clusters that are continuously distributed in latent space. This can be seen in figure $\ref{fig: solution vae}$.

$\\$
$\\$
We will now describe the intuitive reasoning above in more mathematical terms. We can not expect to learn the true conditional probability of the encoder $p (\bf{z} \vert \bf{x})$, since it can be arbitrarily complex. Therefore we turn to a statistical technique called $\textbf{variational inference}$. In this paradigm, we approximate the distribution within a certain family of distributions. We look for the distribution in this family that most closely matches $p (\bf{z} \vert \bf{x})$.  

$\\$
We look for an approximate function $q_{\phi}(\bf{z} \vert \bf{x})$ in the family of multivariate gaussians. If this function is given by a neural network, the parameters $\phi$ are the weights and biases of the network. We also parameterize the decoder as a neural network, which we call $p_\theta (\bf{x} \vert \bf{z})$. The optimal parameters $\phi^*$ and $\theta^*$ for the two neural networks are then given by minimizing a loss function. This loss function contains two parts, the first is the log-likelihood of the data $p_\theta (\bf{x} \vert \bf{z})$. This term describes how good the reconstruction is. The second term is the Kullback-Leibler divergence between $q_\phi(\mathbf{z} \vert \mathbf{x})$ and a prior distribution $p(\bf{z})$. This quantity measures the similarity between two distributions. In this way, we can constrain the distributions in the latent representation to be close to unit gaussians. In the end we have

\begin{equation}
    \nonumber
    \begin{split}
    (\phi^*, \theta^*) 
    &= \ \argmin_{\phi, \theta} \ \mathbb{E}_{z \sim q_{\phi} (\mathbf{z} \vert \mathbf{x})} \left[ - \text{log} \ p_{\theta}(\mathbf{x} \vert \mathbf{z}) \right] + D_{KL}(q_\phi(\mathbf{z} \vert \mathbf{x}) \vert \vert p(\bf{z})), \\
    &= \ \argmin_{\phi, \theta} \ \mathbb{E}_{z \sim q_{\phi} (\mathbf{z} \vert \mathbf{x})} \left[ \frac{(\mathbf{x} - f(\mathbf{z}))^2}{2c}\right] + \mathbb{E}_{\mathbf{z} \sim q_{\phi} (\mathbf{z} \vert \mathbf{x})} [\text{log} \ \frac{q_{\phi} (\mathbf{z} \vert \mathbf{x})}{p(\mathbf{z})}]. \\
    \end{split}
\end{equation}

$\\$
\noindent In the last step we assumed the decoder to follow a gaussian distribution

\begin{equation}
    \nonumber
    p_{\theta}(\mathbf{x} \vert \mathbf{z}) \sim exp \left[ -\frac{(\mathbf{x} - f(\mathbf{z}))^2}{2c} \right].
\end{equation}
The function $f(\bf{z})$ is the mean of this gaussian distribution. We can minimize the loss function with respect to the parameters $\phi$ and $\theta$ by stochastic gradient descent. This is how we can train the variational autoencoder. 

\subsection*{Disentangled representations}

The variational autoencoder allows us to create continuous latent representations of any input data we would like. This latent space is however not disentangled. What we mean by this, is that the different latent dimensions can be mixed. There might be two dimensions encoding for a mixture of mass and electric charge. In a \textbf{disentangled representation} however, every dimension corresponds to one independent degree of freedom. A popular way to obtain such a disentangled representation is the $\boldsymbol{\beta}$\textbf{-variational autoencoder ($\boldsymbol{\beta}$-VAE)}. 

$\\$
When we examine the loss function of the VAE more closely, we see that the first term can be interpreted as a reconstruction loss and the second term can be interpreted as a way to restrict the bottleneck. The first term pushes the model to retain more useful information in order to make the best reconstructions, and the second term makes sure that the latent representation is as small as possible. In a $\beta$-VAE, a parameter $\beta$ is introduced that determines the relative strength of these two terms. The loss function becomes 

\begin{equation}
    \nonumber
    (\phi^*, \theta^*) = \ \argmin_{\phi, \theta} \ \mathbb{E}_{z \sim q_{\phi} (\mathbf{z} \vert \mathbf{x})} \left[-\text{log} \ p_{\theta}(\mathbf{x} \vert \mathbf{z}) \right] + \beta \cdot D_{KL}(q_\phi(\mathbf{z} \vert \mathbf{x}) \vert \vert p(\mathbf{z})). 
\end{equation}

\noindent We need to optimize the parameter $\beta$ in the same way as we optimize the number of layers and nodes in the network, by trial and error. For a suited value of $\beta$, we can indeed obtain a disentangled representation where each dimension is independent of the others. In the context of Physics, this means that each dimension corresponds to an independent degree of freedom. This idea was applied by Iten et al. \cite{Iten2020}. In their paper, it was used to discover heliocentrism, to make predictions about the harmonic oscillator and to discover the conservation of angular momentum in mechanical systems. In my thesis, I applied the same network to determine the degrees of freedom in the Fabry-Pérot resonator. 

$\\$
This completes the short introduction to Deep Learning. It is a very new and exciting field that is evolving each and every day. We have seen progress in applications, better algorithms and a better theoretical understanding. There is still a huge amount of interesting potential applications where Deep Learning can prove to be beneficial. In addition, there are a lot of interesting open questions about the underlying theory that we only begin to understand. As a researcher, it is truly a pleasure to work in this thriving and exciting field!

%% file: chapters/chapter4.tex
\begin{figure}[h!]
    \centering
    \includegraphics[width = 0.4\textwidth]{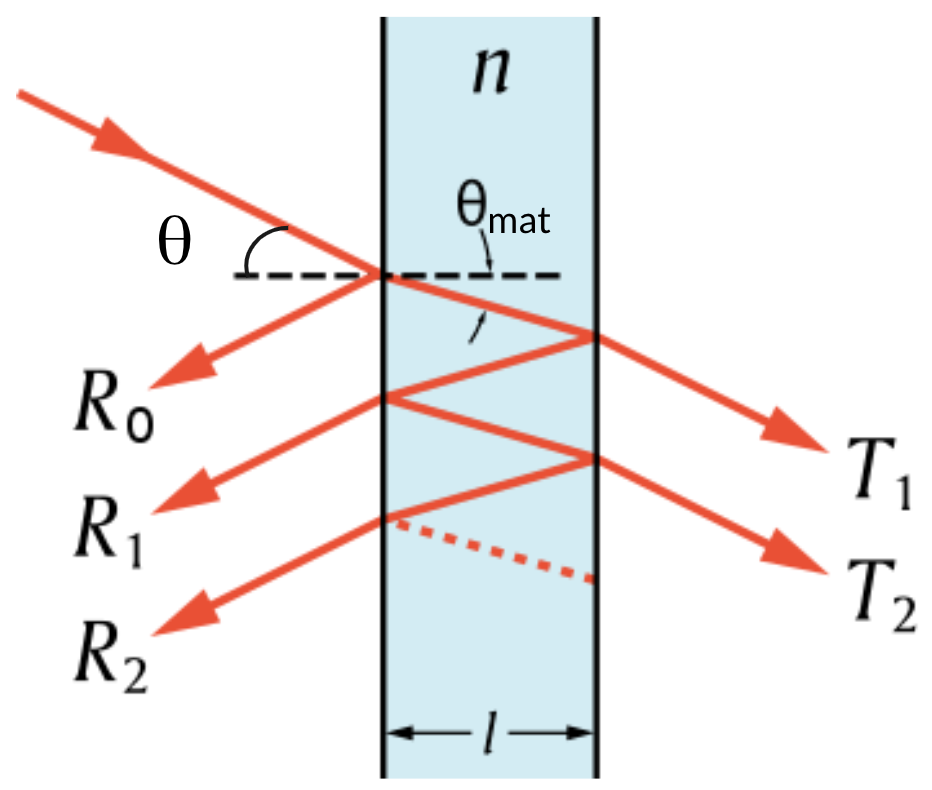}
    \caption{A Fabry-Pérot resonator is a dielectric medium width index of refraction $n$ and width $l$. Light falls in under an angle $\theta$ and is diffracted so that it travels at an angle $\theta_{mat}$ in the material. The multiple reflections of the light inside the resonator cause interference between different outgoing waves, leading to non-trivial reflection and transmission.}
    \label{fig: sketch Fabry-perot}
\end{figure}

\noindent A Fabry-Pérot resonator consists of a thin slice of a dielectric medium. When light falls in on the resonator, the light is reflected several times in the medium. This is shown in figure \ref{fig: sketch Fabry-perot}. The interesting property of this system, is that the total reflected and transmitted light are the sum of many partial waves. The light intensities coming in and going out are thus heavily influenced by the interference of these partial waves. This leads to a non-trivial dependence of the transmission and reflection on the wavelength of the incoming light and the design parameters of the resonator. Since reflection and transmission sum up to 1 in a lossless system, we only look at the transmission in this work. 

$\\$
There are 4 parameters that determine the interference of the partial waves: the wavelength of the incoming light $\lambda$, the incident angle $\theta$, the index of refraction $n$ and the width of the resonator $l$. The total transmission can then be determined to be

\begin{equation}
    T(\lambda, \theta, n, l) = \frac{1}{1 + F \text{sin}^2 (\frac{\delta}{2})},
    \label{eq: spectrum}
\end{equation}

\noindent with

\begin{equation}
    F(\theta, n) = \frac{4R}{(1-R)^2}, \hspace{1 cm} \delta(\lambda, \theta, n, l) = \frac{4 \pi n}{\lambda} l \text{cos} (\theta_{mat}). \hspace{0.6 cm} 
\end{equation}

\noindent In these equations, $F$ is the coefficient of finesse. This parameter is a function of the reflectance $R$ on the sides of the resonator. We assume that this reflectance is the same on both sides. We can get an expression for the reflectance based on the index of refraction of the material $n$ and the angle of incidence $\theta$. The full derivation is given in appendix \ref{Appendix: Fabry-Perot}. It is important to remember for now that $F$ depends on $\theta$ and $n$. 

$\\$
From equation \ref{eq: spectrum}, we can see that $F$ determines the minimal transmission $\frac{1}{1 + F}$, when the sine is equal to one. The transmission will oscillate between 1 and this minimal value. The other parameter $\delta$ is the phase difference between the different partial waves that are transmitted. At last, $\theta_{mat}$ is the angle inside the resonator. This angle can be determined from the index of refraction using Snell's law. The full derivation is given in appendix \ref{Appendix: Fabry-Perot}. 

$\\$
Let us now design a Fabry-Pérot resonator. We choose the material we want to use, fixing the index of refraction $n$. We then determine the width of the medium, fixing the parameter $l$. At last we determine the angle $\theta$ under which the light falls in. In terms of the wavelength $\lambda$, the transmission of this resonator is

\begin{equation}
    T(\lambda) = \frac{1}{1 + F \text{sin}^2 (\frac{\delta_0}{2 \lambda})},
    \label{eq: T for lambda, part 1}
\end{equation}

\noindent with

\begin{equation}
    F(\theta, n) = \frac{4R}{(1-R)^2}, \hspace{1 cm} \delta_0(\theta, n, l) = 4 \pi n l \text{cos} (\theta_{mat}). \hspace{0.6 cm} 
    \label{eq: T for lambda, part 2}
\end{equation}

\begin{equation}
    T(\lambda) = \frac{1}{1 + F \text{sin}^2 (\frac{\delta_0}{2 \lambda})},
    \label{eq: T for lambda, part 1}
\end{equation}

\begin{equation}
    F(\theta, n) = \frac{4R}{(1-R)^2}, \hspace{1 cm} \delta_0(\theta, n, l) = 4 \pi n l \text{cos} (\theta_{mat}). \hspace{0.6 cm} 
    \label{eq: T for lambda, part 2}
\end{equation}

\noindent This set of equations describes the transmission for different values of the wavelength $\lambda$ when the parameters $n$, $l$ and $\theta$ are fixed. We can do the same thing interchanging the role of $\lambda$ and $\theta$. We determine the transmission in function of the angle $\theta$ when $n$, $l$ and the wavelength $\lambda$ of the incoming light are fixed. This leads to

\begin{equation}
    T(\theta) = \frac{1}{1 + F(\theta) \text{sin}^2 (\frac{\delta_0 }{2}\text{cos}(\theta_{mat}))},
    \label{eq: T for theta, part 1}
\end{equation}

\noindent with

\begin{equation}
    F(\theta, n) = \frac{4R}{(1-R)^2}, \hspace{1 cm} \delta_0(\lambda, n, l) = \frac{4 \pi n l}{\lambda}. \hspace{0.6 cm} 
    \label{eq: T for theta, part 2}
\end{equation}

\noindent This set of equations is somewhat more complicated than the equations we had earlier for the wavelength. The reason for this is that $F$ does not depend on $\lambda$, but it is a function of $\theta$. Moreover, equation \ref{eq: T for theta, part 1} contains $\theta_{mat}$, which is in itself a function of $\theta$ and $n$, determined by Snell's law. The transmission $T(\theta)$ is therefore more involved than the transmission $T(\lambda)$. It will be interesting to see if the transmission in function of the angle is also harder to learn for the neural network than the transmission in function of the wavelength.

$\\$
This gives us every expression for the transmission of the Fabry-Pérot resonator that we use in this work. What we need to do next is describe the transmission  $T(\lambda)$ or $T(\theta)$ in terms of a neural network. We do this in three different ways in this thesis. What all of the descriptions have in common, is that they take three of the parameters $\lambda$, $\theta$, $n$ and $l$ or a combination thereof and map them to the transmission in function of the fourth parameter. Each output node of the network then corresponds to a value of the function $T(\lambda)$ or $T(\theta)$. Each section of this chapter covers one neural network. This is summarized below.

$\\$
\textbf{Overview -} In this chapter, we discuss 3 neural networks: 
\begin{itemize}
    \item Section \ref{sec: theta, n, l to T(lambda)}: $\theta$, $n$, $l$ $\rightarrow$ T($\lambda$). We predict the transmission in function of the wavelength. This allows us to know the transmission of a resonator for different colors of the incoming light.
    
    \item Section \ref{sec: lambda, n, l to T(theta)}: $\lambda$, $n$, $l$ $\rightarrow$ T($\theta$). We interchange the roles of $\lambda$ and $\theta$ to look at the transmission in function of the angle. This angular dependence of the transmission could be useful for optical computing. 
    
    \item Section \ref{sec: F, delta to T(lambda)}: $F$, $\delta_0$ $\rightarrow$ T($\lambda$). We again consider the transmission in function of the wavelength. We investigate if this transmission is easier to learn as a function of the parameters $F$ and $\delta_0$ that appeared in equations \ref{eq: T for lambda, part 1} and \ref{eq: T for lambda, part 2}, compared to the transmission as a function of the original parameters.
\end{itemize}

\section{Wavelength dependent transmission}
\label{sec: theta, n, l to T(lambda)}

\begin{figure}[h]
    \centering
    \includegraphics[width=\textwidth]{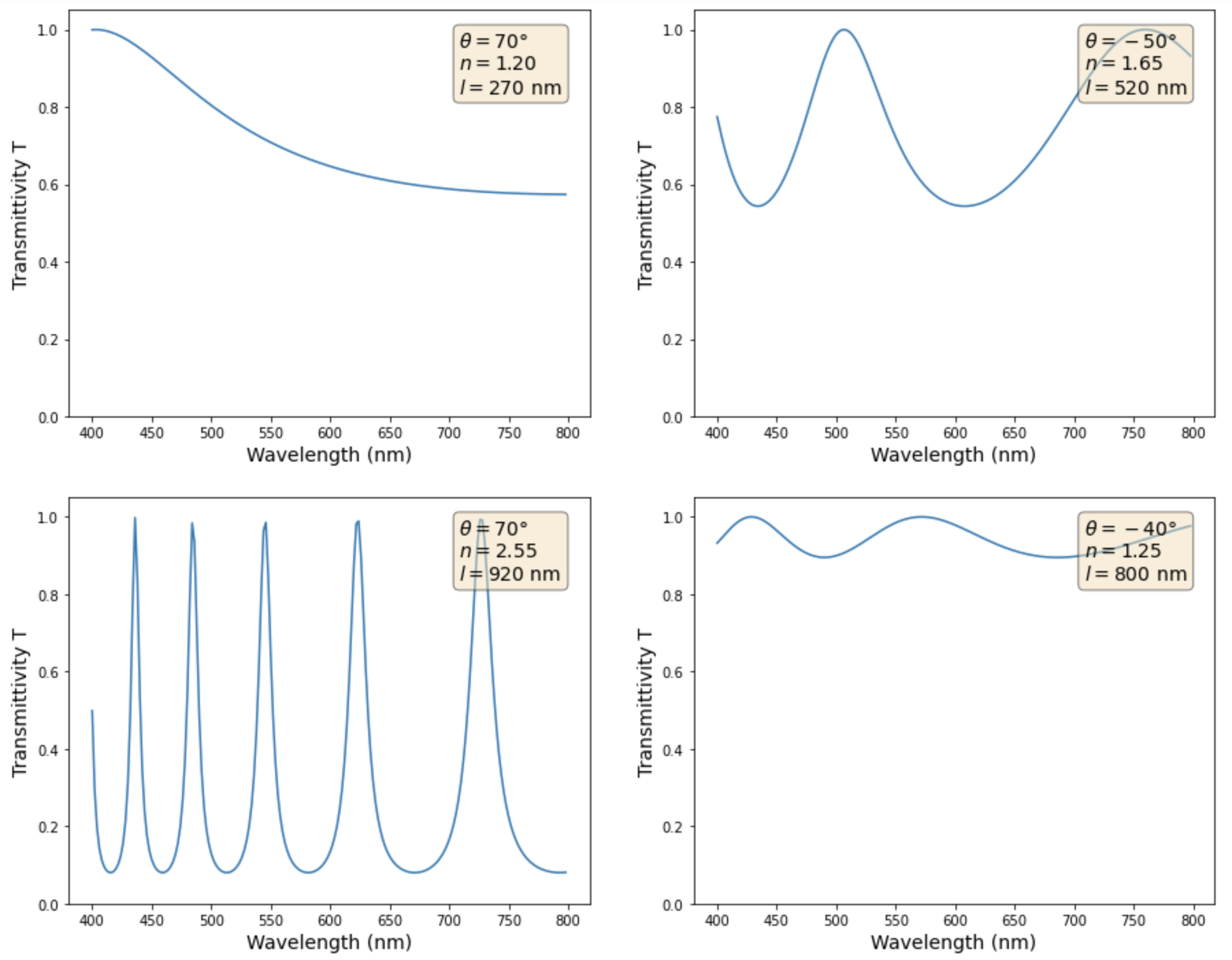}
    \caption{Transmission $T(\lambda)$ of the Fabry-Pérot resonator.}
    \label{fig: dataFP1}
\end{figure}

In this first section, we create a neural network that predicts the transmission $T(\lambda)$ for fixed parameters $\theta$, $n$ and $l$. We can motivate this as follows. Imagine a beam of white light falling in on the resonator. The white light contains equal amounts of every wavelength between 400 nm and 798 nm. We then ask ourselves, what is the color of the light transmitted by the resonator? In order to compute the transmitted color, we need to know the transmission for each wavelength $T(\lambda)$. This transmission is determined by the other parameters of the resonator, namely $\theta$, $n$ and $l$. Some transmissions $T(\lambda)$ in function of these parameters are shown in figure \ref{fig: dataFP1}. To predict the color of the transmitted light, we thus need to find a function that maps $\theta$, $n$ and $l$ to the transmission $T(\lambda)$.

\begin{figure}
    \centering
    \includegraphics[width=0.7\textwidth]{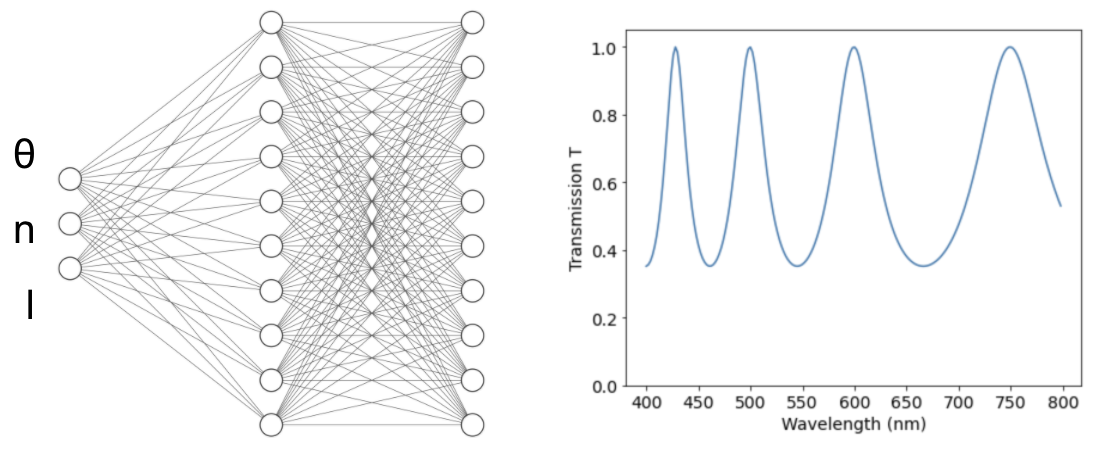}
    \caption{Overview of the neural network we created. The number of nodes is not representative.}
    \label{fig: overview neural network}
\end{figure}

$\\$
We approximate this function by a neural network. The network has three input nodes representing $\theta$, $n$ and $l$. To have the transmission as output, we represent $T(\lambda)$ as a set of 200 discrete points. Each point represents the transmission for a value of the wavelength, where $\lambda$ is chosen between 400 nm and 798 nm in steps of 2 nm. In this way, the transmission can be represented by 200 output nodes, with each node corresponding to the transmission for a certain wavelength. This is shown in figure \ref{fig: overview neural network}. We then have a neural network that can learn the transmission of the Fabry-Pérot resonator.

$\\$
We trained the neural network in a supervised way with $\theta$, $n$ and $l$ as the input data $\bf{x}$ and $T(\lambda)$ as labels $\bf{y}$. The first step is to choose the parameter values we want to include in our data. This is a very important step. We want a broad range of parameter values that we sample sufficiently dense, such that the neural network can meaningfully interpolate between the data it has seen. We constructed our data set with the following parameter values

\begin{equation}
\begin{aligned}
    \nonumber
    \theta &= 0\degree, \pm 15 \degree, \pm 30 \degree, \pm 40\degree, \pm 50\degree, \pm 60\degree, \pm 70\degree, \\
    n &\in [1.05, 3.50], \ \text{stepsize:} \ 0.05, \\
    l &\in [100 \ \text{nm}, 1000 \ \text{nm}], \ \text{stepsize:} \ 10 \ \text{nm}.
\end{aligned}
\end{equation}

\noindent The input data needs to be normalized such that each parameter has a comparable influence. Otherwise, the large values of $l$ would have a much larger effect on the network than the index of refraction $n$. There is however no physical reason to believe that $n$ would be more important than $l$. Therefore, we normalize all parameters to the range $[0, 1]$. For each of these parameters, we can analytically compute the transmission $T(\lambda)$ using equations \ref{eq: T for lambda, part 1} and \ref{eq: T for lambda, part 2}. In this way, we obtain 59150 data samples.

$\\$
As is standard practice in Machine learning, we divide this data into a training set, a validation set and a test set. The training set is the only data from which the neural network learns directly. The network uses this data set to perform gradient descent and adjust its parameters. The validation set is used during training mainly for two reasons, to prevent overfitting and to tune the hyperparameters of the model. The neural network does not learn from the validation set, but instead computes the loss function on this data set during training. 

\begin{figure}
    \centering
    \includegraphics[width=0.7\textwidth]{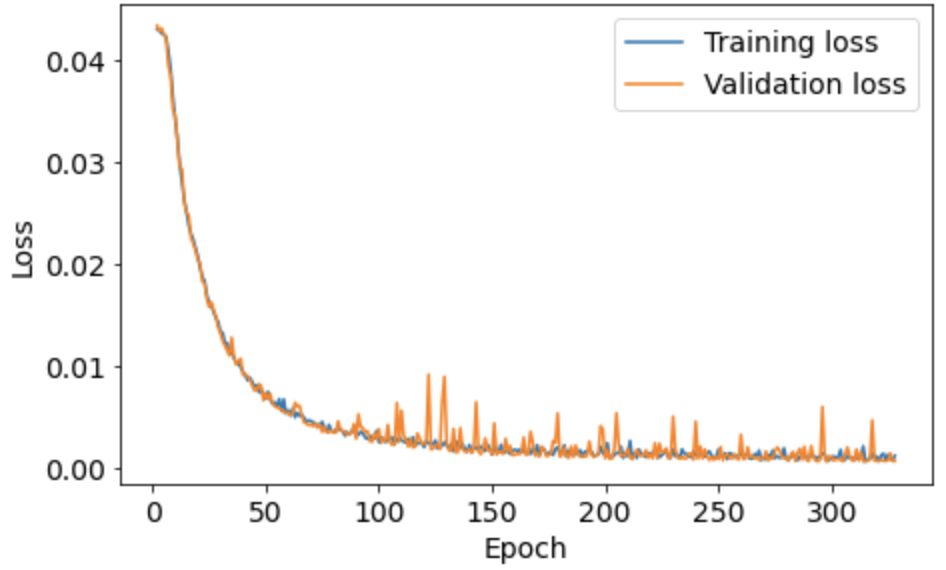}
    \caption{Loss function during training. Training loss and validation loss overlap, so there is no overfitting.}
    \label{fig: loss ReLU}
\end{figure}

$\\$
To prevent overfitting, we can compare the loss on the training set with the loss on the validation set. If overfitting occurs, we would observe that the loss on the training set decreases, but the loss on the validation set increases. It would mean that the neural network is memorizing the training set, but therefore not learning anything meaningful about the unseen validation set. In our experiments, we did not observe overfitting. A typical plot of the loss function during training is shown in figure \ref{fig: loss ReLU}. This plot shows large peaks for the validation loss. This is a consequence of the stochastic nature of the gradient descent. A peak occurs every time a step was taken in the wrong direction. 

$\\$
The second reason we need this validation set is to tune \textbf{hyperparameters}. These are parameters like the number of hidden layers, the number of nodes in each layer or the parameters specifying gradient descent. The hyperparameters determine what the neural network looks like. The strategy to find optimal hyperparameters for a given problem is to train multiple neural networks with different hyperparameters. We then select the neural network with the best performance on the validation set as our final model. Once we have a final model, we evaluate the performance on the test set and this is the final performance of the model. One could ask why an evaluation on the test set is different from an evaluation on the validation set, since both sets are unseen by the neural network. The reason for this is that the validation set has influenced the neural network by selecting the best hyperparameters. When assessing the performance on new data, this data has to be completely independent from the way in which the neural network was created. Therefore, our final performance needs to be computed on the test set. 

$\\$
We randomized our data and then split into a first set of 53235 data samples and a second set of 5915 data samples. We pick this second set as our test set. The first set is split into training and validation set every time we train a new neural network. This is done automatically by using the Deep Learning library Keras. This Python library gives us a whole set of functions and algorithms to easily make and train neural networks \cite{chollet2015keras}. The split in training set and validation set is different every time we train a model. We chose 15\% of the data as validation set and the other 85\% as training set. In the end, there are 45250 data samples in the training set, 7985 data samples in the validation set and 5915 data samples in the test set.

$\\$
Now that we have set up our data, we can start training. Let us describe the hyperparameters we need to tune for the training. We update the weights $\bf{W}$ and biases $\bf{b}$ of the neural network by gradient descent. An important hyperparameter for gradient descent is the \textbf{learning rate} $\eta$. This parameter determines the step size of the parameter update in every iteration, we have

\begin{equation}
    \nonumber
    \mathbf{w}_{i+1} = \mathbf{w}_i - \eta \nabla_{\mathbf{w_i}} L(f_{\mathbf{w_i}}(\mathbf{x}), \mathbf{y}),
\end{equation}
where we take the gradient of the loss function with respect to the parameters we want to update. The notation $\bf{w}$ refers to both the weights $\bf{W}$ and the biases $\bf{b}$. There are different ways to choose the learning rate $\eta$ and the update rule in general. These different ways of doing gradient descent are called \textbf{optimization algorithms}. A great review on optimization algorithms for Deep Learning is given by Ruder \cite{Ruder2016}. We have chosen one of the most popular optimization algorithms called Adam. We used a learning rate $\eta = 0.001$ as is standard in the Keras library. As we come closer to a solution during training, the learning rate is decreased by the Adam optimization algorithm.

$\\$
The next hyperparameter we need to determine is the \textbf{batch size}. Computing gradients of the loss function for all data samples at once would require a lot of memory. Every data sample in the training set should be loaded into RAM at every iteration. The other extreme is to compute the loss and its gradient for every data sample separately. This would lead to many iterations before the network has seen every data sample and a large computation time. The middle ground between these two is called \textbf{Stochastic Gradient Descent (SGD)}. In every iteration, the training set is randomly divided into a number of batches. Each batch contains a fixed number of data samples called the batch size. 

$\\$
The loss function and gradients are then computed for one batch in every iteration. When enough iterations are performed such that all batches in the training data are seen once, we say that an \textbf{epoch} has completed. After an epoch is completed, the training data is again randomly divided into batches and the training continues. 

$\\$
A curious thing about stochastic gradient descent is that it leads to better results compared to using the whole test set. This also became clear in our own experiments. To assess the performance of the networks, we look at the mean absolute error (MAE) on every point of the transmission, given by

\begin{equation}
    \text{MAE} = \frac{1}{200} \vert f_{\mathbf{w}}(\mathbf{x}) - \bf{y} \vert.
\end{equation}

\begin{figure}
    \centering
    \includegraphics[width=0.5\textwidth]{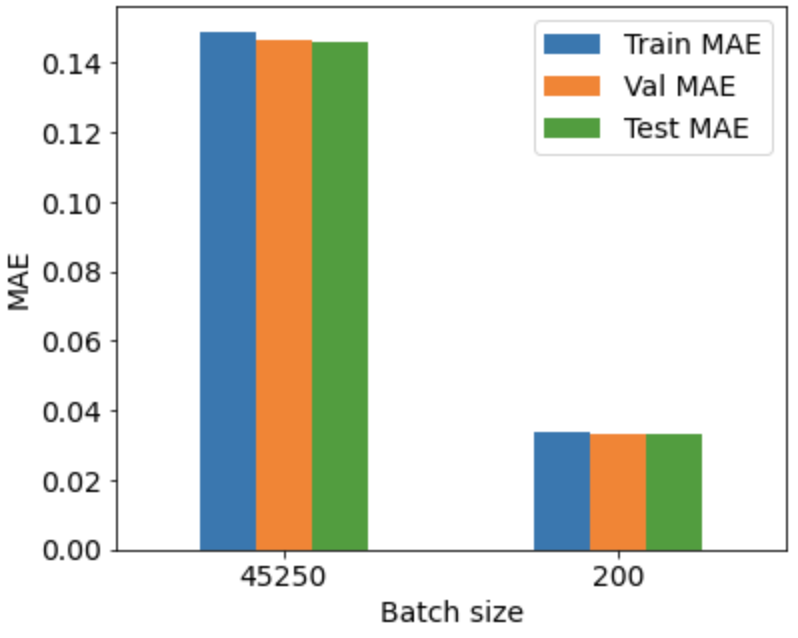}
    \caption{Mean absolute error (MAE) of two neural networks with different batch size. All other hyperparameters were identical.}
    \label{fig: performance SGD}
\end{figure}

\noindent In this equation, $\bf{y}$ is the true transmission, $f_{\mathbf{w}}(\mathbf{x})$ is the prediction of the network and $\mathbf{w}$ are the weights and biases of the network. We divide by 200 since we have 200 output nodes. The MAE for different batch sizes is shown in figure \ref{fig: performance SGD}. The two networks have 4 hidden layers, 100 nodes in each layer and the ReLU activation function. They are optimized using Adam. 

$\\$
We also tried a batch size of 1, but this took several minutes to complete just one epoch. It would have taken us several hours to complete the training, compared to the 10 minutes it took on average to train with SGD. We therefore did not complete the training with batch size 1. It remains however difficult to determine the optimal batch size. Like other hyperparameters, it has to be determined by trial and error. Ultimately, we settled for a batch size of 200.

$\\$
When we are training, we need to determine a point at which we stop. We can do this simply by specifying the number of epochs we want to train for, but we can also do this in a more clever way. Ideally, we want the neural network to stop learning once it has learned every useful pattern in the training set. The only way to decrease the training loss further would be to start memorizing the data and this is not what we want. To determine when learning is complete, we can look at the loss of the validation set. Once this loss does not decrease any further, we can stop training. Since the process is stochastic however, it could be that the validation loss increases in an iteration because the parameters were updated in the wrong direction. Nonetheless, this does not mean that the validation does not decrease in the following iterations. 

$\\$
To account for this, we conclude that the validation loss does not decrease further when it has not decreased for several epochs. The number of epochs to wait for a better validation loss is called the \textbf{patience} hyperparameter. We used a patience of 30 epochs. This got us close to convergence, while simultaneously keeping the training time limited to about 5-15 minutes, depending on the other hyperparameters. In the end, we can summarize the hyperparameters for the training of the neural network in table \ref{table: hyperparameters training}. We use these hyperparameters for every network in this chapter.

\begin{table}[h]
\centering
{\rowcolors{2}{gray!20!}{gray!40!}
    \begin{tabular}{|p{2.8cm}|p{2.8cm}|}
        \hline
        Hyperparameter & Value \\
        \hline
        Optimizer & Adam \\
        Learning rate $\eta$ & 0.001 \\
        Batch size & 200 \\
        Patience & 30 \\
        \hline
    \end{tabular}
}
\caption{Hyperparameters for the training.}
\label{table: hyperparameters training}
\end{table}

\noindent Now that we have discussed the optimal hyperparameters for the training of the neural network, we can determine the optimal architecture for this problem. We gained inspiration from the paper of Peurifoy et al. \cite{Peurifoy2017}. This paper created a neural network to predict the scattering of nanoparticles, based on the width of the layers of the nanoparticles. Since this problem is quite similar to ours, we expect their architecture to also work well for the Fabry-Pérot resonator. 

\begin{figure}
    \centering
     \begin{subfigure}[b]{0.45\textwidth}
         \centering
         \includegraphics[width=\textwidth]{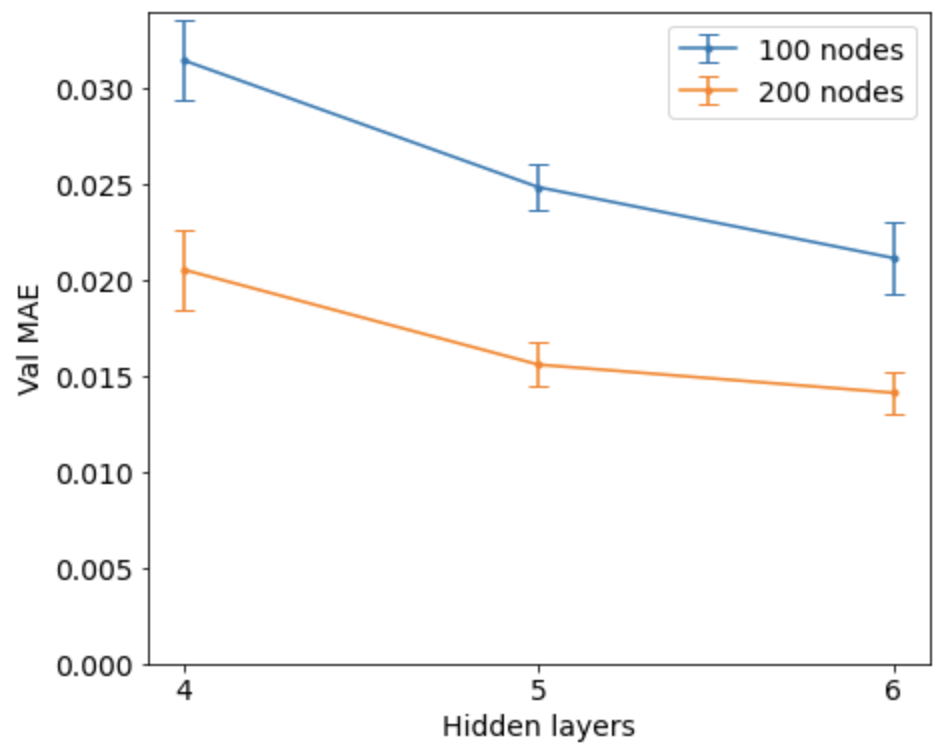}
         \caption{}
         \label{fig: hyperparameters FP1 nodes}
     \end{subfigure}
     \hfill
     \begin{subfigure}[b]{0.45\textwidth}
         \centering
         \includegraphics[width=\textwidth]{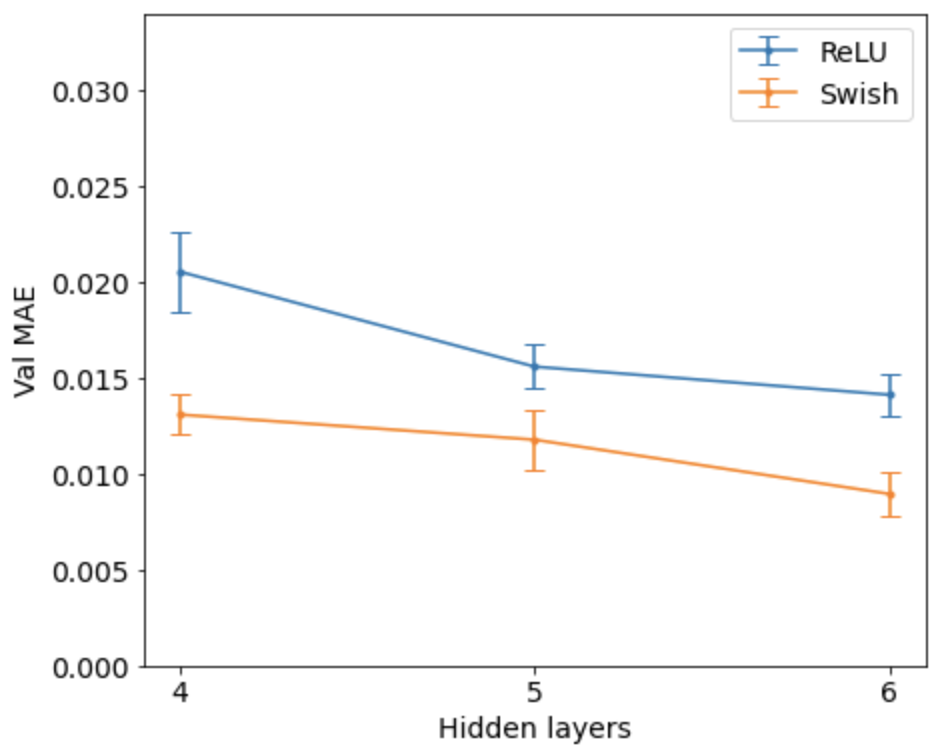}
         \caption{}
         \label{fig: hyperparameters FP1 activation}
     \end{subfigure}
    \caption{The mean absolute error (MAE) on the validation set for different architectures. a) Different numbers of hidden layers and number of nodes in each layer are compared. b) Different numbers of hidden layers and activation functions are compared.}
    \label{fig: hyperparameters FP1}
\end{figure}

$\\$
The architectures we tested contain 4, 5 or 6 hidden layers with 100 or 200 nodes in each of the layers. The activation function for these networks is ReLU. To assess the performance, we look at the mean absolute error (MAE) on the validation set. When a neural network is trained, the weights and biases are initialized randomly. This causes an uncertainty on the performance of the neural network. We therefore trained 5 neural networks for each architecture. We then computed the mean and error bars of the MAE on the validation set for these 5 networks. The results are shown in figure \ref{fig: hyperparameters FP1 nodes}.

$\\$
We see that more hidden layers and more hidden nodes lead to a lower MAE. This is what we would expect. Also note that the error bars are relevant, since they are of the same order of magnitude as the difference in MAE between different architectures.

$\\$
The most popular activation function for deep networks nowadays is ReLU. This is also the one used by Peurifoy et al. Recently however, a study by Google Brain researched the best possible activation function for deep neural networks \cite{ramach2017}. The best possible activation they discovered in their research is the Swish, given by

\begin{equation}
    \nonumber
    \begin{aligned}
        \sigma_{swish}(x) &= x \cdot \sigma_{sigmoid}(x), \\
        &= \frac{x}{1 + e^{-x}}.
    \end{aligned}
\end{equation}

\noindent We used the ReLU function in the experiments comparing different architectures. We might be able to improve our results using Swish instead of ReLU. We investigated this idea for the networks with 200 nodes in each hidden layer.  Results are shown in \ref{fig: hyperparameters FP1 activation}. Note that the upper line for ReLU shows the performance of the same networks as the networks with 200 nodes in figure \ref{fig: hyperparameters FP1 nodes}. Our experiments show that choosing the Swish activation can decrease the MAE even further for this problem. 

\begin{figure}
    \centering
    \includegraphics[width=\textwidth]{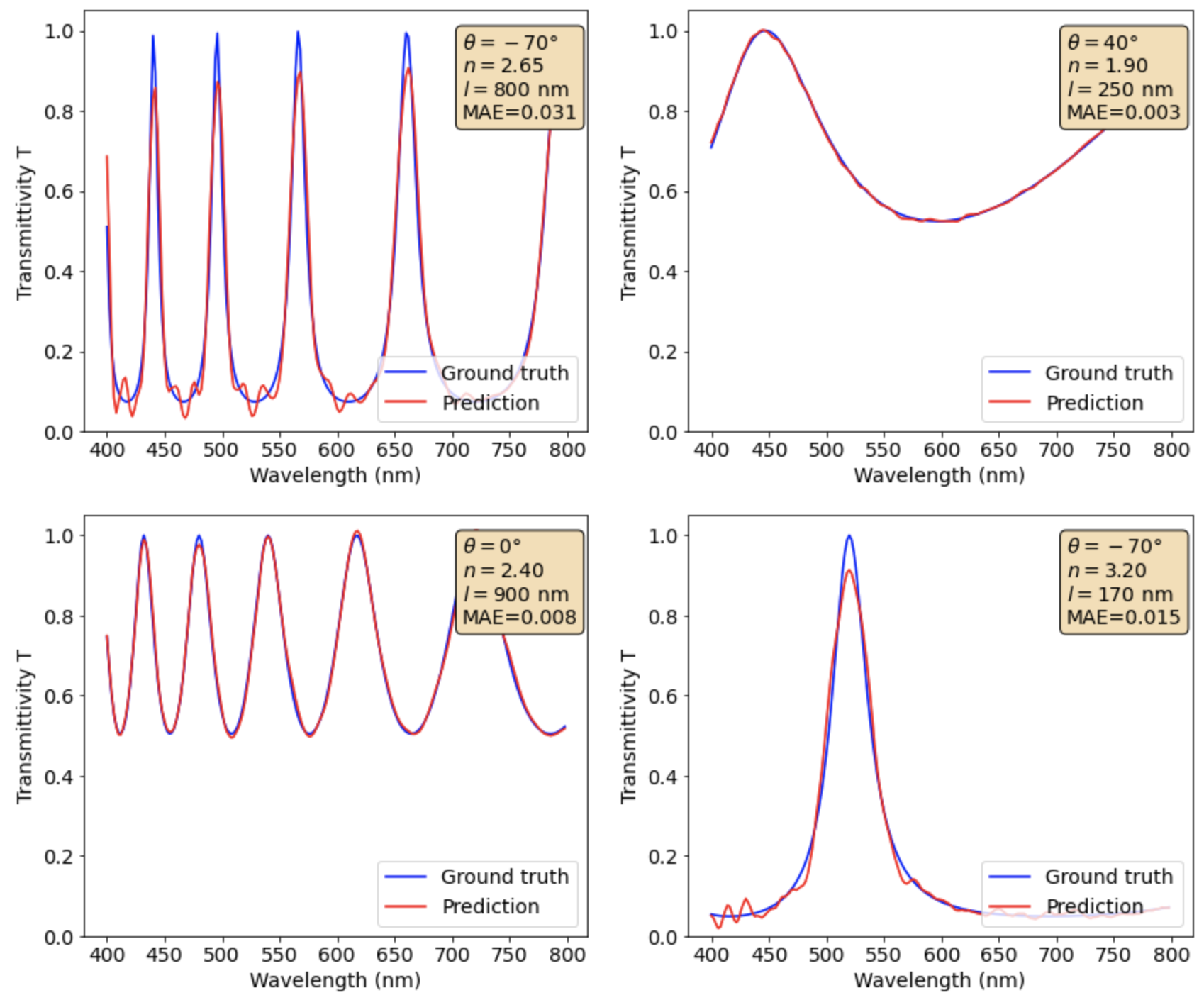}
    \caption{Predictions of transmissions $T(\lambda)$ for different design parameters in the test set. The blue lines are the analytically computed transmissions and the red lines are the predictions. The design parameters are shown in the box in the upper right. This box also shows the mean absolute error (MAE) between the prediction and the ground truth.}
    \label{fig: predictions FP1}
\end{figure}

$\\$
In the end, the best neural network we created has 6 hidden layers, 200 nodes in each hidden layer and the Swish activation function. The hyperparameters for training are summarized in the Supplemental Materials. This network obtains a test MAE of $0.898 \pm 0.120\%$, averaged over 5 neural networks with random weight initializations. The predictions of the best of these 5 network on several transmissions $T(\lambda)$ from the test set are shown in figure \ref{fig: predictions FP1}. 

$\\$
In each plot, the input parameters of the network are shown in the upper right. The plot shows the ground truth in blue. This is the transmission calculated by equation \ref{eq: spectrum}. The prediction of the network is shown in red. The MAE between the ground truth and the predictions is also given in the box in the upper right. The four plots show good predictions. The transmissions with sharp peaks are somewhat harder to predict. This can be expected, since we sample the transmission at 200 points. This means that sharp peaks suffer from the sampling. This leads us to think that a denser sampling of the transmission should lead to better predictions for transmissions with sharper peaks. We did however not test this idea. 

\section{Angle dependent transmission}
\label{sec: lambda, n, l to T(theta)}

In the second set of experiments, we interchange the role of $\lambda$ and $\theta$. The transmission $T(\theta)$ that we obtain in this way is interesting for optical computing. This field tries to do computation with classical wave optics. It is inspired by the properties of a simple lens. When an image is diffracted by a lens, the Fourier transform of this image appears at the focal point. This allows us to do computations in the Fourier space, by placing optical elements at a focal distance behind a lens. 

$\\$
A great benefit of optical computing is that it computes in parallel. This parallel computing is part of what allowed Deep Learning to have the fast computation it needs. A great review of the history of optical computing is found in \cite{Ambs2010}. By selecting components of the light based on $\theta$, we could essentially do the same as a lens, but now on a much shorter distance. 

$\\$
The three input nodes of the neural network are $\lambda$, $n$ and $l$. There are 179 output nodes, corresponding to the transmission $T(\theta)$ for every degree between $-89\degree$ and $89\degree$. We then have to choose the values of the input parameters. These are

\begin{equation}
\begin{aligned}
    \lambda &\in [400 \ \text{nm}, 452 \ \text{nm}], \ \text{stepsize:} \ 4 \ \text{nm}, \\
    n &\in [1.05, 3.50], \ \text{stepsize:} \ 0.05, \\
    l &\in [100 \ \text{nm}, 1000 \ \text{nm}], \ \text{stepsize:} \ 10 \ \text{nm}. 
\end{aligned}
\end{equation}

\noindent We chose the same values for $n$ and $l$ as in the previous problem. We have limited the range for $\lambda$ so we do won't have too much data samples. In this set of experiments, we normalized the parameters to the range $[-1, 1]$ which is more standard in Machine Learning. Nonetheless, we observed no significant difference in performance compared to the normalization to the range $[0, 1]$, like we did in the previous section. We end up with 63700 normalized data samples. 

$\\$
We have split this data into 48730 data samples in the training set, 8600 data samples in the validation set and 6370 data samples in the test set. The test set is the same for every network, while the split in training and validation set was performed randomly every time we trained a new network. We used the same hyperparameters for training as in the previous section, summarized in table \ref{table: hyperparameters training}. 

\begin{figure}[h]
    \centering
    \includegraphics[width=0.5\textwidth]{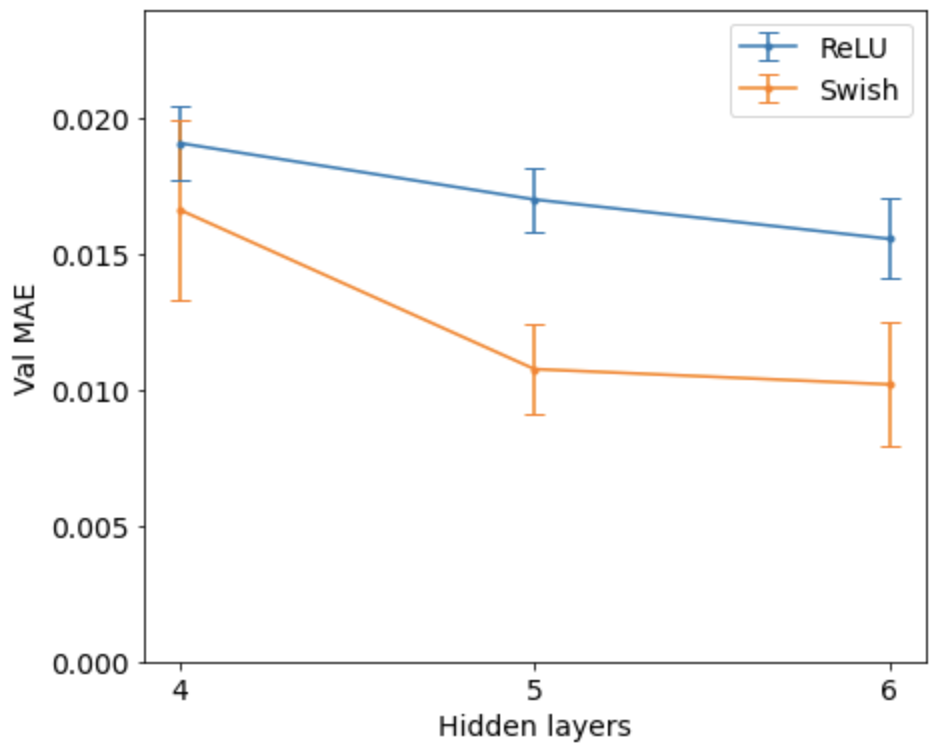}
    \caption{The mean absolute error (MAE) on the validation set for different architectures.}
    \label{fig: hyperparameters FP2}
\end{figure}

$\\$
To obtain the optimal architecture for this problem, we again trained several neural networks with different architectures. We made changes in the architecture to the activation function and to the number of hidden layers. The results are plotted in figure \ref{fig: hyperparameters FP2}. We see that the Swish activation function gives better MAE on the validation set than ReLU. Deeper networks also give better performance.

\begin{figure}[h!]
    \centering
    \includegraphics[width=\textwidth]{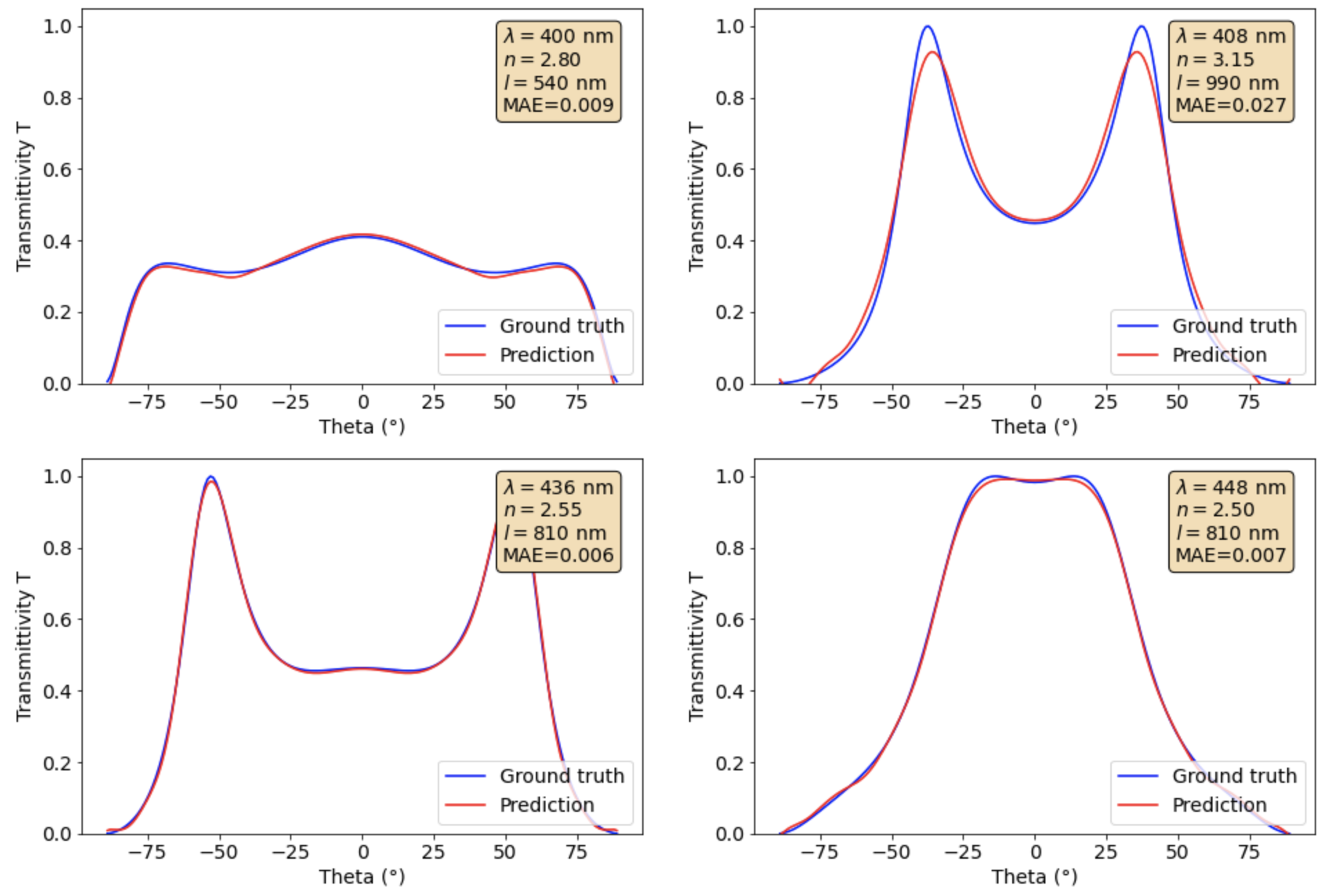}
    \caption{Predictions of transmissions $T(\theta)$ for different design parameters in the test set. The blue lines are the analytically computed transmissions and the red lines are the predictions. The design parameters are shown in the box in the upper right. This box also shows the mean absolute error (MAE) between the prediction and the ground truth.}
    \label{fig: predictions FP2}
\end{figure}

$\\$
The best neural network we created to predict the transmission in function of the angle $\theta$ has 6 hidden layers with 200 nodes in each layer and the Swish activation function. We used the same hyperparameters for training as in section one, given in table \ref{table: hyperparameters training}. The network has a test MAE of $1.029 \pm 0.226 \%$, averaged over 5 neural networks. Predictions of the best of these 5 networks on some transmissions $T(\theta)$ in test set are shown in figure \ref{fig: predictions FP2}. We see that the predictions agree well with the ground truth.

\section{Simplified wavelength dependent transmission}
\label{sec: F, delta to T(lambda)}

The third set-up we investigate is based on our knowledge of the analytical expression of the Fabry-Pérot transmission given in equation \ref{eq: T for lambda, part 1}. For convenience, we repeat this expression here

\begin{equation}
    \nonumber
    T(\lambda, \theta, n, l) = \frac{1}{1 + F \text{sin}^2 (\frac{\delta_0}{2 \lambda})}.
\end{equation}

\noindent The parameters $F$ and $\delta_0$ are themselves functions of $\theta$, $n$ and $l$. Working with these parameters makes the expression for the transmission a lot more simple. We now want to test if it is also easier for a neural network to learn the transmission in function of these two parameters. 

$\\$
To test this idea, we make a neural network with two input parameters $F$ and $\delta_0$. The output is $T(\lambda)$, where the output is given as 200 output nodes as before. We take the same parameter ranges for $\theta$, $n$ and $l$ as the first section of this chapter. This leads to parameters $F \in [0.10, 23.8]$ and $\delta_0 \in [588, 42367]$. These values are normalized to be in the interval $[-1, 1]$. This gives us 45250 samples in the training set, 7985 samples in the validation set and 5915 samples in the test set for a total of 63700 samples.

$\\$
We train with the same hyperparameters for learning as in table \ref{table: hyperparameters training}, but the patience is now 50 instead of 30. We observed that the training would stop too fast otherwise. 

\begin{figure}[h!]
     \centering
     \begin{subfigure}[b]{0.45\textwidth}
         \centering
         \includegraphics[width=\textwidth]{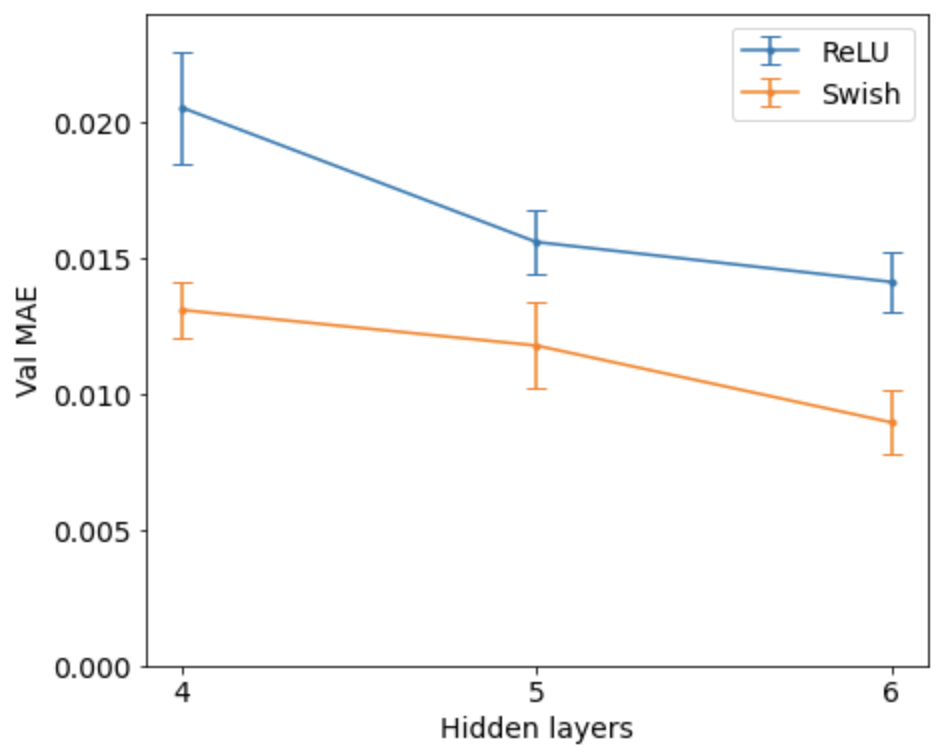}
         \label{fig: hyperparameters FP1 vs FP3, FP1}
     \end{subfigure}
     \hfill
     \begin{subfigure}[b]{0.45\textwidth}
         \centering
         \includegraphics[width=\textwidth]{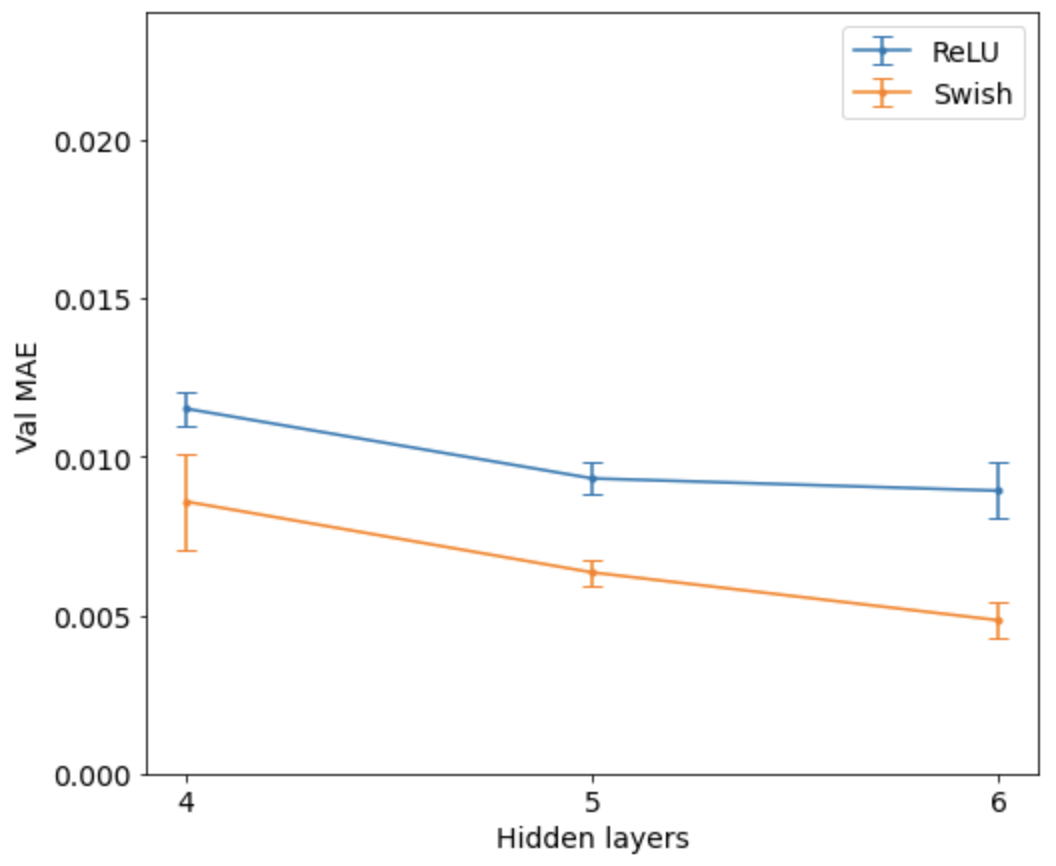}
         \label{fig: hyperparameters FP1 vs FP3, FP3}
     \end{subfigure}
    
    \caption{The mean absolute error (MAE) on the validation set for two different neural networks. a) The networks created in section one, where the original parameters $\theta$, $n$ and $l$ were used as input. b) The networks created in this section where $F$ and $\delta_0$ are used as input.}
    \label{fig: hyperparameters FP3}
\end{figure}

\noindent We tested different architectures for this problem. The hyperparameters we investigated are the number of hidden layers and the activation function. As in the previous sections, we tested networks with 4, 5 and 6 hidden layers and either ReLU or Swish as activation function. In this set of experiments, there are only two input parameters instead of three. The output remains the same. It is interesting to ask how this will effect the optimal hyperparameters. As in earlier sections, we obtained the error bars by training 5 neural networks with different random initialisation.  The results are plotted in figure \ref{fig: hyperparameters FP3}. We observe the same trend as in the other sections that deeper networks work better and that the Swish activation function is preferred over ReLU. 

$\\$
We also see that a neural network that used the simplified parameters $F$ and $\delta_0$ as input leads to better predictions than a network with the original design parameters $\theta$, $n$ and $l$ as input. The difference is also quite significant. This shows that knowing the true degrees of freedom in an optical system is of great use to train neural networks. In this case, we obtained these parameters looking at the analytical expression. For more general optical problems where we do not have an analytical expression, we need other methods to find the degrees of freedom. This is what we explore in the next chapter.

\begin{figure}[h]
    \centering
    \includegraphics[width=\textwidth]{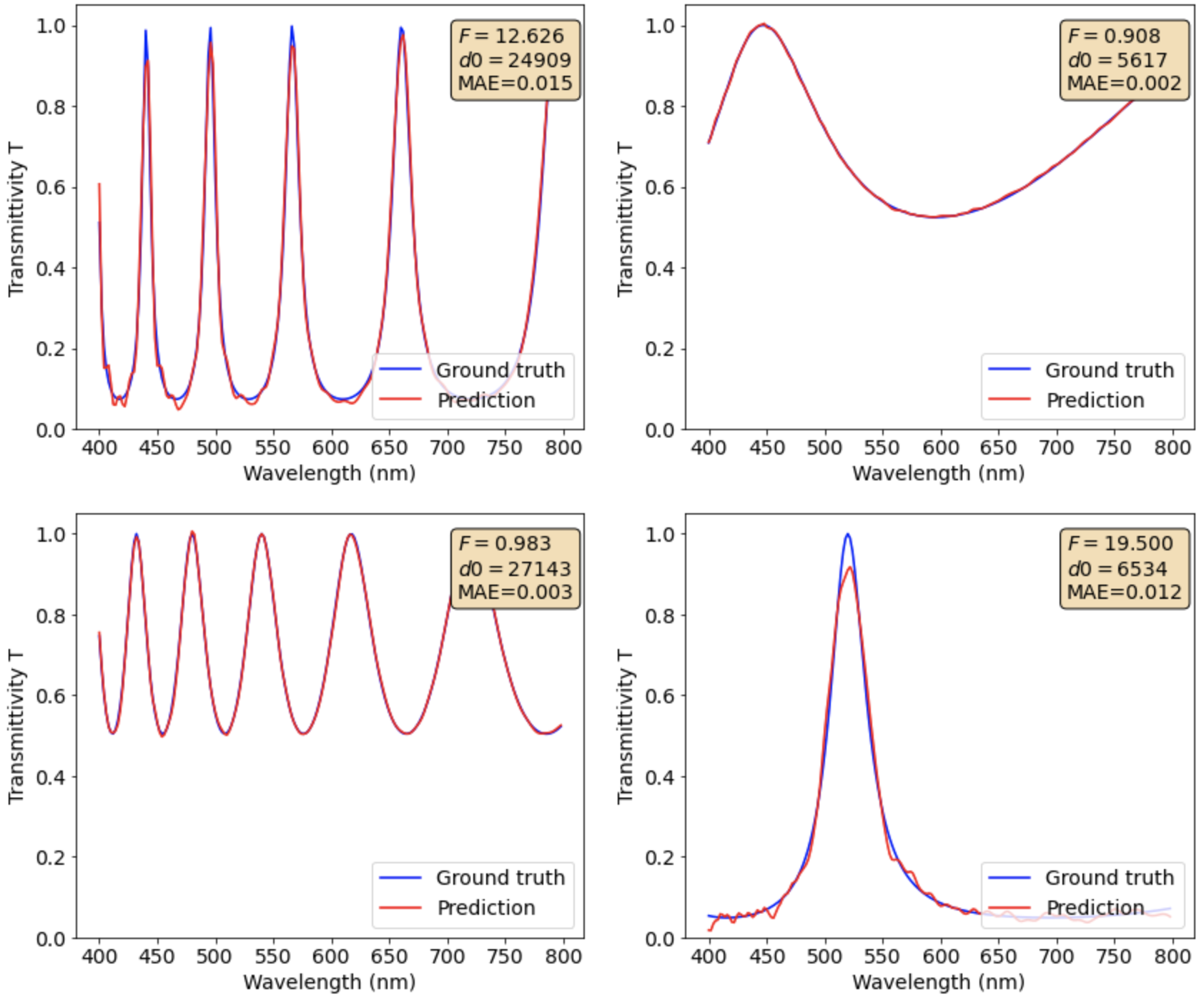}
    \caption{Predictions of transmissions $T(\lambda)$ for different design parameters $F$ and $\delta_0$ in the test set. The blue lines are the analytically computed transmissions and the red lines are the predictions. The design parameters are shown in the box in the upper right. This box also shows the mean absolute error (MAE) between the prediction and the ground truth.}
    \label{fig: predictions FP3}
\end{figure}

\noindent The best neural network we created to predict the transmission $T(\lambda)$ starting from $F$ and $\delta_0$ has 6 hidden layers with 200 nodes in each layer and the Swish activation function. We used the same hyperparameters for training as in section one, given in table \ref{table: hyperparameters training}. This network obtained a test MAE of $0.485 \pm 0.060 \%$, averaged over 5 networks. Some predictions of the best of these 5 networks on the test set are shown in figure \ref{fig: predictions FP3}. These predictions are closer to the ground truth than those in figure \ref{fig: predictions FP1} for the neural network trained on $\theta$, $n$ and $l$.

\section*{Conclusion}

\noindent Finally, we summarize the performance of the best networks for each of the 3 problems in this chapter. We look at the test MAE averaged over 5 networks with the best architecture. In all cases, the best architecture has 6 hidden layers, 200 nodes in each layer and the Swish activation function. The results are given in table \ref{table: test MAE prediction networks}. We conclude that in each case, we were able to create a neural network that is able to make very good predictions for the transmission of the Fabry-Pérot resonator. 

$\\$
We observe that it was easier to learn the transmission $T(\lambda)$ than to learn the transmission $T(\theta)$. This is what we expected based on the analytical formulas for these spectra in equations \ref{eq: T for lambda, part 1} and \ref{eq: T for theta, part 1}. We also note that the transmission can be better predicted with $F$ and $\delta_0$ as input, compared to using the original design parameters $\theta$, $n$ and $l$ as input. From the analytical expression in equation \ref{eq: spectrum} this makes sense. The function we need to learn for $F$ and $\delta_0$ is a lot simpler than the full function for $\theta$, $n$ and $l$. This indicates that functions that appear easier analytically, are also easier to learn by a neural network.

$\\$
\begin{table}[h!]
\centering
{\rowcolors{2}{gray!20!}{gray!40!}
    \begin{tabular}{|p{2.8cm}|p{2.8cm}|}
        \hline
        Network & Test MAE \\
        \hline
        $\theta, n, l \rightarrow T(\lambda)$ & $0.898 \pm 0.120 \%$ \\
        $\lambda, n, l \rightarrow T(\theta)$ & $1.029 \pm 0.226 \%$ \\
        $F, \delta_0 \rightarrow T(\lambda)$ & $0.485 \pm 0.060 \%$ \\
        \hline
    \end{tabular}
}
\caption{Performance of the best neural networks in each section.}
\label{table: test MAE prediction networks}
\end{table}

%% file: chapters/chapter5.tex
\noindent In the previous section, it became clear that a neural network performed better when trained on $F$ and $\delta_0$ compared to when it was trained on the original design parameters $\theta$, $n$ and $l$. The reason for this, is that $F$ and $\delta_0$ are the actual degrees of freedom of the resonator. We knew this because we had an analytical solution of the problem. In more general problems, such an analytical solution does not always exist. In these situations, Maxwell's equations have to be solved, quite often computationally. It is not clear how the parameters specifying the problem influence one another in these simulations. It would therefore be of great use to find a way to uncover the true degrees of freedom in an optical system.

$\\$
Deep Learning could provide a way to do this. In a recent article, Deep Learning was used to discover physical concepts like heliocentrism \cite{Iten2020}. The authors of the paper used a $\beta$-VAE as discussed in chapter \ref{chap 3: deep learning} to encode parameters of physical interest in the latent space. To discover heliocentrism, the neural network was shown a time series of the angles of Mars and the Sun as seen from Earth. The network then needed to compress these two angles into 2 latent variables to predict future positions of Mars and the Sun as simply as possible. The network stored the angles of Earth and Mars as seen from the Sun in the latent representation. Since the Sun is in the middle of the solar system, these angles have the simplest time evolution. 

$\\$
In another problem, the network was given a time series of the classical harmonic oscillator. Based on the previous positions of a mass attached to a spring, the network was asked to predict future positions. The network learned that it only needed the mass $m$ and the spring constant $k$ to correctly predict the motion of the harmonic oscillator. This shows that the $\beta$-VAE is able to discover the degrees of freedom in a physical system.

$\\$
There was also a recent application of autoencoders to inverse design in Nanophotonics by Kiarashinejad et al. \cite{Kiarashinejad2020}. They used the autoencoder as dimensionality reduction for both the design and the optical response. This led to a reduction in computation. Moreover, it gave them physical insights into how different design parameters influenced the optical response.

$\\$
Inspired by this promising work, we tried to discover the degrees of freedom for Fabry-Pérot. We want the network to learn unsupervised from the transmissions $T(\lambda)$ alone. By asking to reconstruct the transmissions after going through a bottleneck, the network will be forced to learn $F$ and $\delta_0$. 

\section{Proof of concept}

\begin{figure}[h!]
    \centering
    \includegraphics[width=0.8\textwidth]{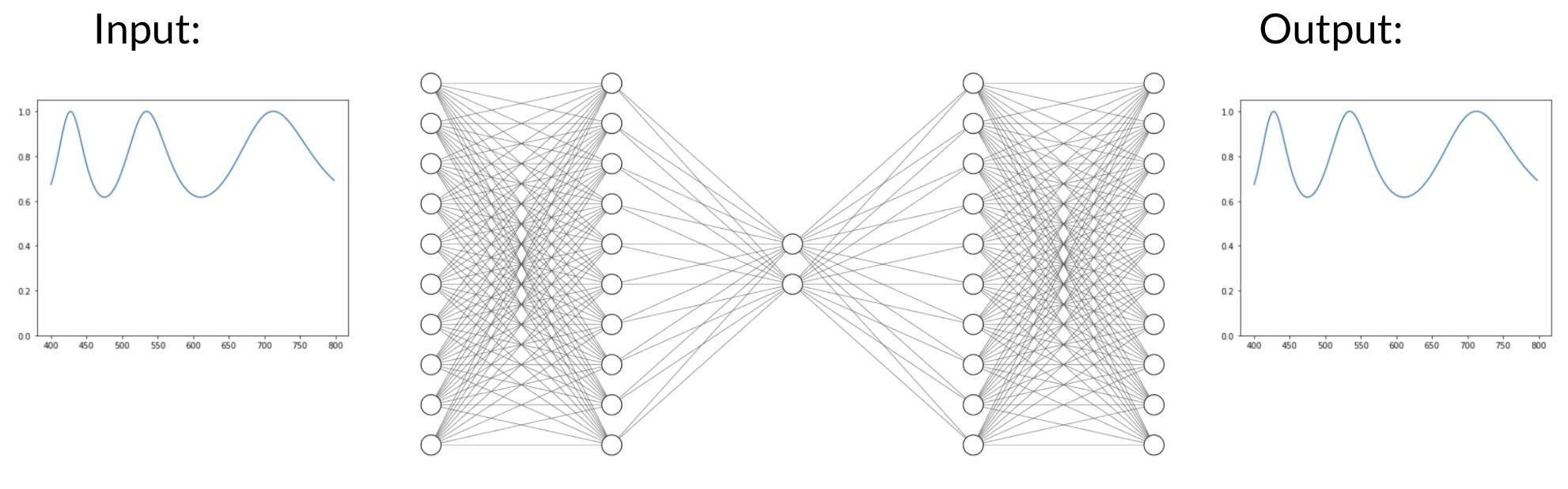}
    \caption{Architecture of the encoder-decoder. A transmission $T(\lambda)$ is encoded into a latent representation. This latent representation is then decoded to make a reconstruction of the original transmission.}
    \label{fig: beta-VAE architecture}
\end{figure}

We encode the transmission $T(\lambda)$ as a set of 200 discrete points for wavelengths between 400 nm and 798 nm, with an interval of 2 nm. Then we construct an encoder-decoder architecture. The encoder compresses the transmission into a latent representation. In principle, this latent representation needs 2 dimensions to store $F$ and $\delta_0$. A key property of the $\beta$-VAE is however, that it only stores information in a latent dimension if it needs this dimension to make accurate reconstructions. We therefore use 5 latent dimensions in our experiments and hope to see information stored in 2 of these dimensions. The other 3 dimensions should be empty. This means that every input is mapped to a unit gaussian $\mathcal{N} (0, 1)$ in these empty dimensions. The decoder then maps the latent representation to a reconstructed transmission. The encoder-decoder architecture is shown in figure \ref{fig: beta-VAE architecture}.

$\\$
The first thing we tested is whether this architecture is capable of learning $F$ and $\delta_0$. Moreover, we would like to know how deep our network should be. To answer these questions, we train an encoder and decoder by supervised learning. We can do this, since we already know the parameters $F$ and $\delta_0$ we are looking for. In more general cases, this is not always possible. 

$\\$
We propose a decoder network with the same structure as in chapter \ref{chap 4: predicting transmission}. It maps the parameters $F$ and $\delta_0$ to a transmission $T(\lambda)$. We propose a decoder network with 4 hidden layers, 100 nodes in each layer and the Swish activation function. We try to keep the network as small as possible, since the loss function of the $\beta$-VAE makes it hard to train large networks. The encoder network maps a transmission $T(\lambda)$ to the parameters $F$ and $\delta_0$ that were used to create it. The network has the same architecture as the decoder, 4 hidden layers, 100 nodes in each layer and the Swish activation function. 

\begin{figure}
    \centering
    \includegraphics[width=\textwidth]{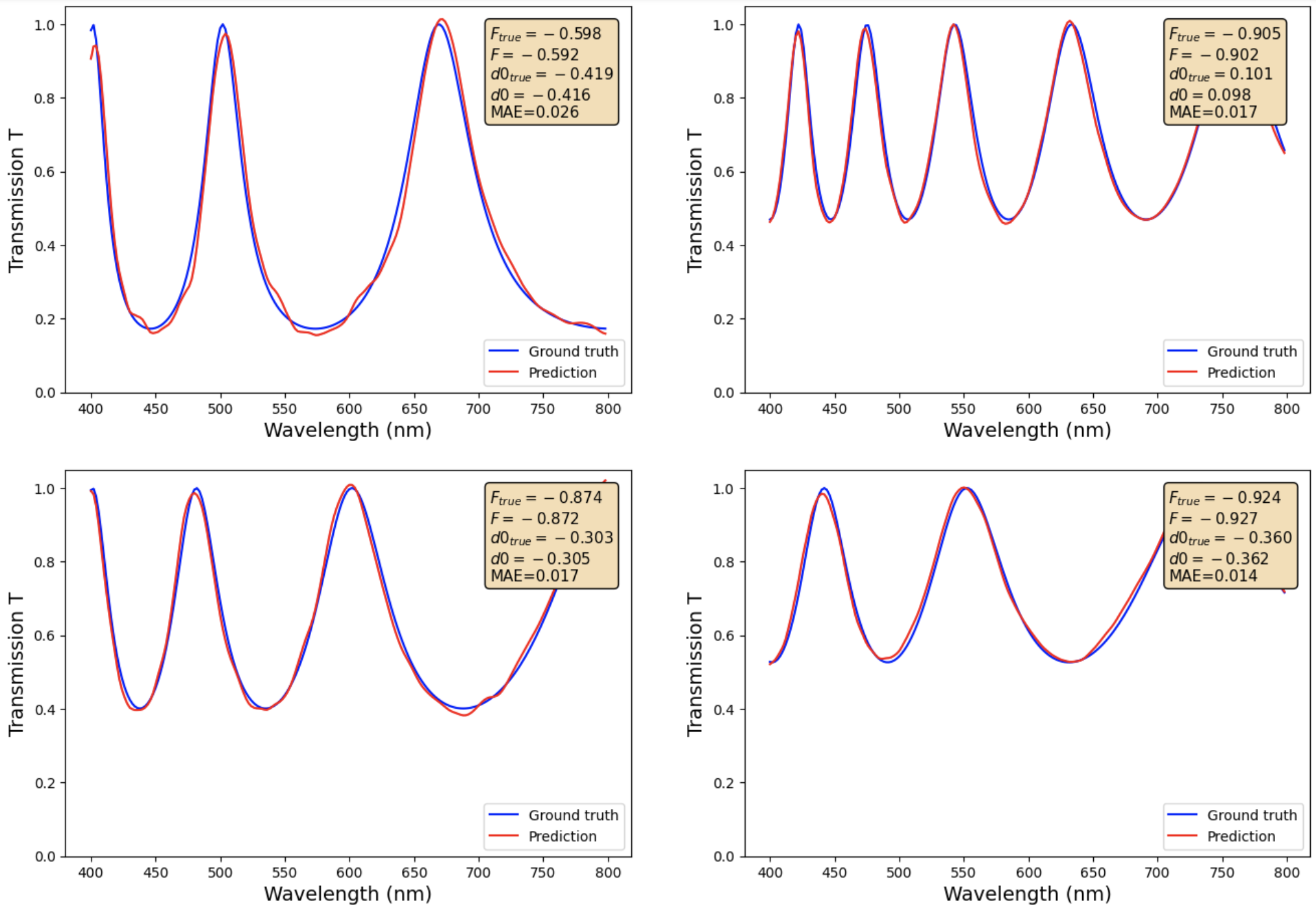}
    \caption{Reconstructions of transmissions $T(\lambda)$ in the test set for the autoencoder trained by supervised learning. The box in the upper right show the true parameters used to compute the transmission in the dataset and the parameters predicted by the encoder. The predicted parameters are used by the decoder to make a reconstruction of the transmission. The mean absolute error (MAE) is computed between the reconstruction and the original transmission.}
    \label{fig: supervised FD}
\end{figure}

$\\$
After training the encoder and the decoder by supervised learning, we can put them together to emulate an autoencoder. We apply the encoder to a transmission $T(\lambda)$ to get the parameters $F$ and $\delta_0$. Then we use this as input for the decoder to reconstruct $T(\lambda)$. Results on some transmissions from the test set are shown in figure \ref{fig: supervised FD}.

$\\$
For the transmissions in the test set displayed, the parameters $F$ and $\delta_0$ are close to the true parameters for the transmissions in the test set. On the full test set, we obtain an MAE on the reconstructions of 3.425\%. Note that this contains the inaccuracy of both networks. The encoder will not give the true parameters $F$ and $\delta_0$. The decoders starts from these incorrect values and needs to make further predictions. This explains why we obtain worse results than in chapter \ref{chap 4: predicting transmission}. 

$\\$
The final architecture is essentially the same as an autoencoder. The only difference is the training. Normally, an autoencoder is trained unsupervised without any knowledge of the parameters it should have in the bottleneck. What we have done here proves however, that a good autoencoder exists for this problem. Moreover, we see that a moderately complex architecture is able to do the job. We therefore expect to see good results with this architecture when we go to the unsupervised learning.

$\\$
An important difference that we have when going to unsupervised learning is that we will not be working with a deterministic autoencoder like the one we have now trained. As explained in chapter \ref{chap 3: deep learning}, much can be gained by making the encoder and decoder probabilistic. The latent representation and the reconstructions are then probability distributions instead of fixed values. This makes sure that the latent representation is continuous, which is desirable since $F$ and $\delta_0$ are continuous variables. This architecture is called a variational autoencoder (VAE). We also work with a variation of the network called the $\beta$-VAE. This variation makes sure that we obtain a disentangled representation. We have proven here that a deterministic autoencoder is capable of learning $F$ and $\delta_0$ and that it can use them to make good reconstructions of the transmissions $T(\lambda)$. Since a $\beta$-VAE is an improvement compared to a simple autoencoder, we suspect that a $\beta$-VAE should also able to learn these parameters. 

\section{Unsupervised training}

We now train a $\beta$-VAE on the transmissions $T(\lambda)$ by unsupervised learning. We try to recover $F$ and $\delta_0$ in the latent representation. We use an encoder with 4 hidden layers with 100 nodes in each layer and the Swish activation function. We saw that the encoder could be written as $q_\phi(\mathbf{z} \vert \mathbf{x})$. The decoder has the same architecture and is written as $p_{\theta}(\mathbf{x} \vert \mathbf{z})$. It was shown above that such an architecture should be able to learn the parameters $F$ and $\delta_0$. 

$\\$
To train the $\beta$-VAE, we need to optimize the parameters $\phi$ and $\theta$ of the encoder and decoder. The optimal parameters minimize the loss function of the $\beta$-VAE given by

\begin{equation}
    \nonumber
    (\phi^*, \theta^*) = \ \argmin_{\phi, \theta} \ \mathbb{E}_{z \sim q_{\phi} (\mathbf{z} \vert \mathbf{x})} \left[ \frac{(\mathbf{x} - f(\mathbf{z}))^2}{2c}\right] + \beta \cdot D_{KL}(q_\phi(\mathbf{z} \vert \mathbf{x}) \vert \vert p(\mathbf{z})). 
\end{equation}

\noindent The function $f(\bf{z})$ is the reconstruction of the transmission. We see thus that the first term gives an MSE loss on the reconstruction. This term will make sure that the $\beta$-VAE is able to make good reconstructions. The second term, the KL divergence, makes sure that the latent representation remains as small as possible. The probability distributions $q_\phi(\mathbf{z} \vert \mathbf{x})$ are kept close to their priors $p(\bf{z})$. These priors are unit gaussians $\mathcal{N}(0, 1)$.

$\\$
Before we can start training, we need to determine the hyperparameters for training. We use the Adam optimization algorithm to minimize the loss. Furthermore, we choose a batch size of 100. We do not look at the validation loss to stop this time. Instead, we will train the network for a fixed amount of 100 epochs. The latent representation has a dimension of 5. 

\begin{figure}
    \centering
    \includegraphics[width=\textwidth]{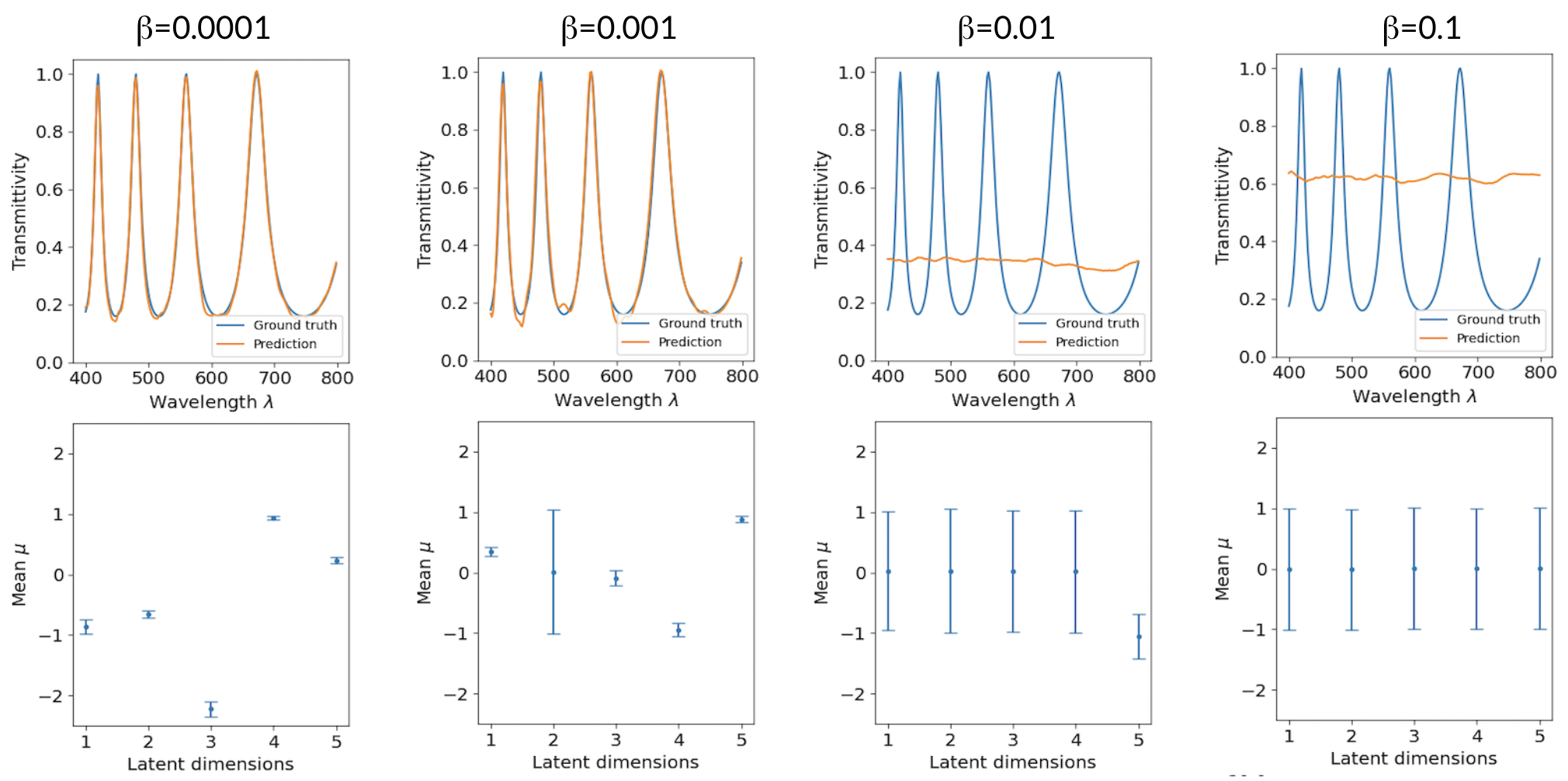}
    \caption{Reconstruction of a transmission from the test set by four $\beta$-VAEs with different values of $\beta$. The lower panels show the mean and standard deviation predicted by the encoder. A latent vector $\bf{z}$ is sampled from this multivariate gaussian distribution to obtain a reconstruction of the transmission.}
    \label{fig: reconstructions}
\end{figure}

\noindent We can train the $\beta$-VAE on the same transmissions $T(\lambda)$ that we had in chapter \ref{chap 4: predicting transmission}. We use the same split into training set, validation set and test set. The parameter $\beta$ is a new hyperparameter that we need to tune. Therefore, we train 4 different networks for different values of $\beta$. The results are shown in figure \ref{fig: reconstructions}. 

$\\$
The figure shows reconstructions of a transmission $T(\lambda)$ from the test set for four different networks with different values of $\beta$. The upper panels show the reconstruction that was obtained. We see that if $\beta$ is too big, we do not get a good reconstruction. We can understand why by looking at the latent representation in the lower panels. Remember that the latent representation are distributions $p(\mathbf{z} \vert \mathbf{x})$, which we chose to be gaussians. The plots show the mean and standard deviations of these gaussian distributions. 

$\\$
If we look at the latent representation for $\beta=0.1$, we see that all dimensions contain the prior distribution $\mathcal{N}(0, 1)$. This means that no information is contained in any dimension. This is what we expect, since a higher $\beta$ leads to a higher importance of the KL divergence which keeps the distribution $p(\mathbf{z} \vert \mathbf{x})$ close to its prior distribution. 

\begin{figure}
    \centering
    \includegraphics[width=0.5\textwidth]{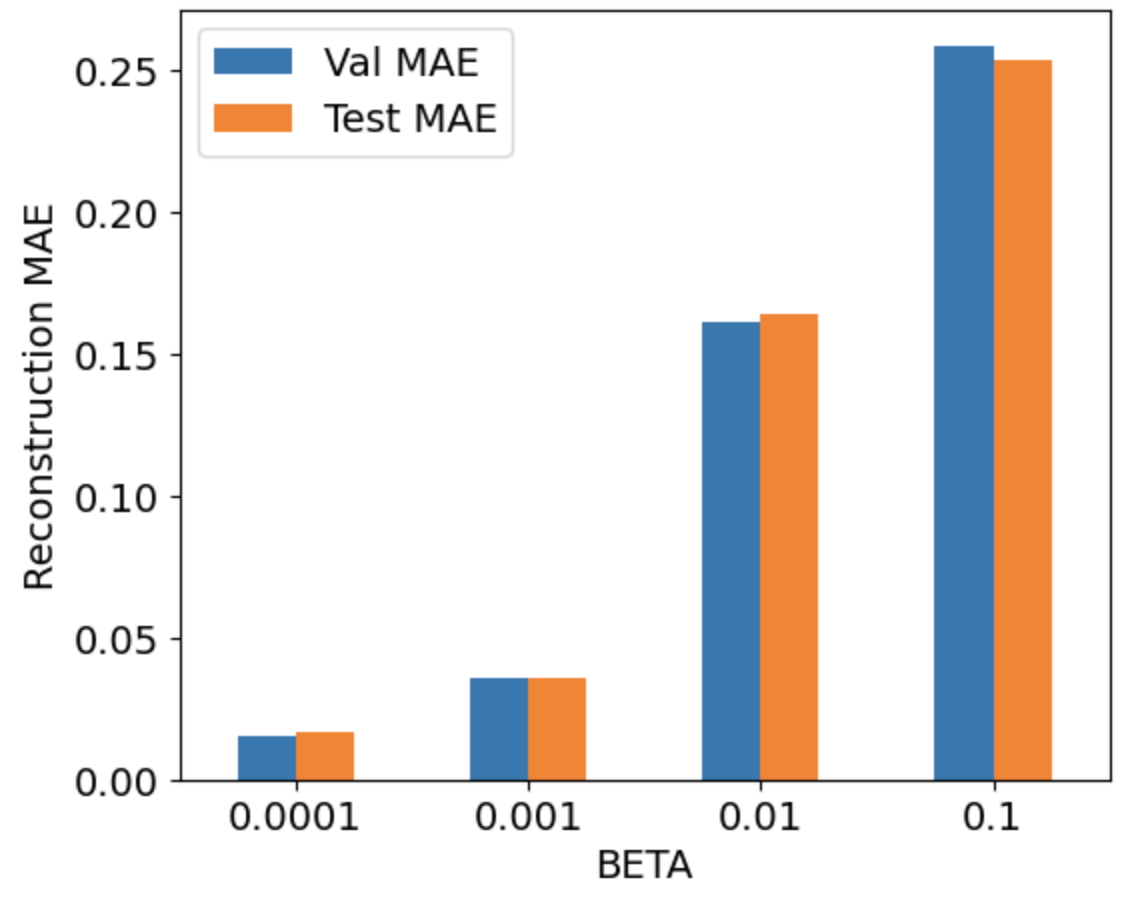}
    \caption{Mean absolute error (MAE) on the reconstructions of the transmissions in the validation and test set. Four networks with different values of $\beta$ are compared. The other hyperparameters stayed the same.}
    \label{fig: BETA plot}
\end{figure}

$\\$
$\\$
For the network with $\beta=0.01$, we see however that one of the dimensions is different from its prior distribution. An equilibrium has been found where one latent dimension contains information and the others remain empty. We also see that one dimension is not enough to make good reconstructions. For the network with $\beta=0.001$, we see that all dimensions contain information, except for the second. With four latent dimensions containing information, accurate reconstructions can be made. Finally, for the network with $\beta=0.0001$, we see that all latent dimensions contain information. This leads to even better predictions. The reconstruction MAE on the full test set is plotted in figure \ref{fig: BETA plot} for the different values of $\beta$.

$\\$
We are now very interested to know what is actually encoded in the latent representation. To investigate this, we analyze the correlation between the latent dimensions and the parameters $F$ and $\delta_0$. We compute the latent vectors and the parameters $F$ and $\delta_0$ for all transmissions in the test set. We plot these latent vectors in a scatterplot with either the parameter $F$ or the parameter $\delta_0$. If a latent dimension would contain one of these parameters, we would observe a straight line. We do not look at the latent representation of the network with $\beta=0.1$, since its latent representation does not contain information. 

\begin{figure}
    \centering
    \includegraphics[width=\textwidth]{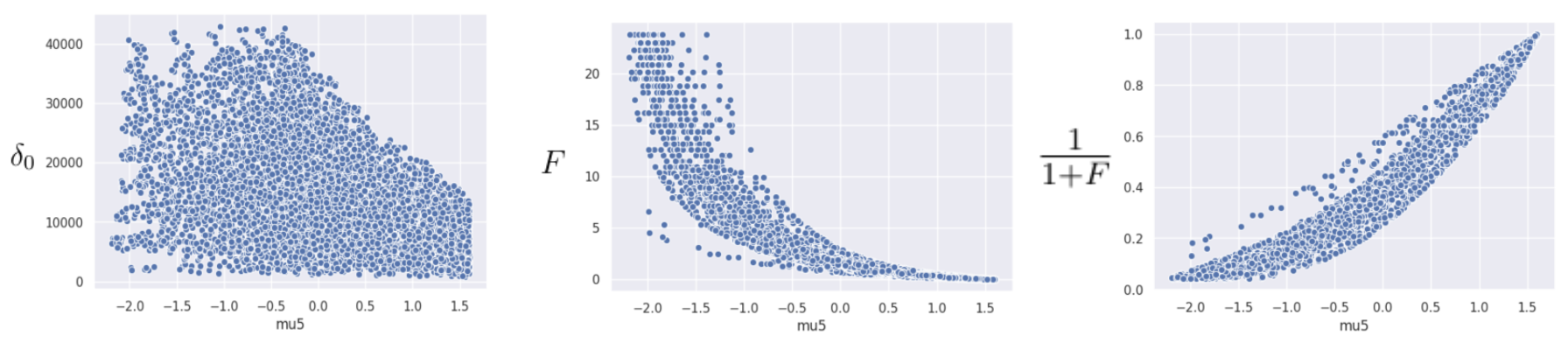}
    \caption{The latent space for the network with $\beta=0.01$. Only the fifth dimension of this latent representation contains information on the transmission. The mean of this dimension is computed for all transmission in the test set and compared to the parameters $\delta_0$, $F$ and $\frac{1}{1 + F}$ for these transmissions.}
    \label{fig: latent space F}
\end{figure}

$\\$
The first network we look at is therefore the one with $\beta=0.01$. Only the fifth latent dimension contains information, while the other dimensions are still equal to their prior distribution. We plot the correlation between the mean of this fifth dimension with $\delta_0$ and $F$ in figure \ref{fig: latent space F}. There seems to be no correlation with $\delta_0$, the data is scattered randomly in latent space. There does seem to be some correlation with $F$, but it has a very specific form. We can recognize the function $1/x$ in it. This should ring a bell. Remember that transmission is given by

\begin{equation}
    T(\lambda, \theta, n, l) = \frac{1}{1 + F sin^2 (\frac{\delta}{2})}.
\end{equation}

\noindent If we would ignore the sine function, the transmission would be equal to $\frac{1}{1 + F}$. We therefore suspect that this transformation of $F$ is stored in the latent representation. If we plot this variable in a scatterplot, we can indeed see a nearly linear correlation.

$\\$
We now turn to the latent dimensions for the networks with $\beta = 0.001$ and $\beta = 0.0001$. These networks are able to give a good reconstruction of the transmission, so we expect that they learn something about $\delta_0$. On closer inspection, we see that one of the latent dimensions of both networks contains the parameter $F$, in the same way as for $\beta = 0.01$. 

\begin{figure}
    \centering
    \includegraphics[width=\textwidth]{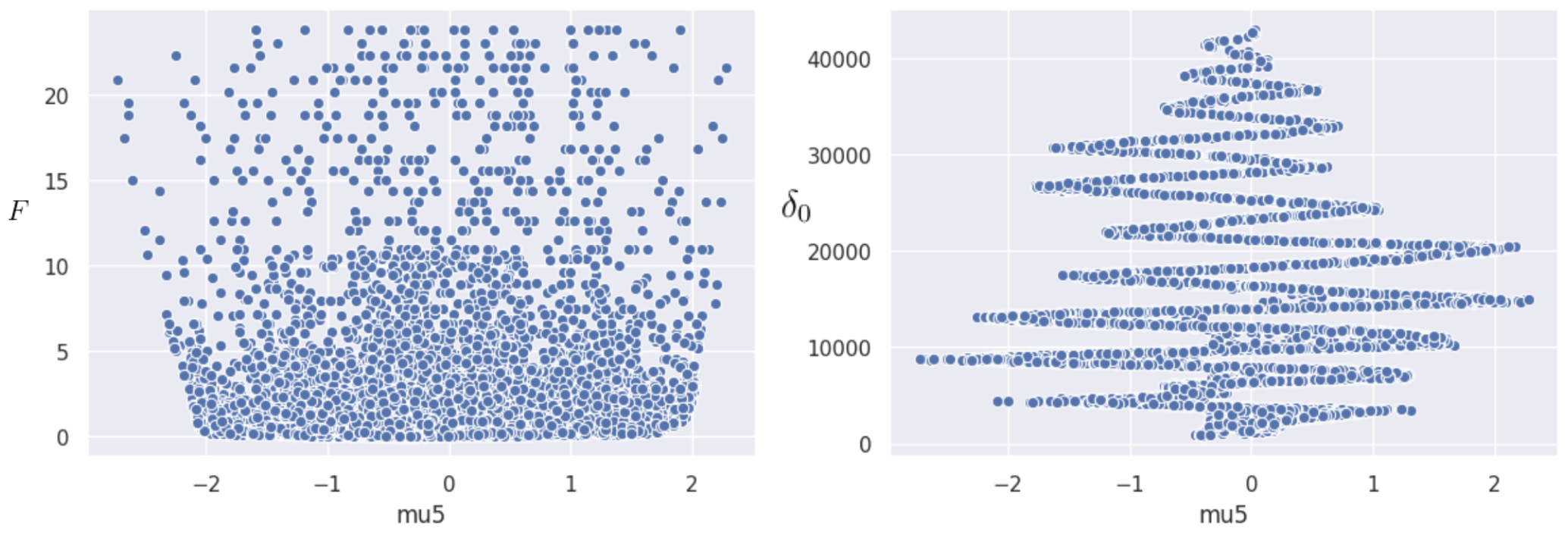}
    \caption{One dimension of the latent space for the network with $\beta=0.0001$. The mean of this dimension is computed for all transmission in the test set and compared to the parameters $F$ and $\delta_0$ for these transmissions. The other latent dimensions of this network and the network with $\beta=0.001$ containing information on $\delta_0$ lead to similar plots.}
    \label{fig: latent space D}
\end{figure}

$\\$
The other dimensions all contain similar information on $\delta_0$, encoded in a very intriguing way. In every dimension, both for $\beta = 0.0001$ and $\beta=0.001$, we observe similar scatterplots. They are shown in figure \ref{fig: latent space D}. There seems to be no correlation with $F$, but there is however a clear correlation with $\delta_0$. We see a wavelike pattern emerge. We are not sure how to interpret this pattern. 

$\\$
$\\$
$\\$
\section{Outlook}

In the experiments that we performed, we were able to retrieve the parameter $F$ approximately in the form of $\frac{1}{1 + F}$. This is a very nice result, because we were able to relate it to the analytical expression for the transmission.

$\\$
The parameter $\delta_0$ was also found in a wavelike pattern. It is unclear to us what this means or what is actually encoded in these latent representations. One hypothesis is that the wavelike pattern is caused by the MSE loss function not being appropriate for the problem. A possible way to change this is change the loss function. We might use the Fourier transform of the transmission $T(\lambda)$, since this takes the oscillation of the transmission $T(\lambda)$ into account. This is also what we do for inverse design, as we will see in the next chapter. In future work, we would like to explore new loss functions to find $\delta_0$. 

$\\$
The reason we chose the $\beta$-VAE instead of the ordinary VAE was to create disentangled representations. In these representations, the latent dimensions are independent from each other. If we look at $F$, we see that we could disentangle this parameter from the rest. In every network, there was only one latent dimension that stored $\frac{1}{1 + F}$, except for the network $\beta = 0.1$ where there was no information in the latent representation. This indicates that the information on $F$ did not get mixed in other latent dimensions. For $\delta_0$ however, we saw distinct but similar encodings in different latent dimensions. The information on $\delta_0$ seems to be spread across multiple latent dimensions. 

%% file: chapters/chapter6.tex
\noindent In chapter 4, we created a neural network that predicts the transmission $T(\lambda)$ of the Fabry-Pérot resonator. We said that this transmission tells us the outcoming color, when white light falls into the resonator. We now pose the inverse question. How can we design a Fabry-Pérot resonator to produce a certain transmitted color? In other words, how can we obtain a specific transmission $T(\lambda)$.

$\\$
To answer this question, we turn to inverse design. As explained in chapter 2, there have been two main approaches to inverse design. The design uses either evolutionary algorithms or gradient-based algorithms. Since we are working with a fully differentiable neural network, it is most natural to use a gradient-based approach. We found inspiration in the paper by Peurifoy et al. \cite{Peurifoy2017}.

\begin{figure}
    \centering
    \includegraphics[width=0.9\textwidth]{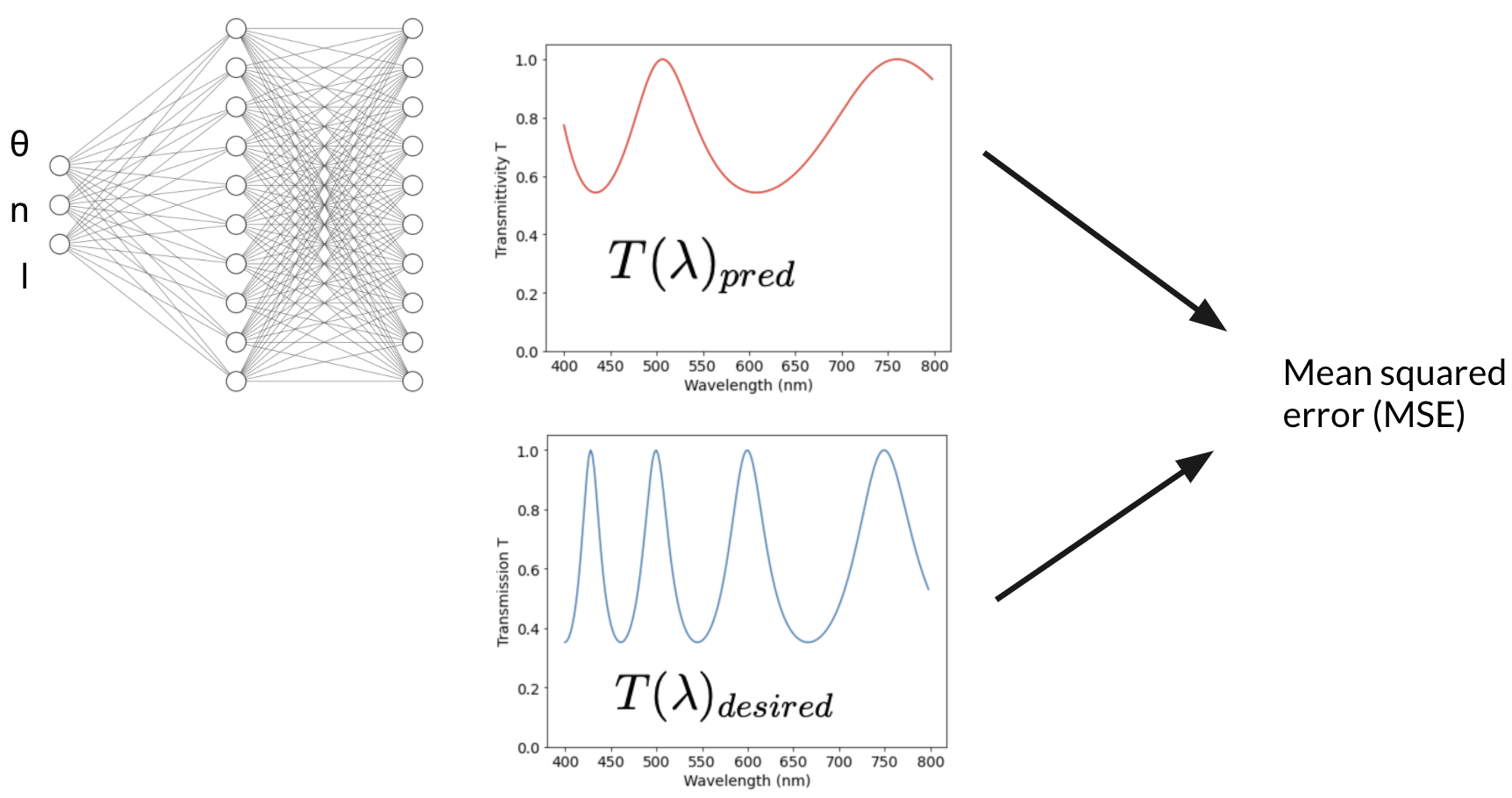}
    \caption{Overview of gradient-based inverse design. A trained neural network is used to predict the transmission of initial design parameters. This transmission is compared to the desired transmission by computing the mean squared error (MSE). The gradient of this loss updates the design parameters while the weights and biases of the network stay fixed.}
    \label{fig: overview inverse design}
\end{figure}

$\\$
The approach we used for this inverse design problem is laid out in figure \ref{fig: overview inverse design}. In the first iteration, the input parameters are initialized at a random value. The neural network is then used to compute the transmission $T(\lambda)_{pred}$ corresponding to the initial design parameters. The predicted transmission is then compared to the transmission $T(\lambda)_{desired}$ that we want to obtain. Their similarity is computed as the mean squared error (MSE). If we could minimize this MSE loss function to be zero, we would obtain two identical transmissions $T(\lambda)$. To do this, we take the gradient of the MSE with respect to the design parameters $\theta$, $n$ and $l$ and update these parameters. The weights and biases of the neural network stay fixed.  After a few iterations, the algorithm converges to the parameters $\theta$, $n$ and $l$ with a minimal MSE. These are the design parameters producing a transmission $T(\lambda)_{pred}$ that is most similar to the transmission $T(\lambda)_{desired}$.

$\\$
To perform gradient descent, we need to choose an algorithm that determines the step size in each iteration. Some very good optimization algorithms have already been created for the training of deep neural networks. An important difference is however that neural networks have thousands of parameters, while we are optimizing just two or three input parameters. There might be algorithms that are better suited than those used for Deep Learning. Our experiments showed however that the Adam optimizer, one of the most popular optimization algorithms for neural networks, leads to results that are as good as we can hope for. 

\begin{figure}[h!]
    \centering
    \includegraphics[width=0.8\textwidth]{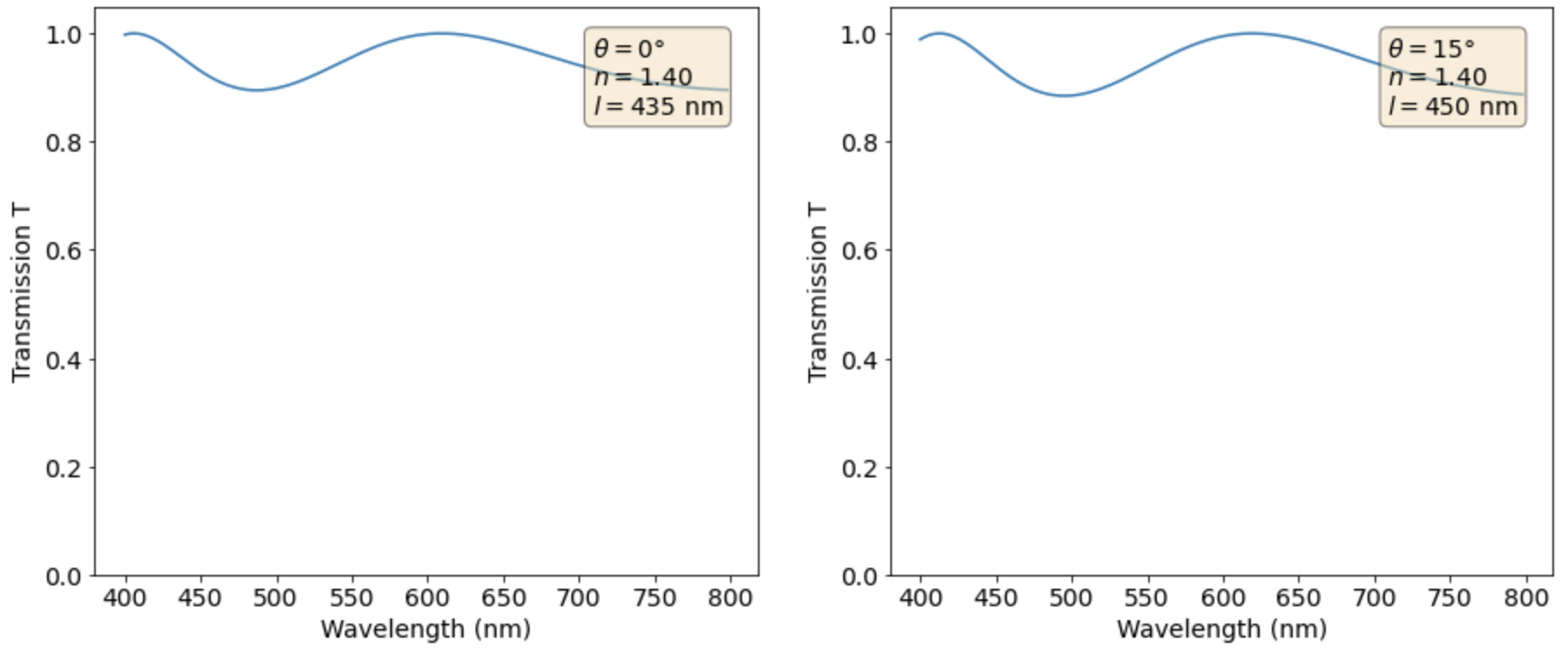}
    \caption{Two similar transmissions $T(\lambda)$ with different design parameters.}
    \label{fig: non-uniqueness}
\end{figure}

$\\$
In our experiments, we encountered two problems. These are the non-uniqueness of the optimal design parameters and the appearance of local minima of the loss function. The non-uniqueness means that two different sets of design parameters produce identical results. On an intuitive level, this can be explained by thinking about interference. The only thing that influences interference is the phase difference between waves $\delta_0$. When two sets of parameters lead to the same phase difference $\delta_0$, the same interference is obtained. For example, if we increase the width of the resonator $l$ and decrease the index of refraction $n$, we can do this in a way that the optical path length $nl$ stays constant. This leads to the same transmission $T(\lambda)$. There are similar relations between $n$ and $\theta$ and between $l$ and $\theta$. The non-uniqueness of the design parameters is illustrated in figure \ref{fig: non-uniqueness}.

\noindent To overcome the non-uniqueness, we have to look at the actual degrees of freedom $\delta_0$ and $F$. In this case, there is a 1-to-1 correspondence between the parameters $\delta_0$ and $F$ and the transmission $T(\lambda)$. Solving the inverse design problem in these variables leads to unique solutions. 

$\\$
The other problem we encountered is the appearance of local minima. This is a consequence of the choice of loss function. The mean squared error (MSE) is not a monotonous function in $\delta_0$. The parameter $\delta_0$ tells us how many times the transmission $T(\lambda)$ oscillates between 1 and a minimal value for wavelengths between 400 nm and 800 nm. It is possible to have a transmission with twice the number of oscillations, but still a rather low MSE. The algorithm would have to climb a hill where the MSE first increases before getting to the global optimum. In order to overcome this problem, we look at the Fourier transform of the transmission $T(\lambda)$.

$\\$
In this chapter, we first perform inverse design on $F$ and $\delta_0$. In this set-up, we do not have the problem of non-uniqueness. We do have the problem of local minima. We discuss how this problem can be overcome using the Fourier transform of the transmission $T(\lambda)$. Using the insights of this first part, we then turn to the harder problem of inverse design on $\theta$, $n$ and $l$. We discuss how the problem of local minima can be overcome with the same Fourier-based technique as for $F$ and $\delta_0$. Then we discuss how the non-uniqueness manifests itself. Finally in the last part of this chapter, we show how to use the methods of inverse design to design the Fabry-Pérot resonator that most closely matches an arbitrary transmission $T(\lambda)$. 

\section{Designing optimal $F$ and $\delta_0$}

\begin{figure}[h!]
     \centering
     \begin{subfigure}[b]{\textwidth}
         \centering
         \includegraphics[width=0.7\textwidth]{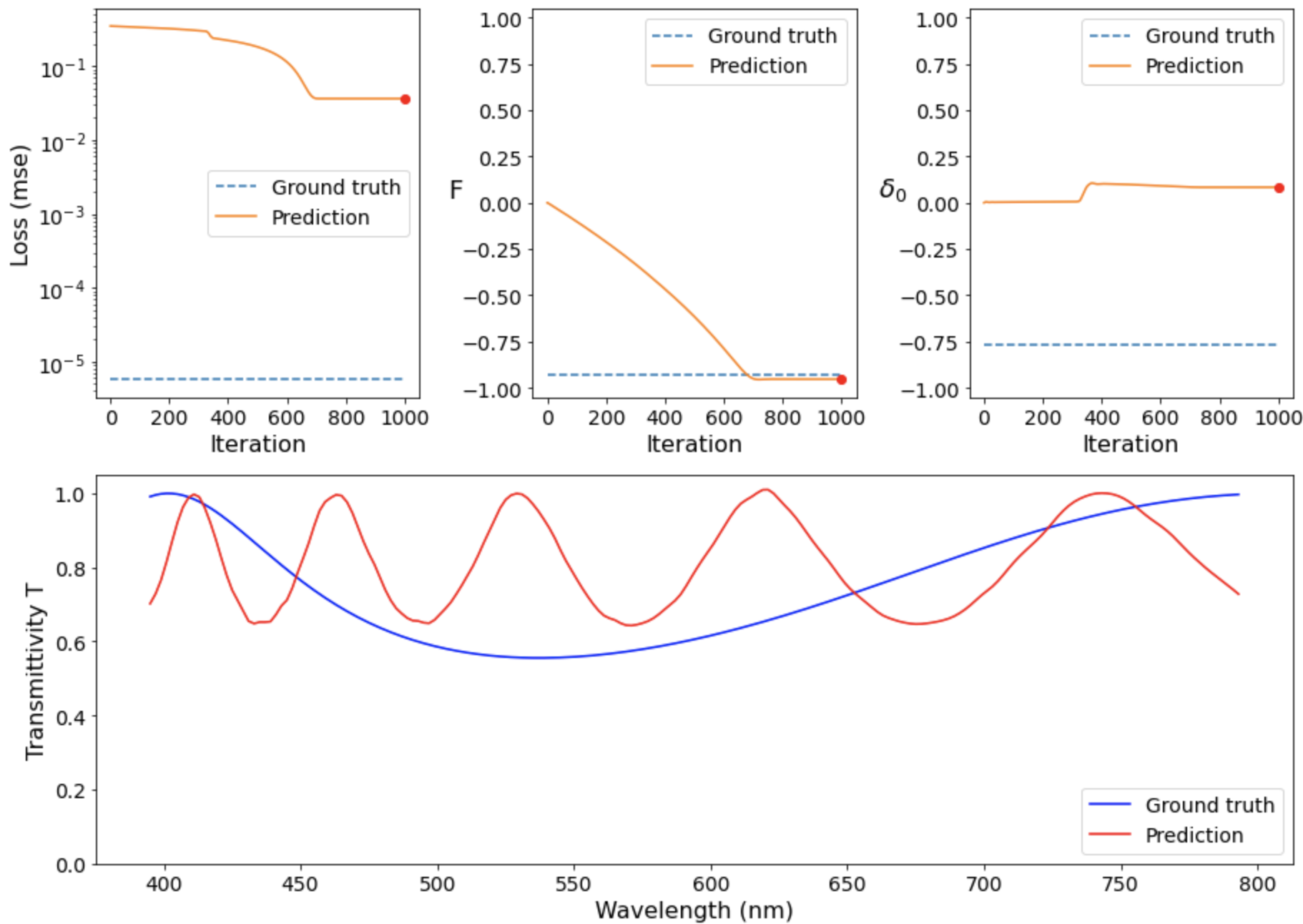}
         \caption{}
         \label{fig: inverseFD 82a}
     \end{subfigure}
     \hfill
     \begin{subfigure}[b]{\textwidth}
         \centering
         \includegraphics[width=0.7\textwidth]{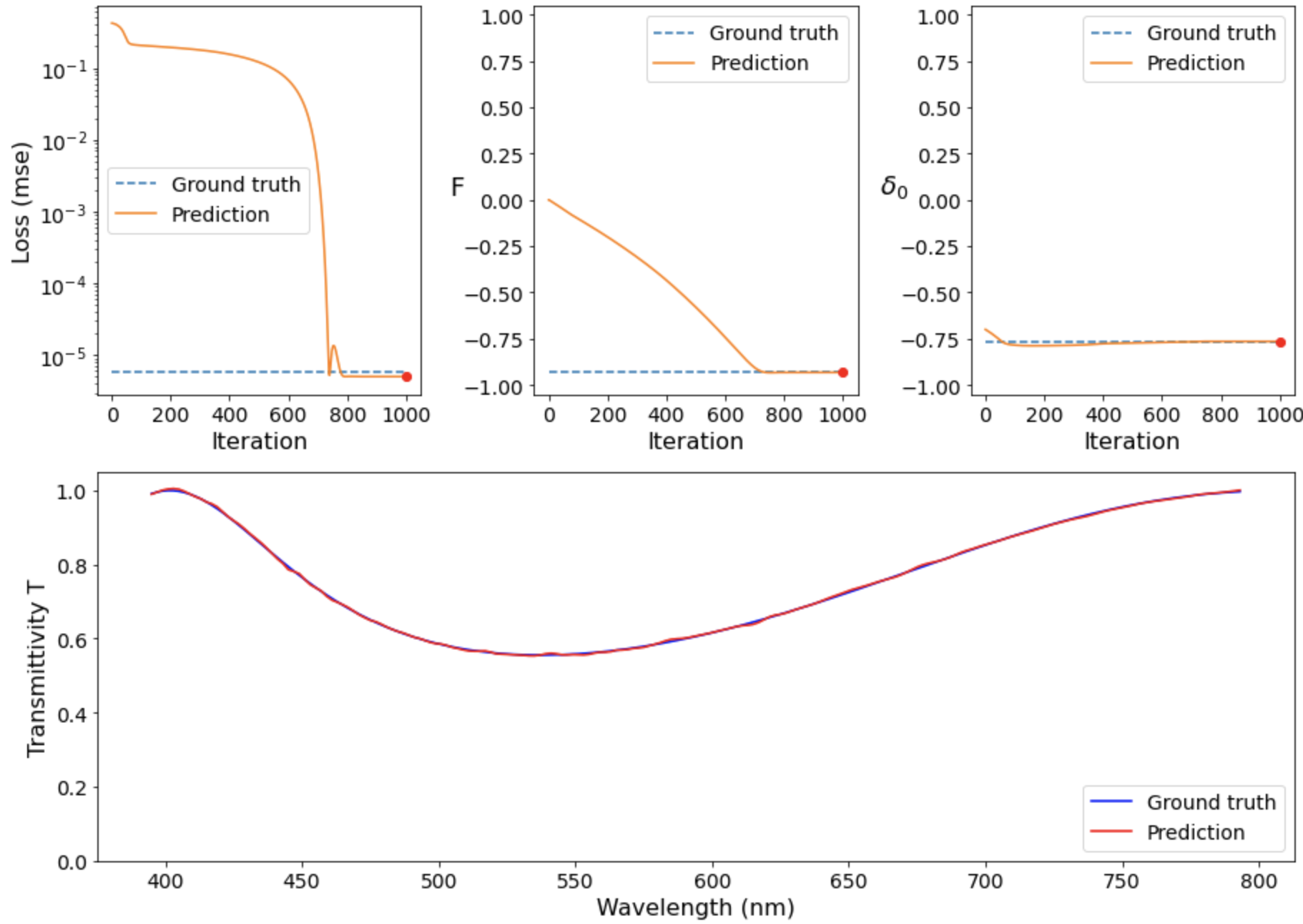}
         \caption{}
         \label{fig: inverseFD 82b}
     \end{subfigure}
        \caption{Inverse design for a transmission in the test set. The lower panels show the desired transmission in blue and the transmission predicted for the obtained design parameters in red. The upper panels show the evolution of the design parameters and the loss during training. a) Local minimum obtained by initializing $\delta_0=0$. The parameters $F$ and $\delta_0$ converge to fixed values, but the transmissions do not look alike. b) Global minimum obtained by initializing $\delta_0=-0.70$. The predicted transmission is very close to the desired transmission.} 
        \label{fig: inverse FD 82}
\end{figure}

We start by inverse design on the parameters $F$ and $\delta_0$. As mentioned earlier, we encountered the problem of local minima. Ideally, our initial values for $F$ and $\delta_0$ could be any random value and the optimization algorithm would still converge to a global optimum. This is however not the case, as figure \ref{fig: inverse FD 82} indicates. The figure shows inverse design on a transmission $T(\lambda)$ from the test set. 

$\\$
In these images, the upper panels represent the evolution of the parameters $F$ and $\delta_0$. The dotted line shows the true parameters from which the desired transmission $T(\lambda)_{desired}$ was computed. This desired transmission is plotted in blue in the bottom left. We call this the ground truth, as is standard in Machine Learning. The red line shows the prediction $T(\lambda)_{pred}$ of the neural network for the proposed parameters $F$ and $\delta_0$. The plot in the upper left shows the mean squared error (MSE) between those two transmissions. The blue dotted line indicates what the loss would be for predictions with the correct parameters $F$ and $\delta_0$. We take this as a baseline since the predictions of the neural network are not perfect. We can not expect a much lower loss than we would obtain for the true parameters. 

$\\$
In figure \ref{fig: inverseFD 82a} we get stuck in a local minimum. The parameters converged to fixed values, but the transmissions do not look alike. If we look at the loss, we see that it is far higher than what it would be for the true design parameters. Figure \ref{fig: inverseFD 82b} shows what happens if we choose an initial value closer to the true $\delta_0$. Now, the gradient descent does converge to a global optimum. We see that the transmissions are indeed very similar. Looking at the loss, we even see that it is lower than it would be for the true parameters. This is an intriguing result! It suggests that the inverse design is correcting for the inaccuracy of the neural network predicting $T(\lambda)$.  

\begin{figure}[h!]
     \centering
     \begin{subfigure}[b]{\textwidth}
         \centering
         \includegraphics[width=0.8\textwidth]{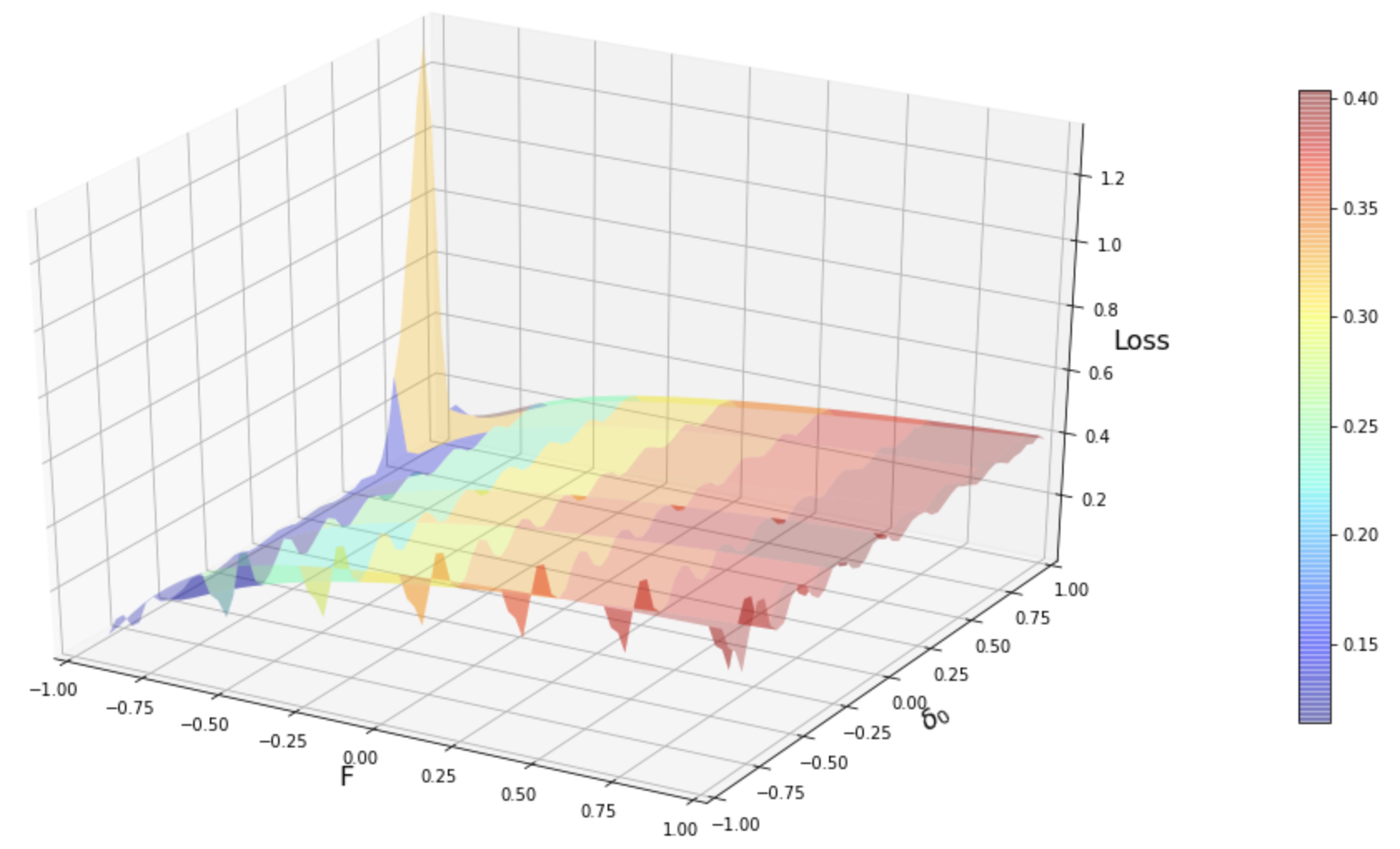}
         \caption{}
         \label{fig: loss FD 3d}
     \end{subfigure}
     \hfill
     \begin{subfigure}[b]{\textwidth}
         \centering
         \includegraphics[width=0.8\textwidth]{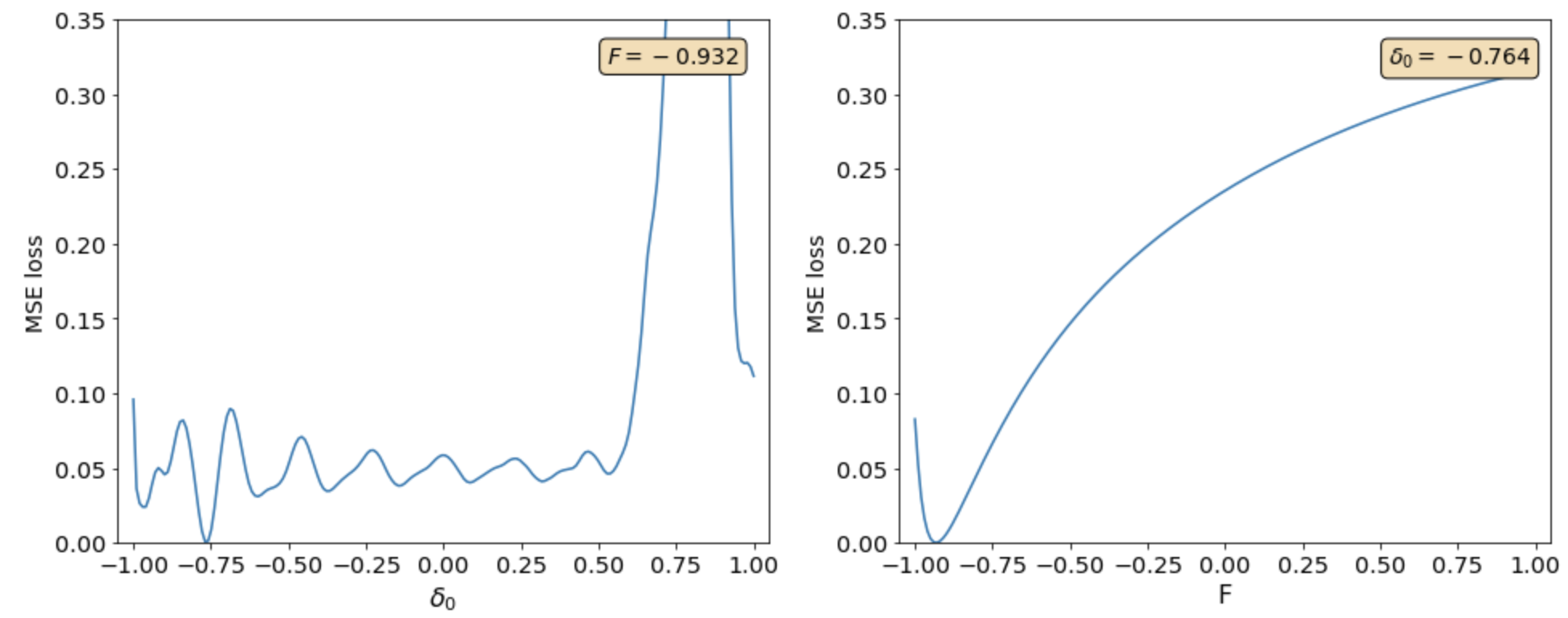}
         \caption{}
         \label{fig: loss FD slice}
     \end{subfigure}
       \caption{Mean squared error (MSE) on the transmissions in function of $\delta_0$ and $F$. The desired transmission is a transmission $T(\lambda)$ from the test set. a) The loss function that we want to optimize. The loss shows a lot of local minima in $\delta_0$. b) Two sections of the loss for the true design parameters. The first plot shows a section for the true parameter $F = -0.932$. The second plot shows a section for the true parameter $\delta_0 = -0.764$. } 
    \label{fig: loss FD}
\end{figure}

$\\$
The underlying reason for the local minimum is that the mean squared error (MSE) does not capture the similarity of two transmission as a convex function of $\delta_0$ with only one global minimum. This is illustrated in figure \ref{fig: loss FD}. In these plots, we chose a transmission from the test set and computed the MSE between this transmission and a predicted transmission as a function of $F$ and $\delta_0$. 

$\\$
In figure \ref{fig: loss FD 3d}, we see that the loss landscape appears to have waves along the direction of $\delta_0$. This leads us to believe that the local minima are caused by $\delta_0$. We investigate this by looking at sections of the loss landscape. The loss landscape has its global minimum for the true parameters of $F$ and $\delta_0$. We perform a section along $F$ and a section along $\delta_0$ that contains this global minimum. This is shown in figure \ref{fig: loss FD slice}.

$\\$
In function of the parameter $F$, the MSE loss is very suited for gradient descent. There is only one global minimum to which we can converge. The story is different for $\delta_0$. The MSE loss shows very sharp local minima, from which it is difficult to escape. Gradient descent is only able to find the global minimum if we can somehow get close to the global optimum.

$\\$
The approach we propose to get close to the global optimum is to look at the Fourier transform  of the transmission $T(\lambda)$. The premise is that since $\delta_0$ is linked to the number of oscillations of the transmission, this information on $\delta_0$ can be captured in the Fourier transform. 

$\\$
We used the fast Fourier transform algorithm (FFT) to compute the Fourier power spectrum of $T(\lambda)$. This gave us the power for each frequency in the range from 1 to 100. Note that the frequency is the number of oscillations that the transmission makes between 400 nm and 800 nm. If we look at the transmissions in the data set, we saw that no transmission oscillated more than 10 times. Therefore, we only look at the power in frequencies up to 10. We normalize the power in these 10 frequencies such that the total power is 1. The result then only depends on the frequency of the oscillation and not on its amplitude. 

\begin{figure}[h]
    \centering
    \includegraphics[width=0.8\textwidth]{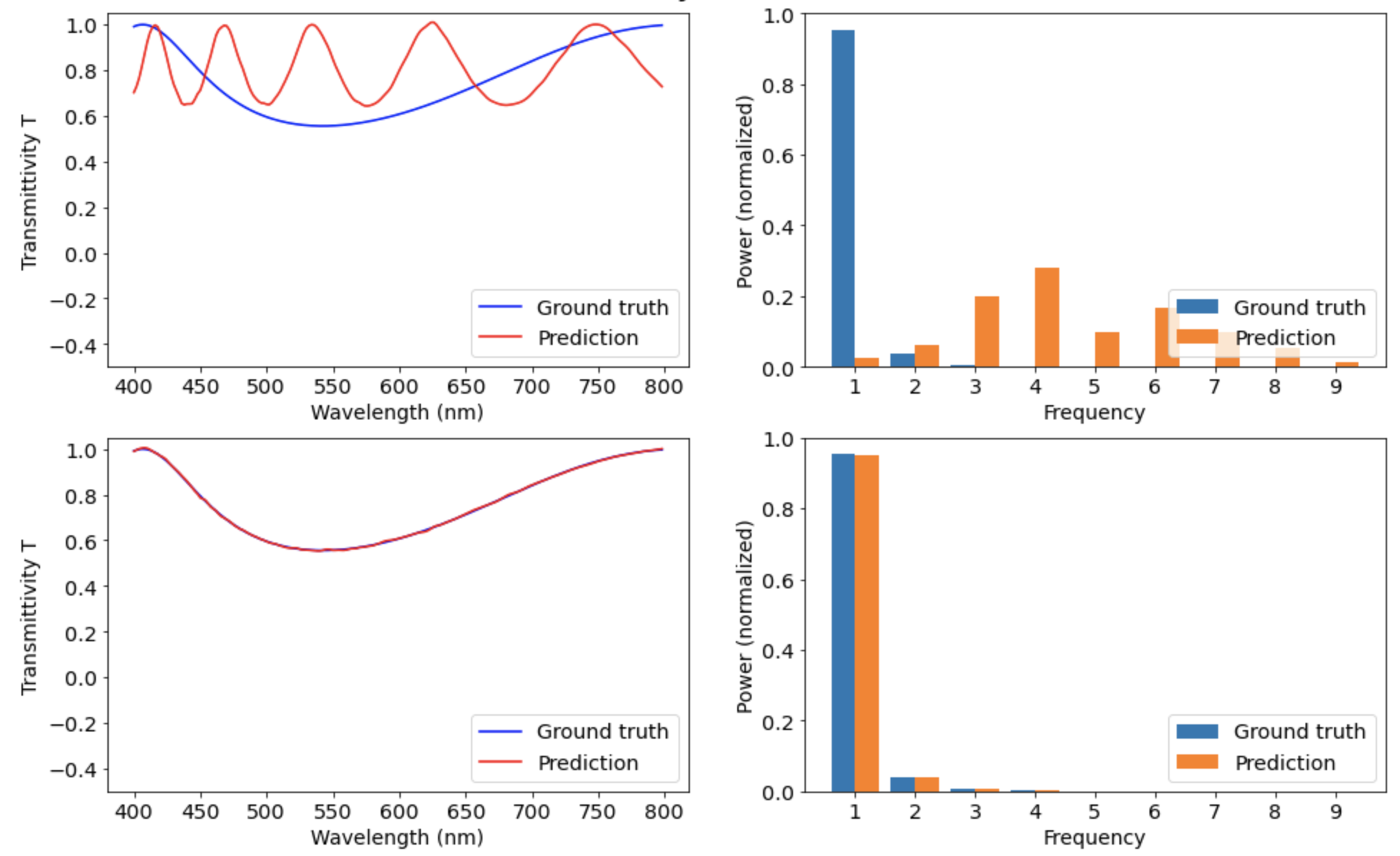}
    \caption{The transmissions in the local and global minima obtained earlier. The Fourier power spectrum of the true and predicted transmissions is shown on the right.}
    \label{fig: fourier local-global minimum}
\end{figure}

\noindent To see how the Fourier transform can help us, we take a look at figure \ref{fig: fourier local-global minimum}. The local and global minima we obtained earlier are plotted together with their normalized Fourier power spectrum. We see that for the local minimum, the predicted and desired transmissions have a very different Fourier power spectrum. For the second solution where we obtain a global optimum however, the Fourier power spectra of the desired and predicted transmissions are nearly the same. This means that we can look at the Fourier power spectrum to differentiate between a local and a global minimum. 

$\\$
We saw earlier that the gradient descent on the mean squared error (MSE) of the transmissions got stuck in local minima. Nonetheless, if we started from an initial value $\delta_0$ close to the global minimum, the gradient descent was able to find this global minimum. We now use the Fourier power spectra of the transmissions to obtain such a good initial value $\delta_0$ close to the global optimum.  . 

\begin{figure}[h!]
     \centering
     \begin{subfigure}[b]{\textwidth}
         \centering
         \includegraphics[width=0.8\textwidth]{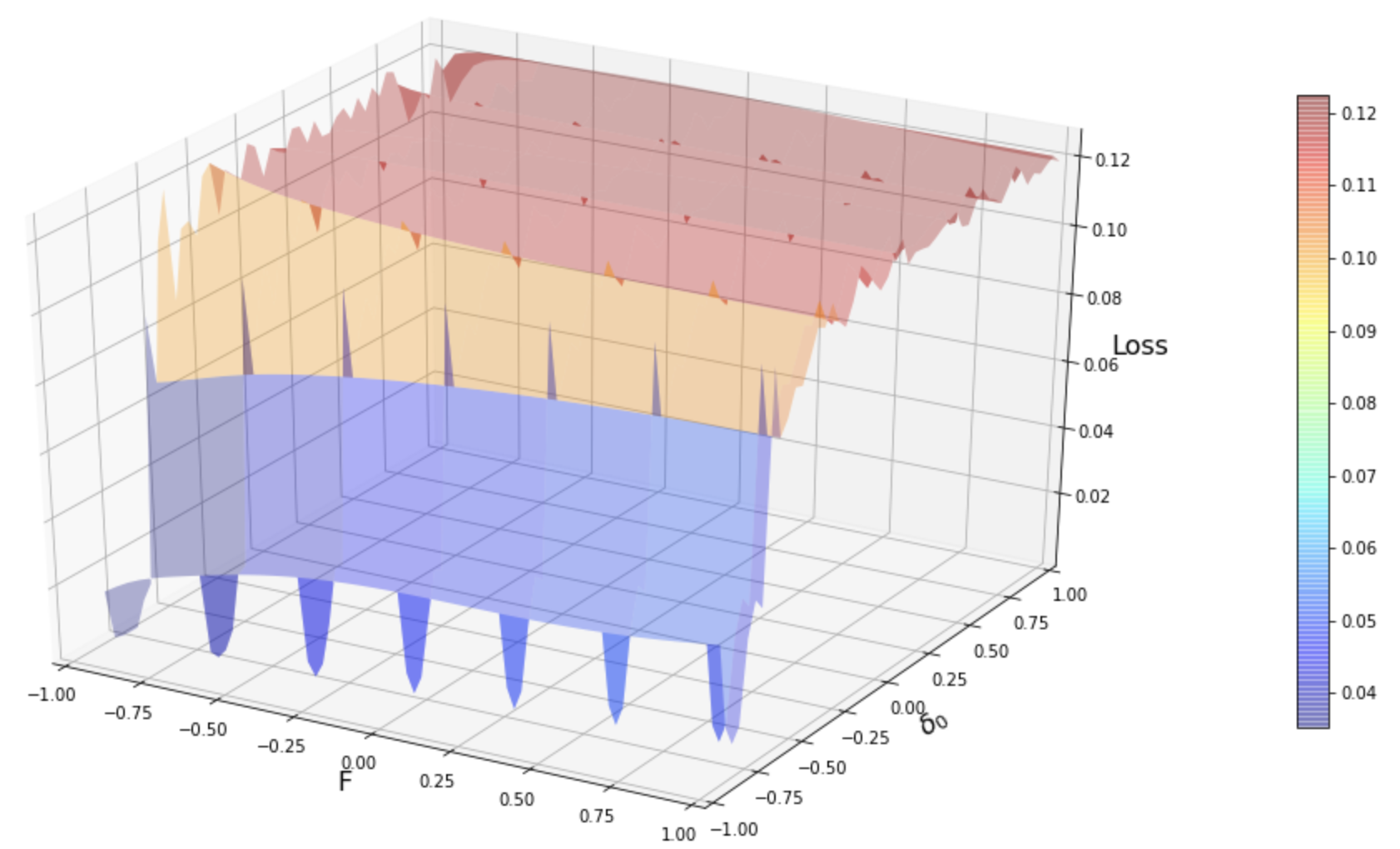}
         \caption{}
         \label{fig: fourier FD 3d}
     \end{subfigure}
     \hfill
     \begin{subfigure}[b]{\textwidth}
         \centering
         \includegraphics[width=0.8\textwidth]{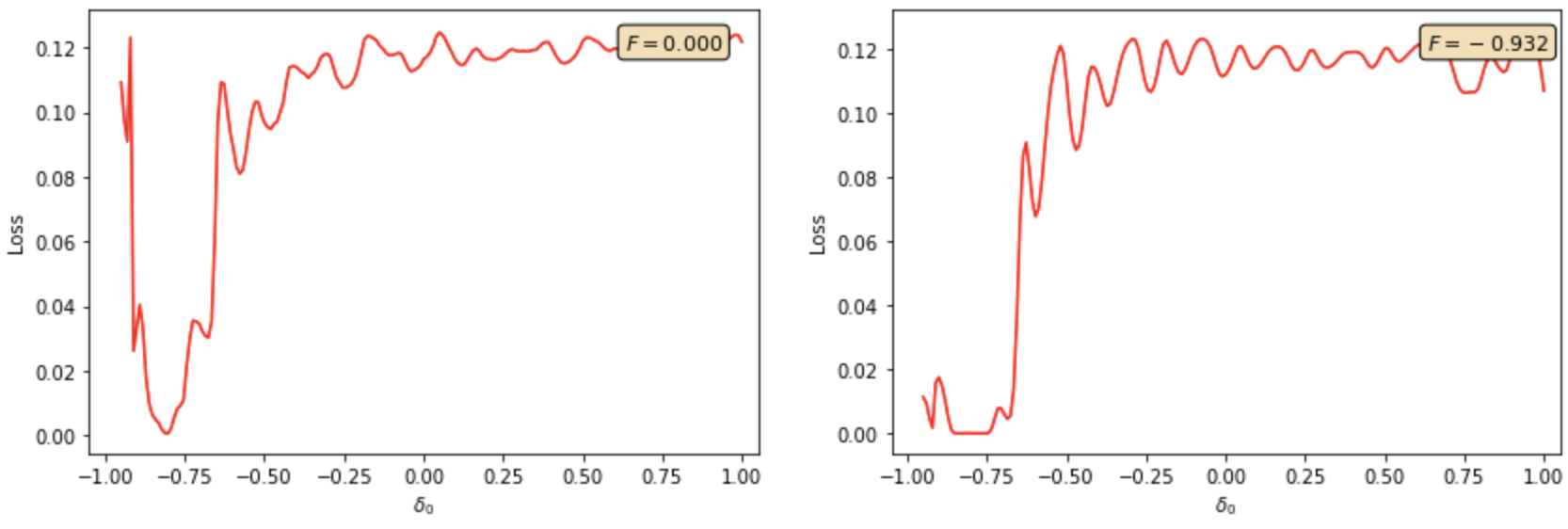}
         \caption{}
         \label{fig: fourier FD slice}
     \end{subfigure}
       \caption{Mean squared error on the Fourier power spectra of the transmissions in function of $\delta_0$ and $F$. The desired transmission is a transmission $T(\lambda)$ from the test set. a) Loss landscape, sections for $F$ constant seem similar. b) Two sections of the loss as a function of $\delta_0$. The first plot shows a section for $F=0$. The second plot shows a section for the true parameter $F = -0.932$. } 
    \label{fig: fourier FD}
\end{figure}

$\\$
We introduce a second loss function, the mean squared error (MSE) on the Fourier power spectrum. This loss function is shown in figure \ref{fig: fourier FD}. Since we normalized the Fourier power spectrum, this new loss function should not be dependent on the amplitude of the oscillation and therefore not be dependent on $F$. This means that every section of the loss landscape along $\delta_0$ will approximately be the same. We see in figure \ref{fig: fourier FD 3d} that this indeed seems to be the case.

$\\$
We can use this fact to initialize $\delta_0$. Two sections for $F=0$ and the true parameter $F=-0.932$ are plotted in figure \ref{fig: fourier FD slice}. The value of $\delta_0$ at minimum of the section for the true parameter of $F$ is the value of $\delta_0$ at the global minimum. We see however that the section for $F=0.0$ has its minimum at nearly the same value of $\delta_0$. This means that we can get close to the true parameter of $\delta_0$ by minimizing the section for $F=0.0$.

$\\$
To minimize a section, we do not need gradient descent. We can simply compute the loss for 200 values of $\delta_0$. We then select the minimum of these values to obtain an initial value for $\delta_0$. Once we have obtained an initial value for $\delta_0$, we perform gradient descent as usual on the MSE of the transmissions. Since the gradient descent is now initialized close to the global minimum, we expect good results.

$\\$
Unfortunately, the assumption that the sections of the MSE on the Fourier power spectra are equal for all values of $F$ is not perfectly valid, as we already see in figure \ref{fig: fourier FD slice}. It might be that minimizing the section for $F=0$ leads to an initial value $\delta_0$ that is still quite far away from the global optimum. To mediate this problem, we re-initialize $\delta_0$ after a few iterations of gradient descent, when we are closer to the true parameter $F_{true}$. We propose to make a new guess every 100 epochs. For values of $F$ that are closer to the true parameter, the sections will be more similar to the section with the global minimum. This way, the initial value for $\delta_0$ gets closer to the true parameter every time we re-initialize.

$\\$
In the end, we are not interested in matching the Fourier transforms of the transmissions, but only interested in matching the transmissions themselves. In principle, a matching Fourier transform should imply matching transmissions, but this is not the case since we do not take information on the phase of different frequencies into account. This could mean that we initialize $\delta_0$ at a value that minimizes the loss on the Fourier spectra, while the predicted transmissions are still far away. 

$\\$
To make up for this disagreement, we propose to replace the simple mean squared error on the Fourier spectra by a combined loss function. This combined loss is the sum of the MSE on the transmissions and the MSE on the Fourier power spectra of the transmissions. In this way, we can take the MSE loss on the transmissions into account. 

$\\$
At last, one could argue that we might not need the Fourier transform at all. Minimizing the original MSE on the transmissions on sections of constant $F$ every 100 epochs might also provide us with a good initialization and re-initialization for $\delta_0$. We therefore also test the performance when making the initialization for $\delta_0$ based on the original MSE on the transmissions. 

$\\$
We now have three proposed loss functions to give good initializations of $\delta_0$: the MSE on the Fourier power spectra, the MSE on the transmissions or a combination of both. Note that these loss functions are only used to determine an initial value of $\delta_0$ and to re-initialize this value every 100 epochs. In all three cases, the gradient descent is performed on the MSE of the transmissions.

\begin{figure}
    \centering
    \includegraphics[width=\textwidth]{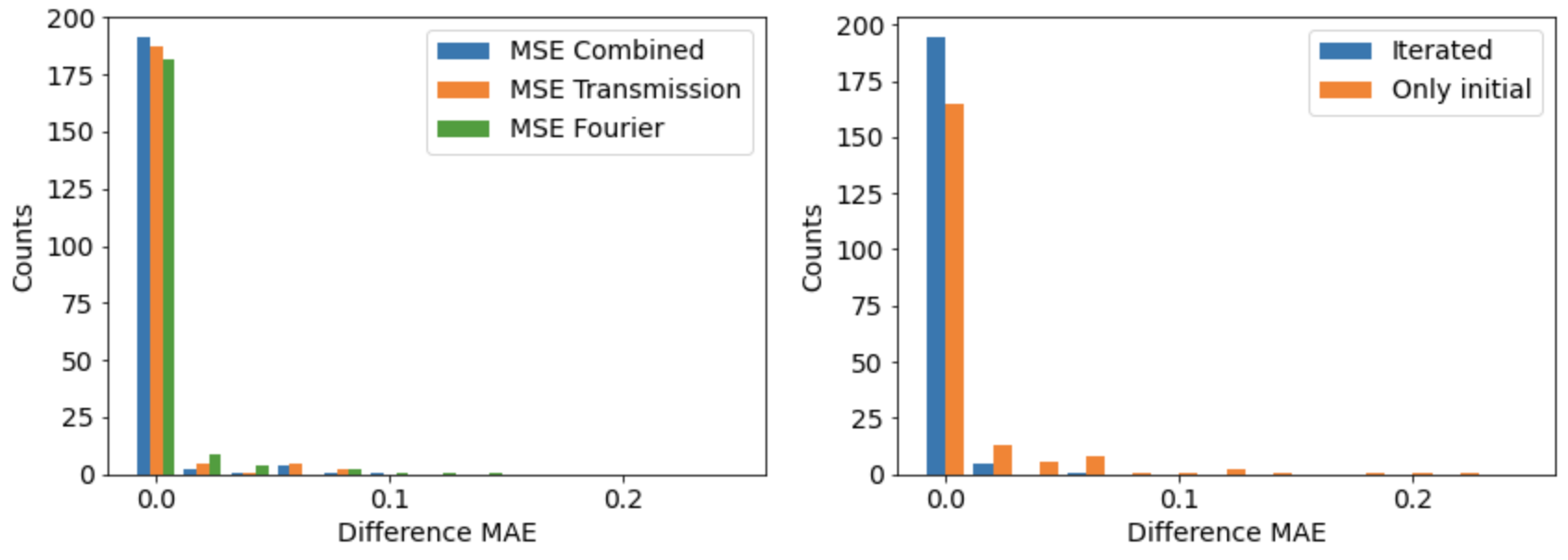}
    \caption{We performed inverse design on 200 transmissions in the test set. We look at the MAE between the desired transmission and the final predicted transmission. We then take difference between this MAE and the MAE we would obtain with the true parameters $F$ and $\delta_0$. Transmissions in the first bin have a MAE of less than $1\%$. In the first plot, we compare three loss functions to initialize $\delta_0$. After 100 epochs, the loss function was used to re-initialize $\delta_0$. The second plot compares our method when we re-initialize $\delta_0$ to the case where we only use the loss for initialization of $\delta_0$. We used the combined MSE in the second plot.}
    \label{fig: inverse design FD result}
\end{figure}

$\\$
In our experiments, we apply this method to 200 transmissions in the test set. When the inverse design was complete, we compute the mean absolute error (MAE) between the desired transmission $T(\lambda)_{desired}$ and the final predicted transmission $T(\lambda)_{pred}$. To take the inaccuracy of the predicting neural network into account, we compare this to the MAE that would be obtained with the true design parameters. Results on 200 transmissions from the test set are shown in figure \ref{fig: inverse design FD result}. 

$\\$
For the transmissions in the first bin, the difference in MAE compared to what we would have for the true parameters was lower than 0.01. Most of the transmissions end up in this bin, so we can conclude that all three loss functions lead to good results. Of the three, the combined loss is superior to the others, but only by a small margin. We observe that there are some outliers with a high MAE. When looking at these transmissions, we saw that they barely oscillate. It is to be expected that the Fourier transform of these transmissions does not lead to much improvement. 

$\\$
At last, we compare the method when we make a guess for $\delta_0$ every 100 epochs or when we only use it to make an initial guess. We used the MSE Combined loss function to make the comparison. It is clear that making a new guess for $\delta_0$ leads to more transmission with a low MAE. 

\section{Designing optimal $\theta, n, l$}

We now perform inverse design on $\theta, n$ and $l$. We need to make some adaptations compared to our method in the earlier section. We still perform gradient descent on the MSE of the predicted and desired transmissions $T(\lambda)$. To get the initial values right, we look at the Fourier transform of the transmission. In this case, there is no design parameter that clearly determines the Fourier transform. Therefore, we need to compute the minimum for every design parameters. 

$\\$
We compute this minimum in three steps. In the first step, we pick $n$ and $l$ randomly and choose the value of $\theta$ that minimizes the loss for these fixed parameters. Then we use the value obtained for $\theta$ to compute a minimum value of $n$, using a random $l$. At last, we use our guess for $\theta$ and $n$ to compute $l$. In this 3-step process, we end up with initial values for $\theta$, $n$ and $l$. This is illustrated in figure \ref{fig: theta, n, l loss functions 82}. Again, there are three reasonable loss functions we can use to guess the design parameters: the MSE on the transmissions, the MSE on the Fourier power spectrum or a combination of both. 

\begin{figure}[h!]
     \centering
     \begin{subfigure}[b]{0.32\textwidth}
         \centering
         \includegraphics[width=\textwidth]{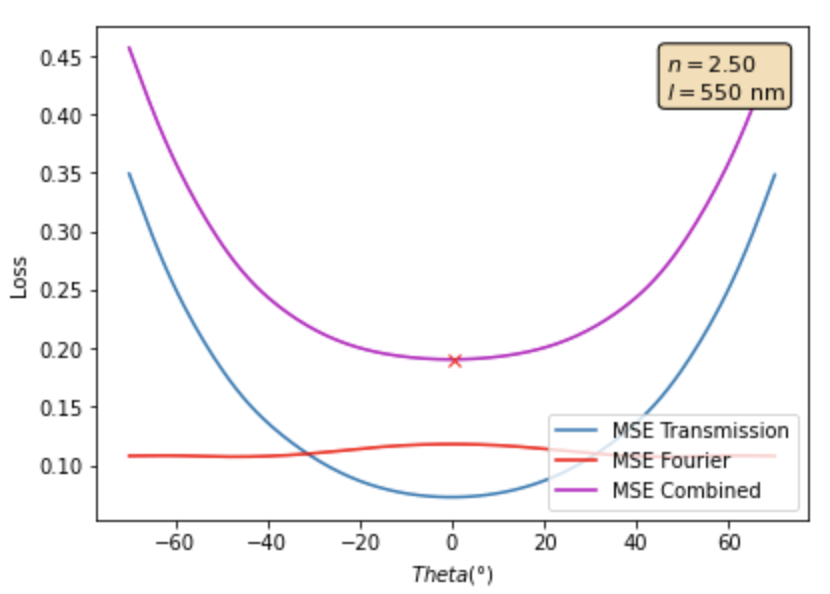}
         \caption{}
         \label{fig: theta loss 82}
     \end{subfigure}
     \hfill
     \begin{subfigure}[b]{0.32\textwidth}
         \centering
         \includegraphics[width=\textwidth]{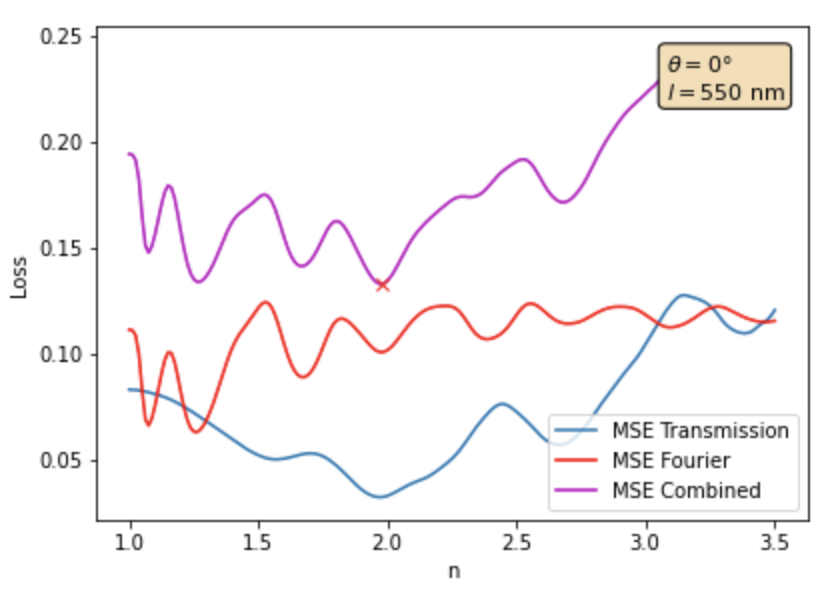}
         \caption{}
         \label{fig: n loss 82}
     \end{subfigure}
     \begin{subfigure}[b]{0.32\textwidth}
         \centering
         \includegraphics[width=\textwidth]{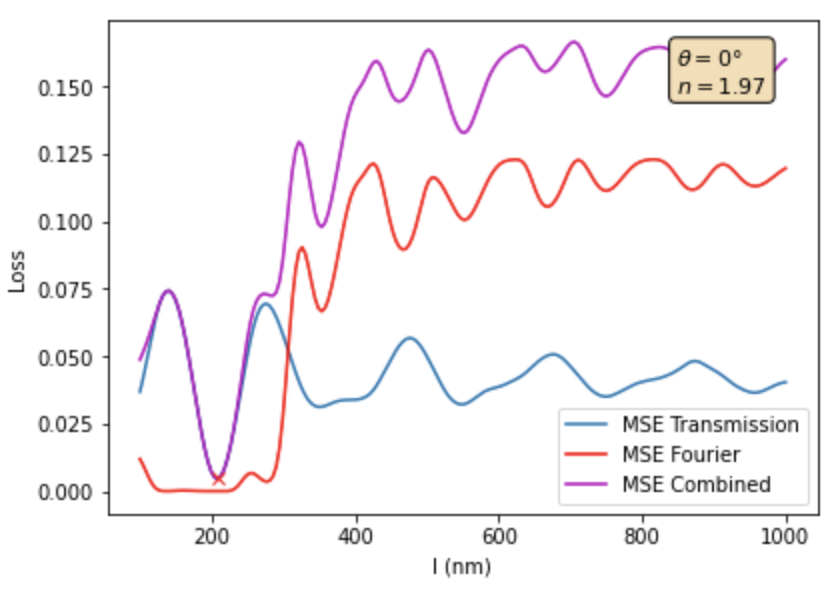}
         \caption{}
         \label{fig: l loss 82}
     \end{subfigure}
\caption{Before the gradient descent, we first minimize the loss for all three parameters.  The three images show the three consecutive minimizations we perform. a) Step 1: We minimize $\theta$ for random values of $n$ and $l$. b) Step 2: We minimize $n$ with the initial value obtained for $\theta$ and a random $l$. c) Step 3: We minimize $l$ using the initial values we obtained for $\theta$ and $n$.}
\label{fig: theta, n, l loss functions 82}
\end{figure}

\begin{figure}
    \centering
    \includegraphics[width=\textwidth]{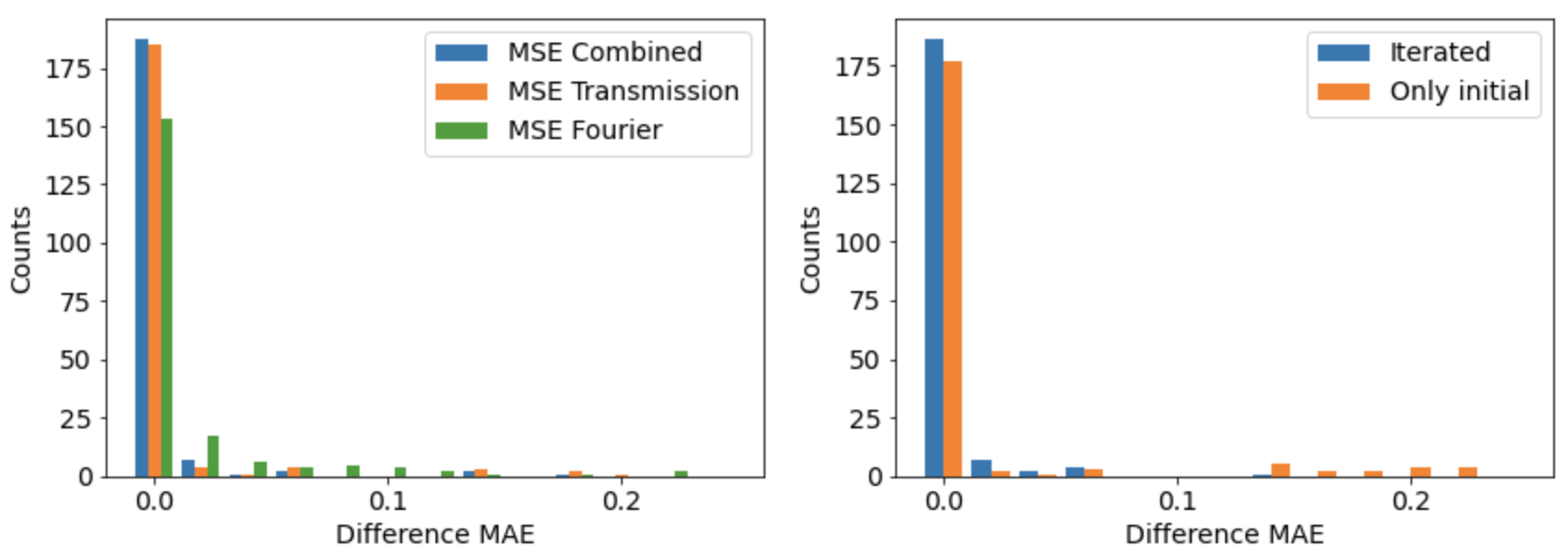}
    \caption{We performed inverse design on 200 transmissions in the test set. We look at the MAE between the desired transmission and the final predicted transmission. We then take difference between this MAE and the MAE we would obtain with the true parameters. Transmissions in the first bin have a MAE of less than $1\%$. In the first plot, we compared three loss functions to initialize the design parameters. After 100 epochs, the loss was used to re-initialize these parameters. The second plot compares our method when we re-initialize the design parameters to the case where we only initialize once. We used the combined MSE in the second plot.}
    \label{fig: inverse design MAT result}
\end{figure}

$\\$
We test this method on 200 transmissions from the test set. Again, we make a new guess every 100 epochs. Results are shown in figure \ref{fig: inverse design MAT result}. The transmissions in the first bin have to a MAE of less than 0.01. Of the three secondary losses, the combined MSE gives the best results. We also see that using the MSE on the transmission to guess the parameters works quite well. Looking at the MSE on the Fourier transform alone gives the worst results. 

$\\$
$\\$
We also check if we really need to make a guess every 100 epochs. We use the best secondary loss function, the combined MSE, to make the comparison. We compare to the case where the combined loss function is only used to determine the initial guess. Results are shown in figure \ref{fig: inverse design MAT result}. The results are not as significant as for $F$ and $\delta_0$. Making a new guess every 100 epochs does lead to a lower MAE, but only by a small margin.

\begin{figure}[h!]
     \centering
    \includegraphics[width=\textwidth]{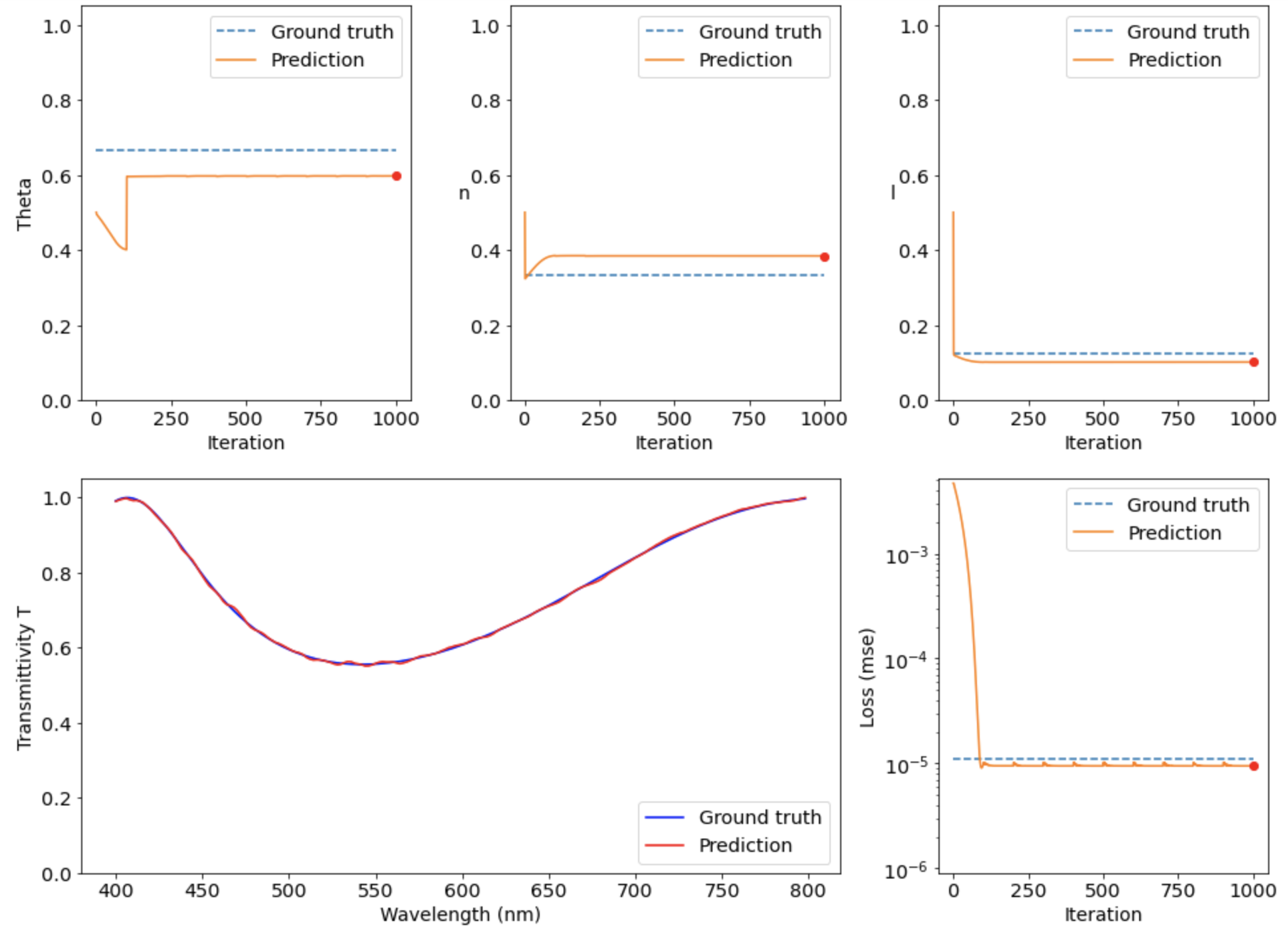}
    \caption{Inverse design for a transmission in the test set. The lower panels show the desired transmission in blue and the transmission predicted for the obtained design  parameters  in  red. The upper  panels show the evolution of the design parameters. The panel in the lower right shows the evolution of the MSE loss on the transmissions during training. We observe that $\theta$ makes a discontinuous jump when the parameter is re-initialized at 100 epochs.}
    \label{fig: inverse MAT 82}
\end{figure}

$\\$
As already discussed, the inverse design converges to a solution that is not unique. To see this, we look at what happens for a transmission in the test set shown in figure \ref{fig: inverse MAT 82}. The design parameters converge to a global minimum. The optimal parameters are however not the same as the true parameters used to generate the spectrum. This means that we have found a different design for the Fabry-Pérot resonator with the same transmission $T(\lambda)$.

\clearpage

\section{Application: the block pulse and the elephant}

\begin{figure}[h!]
     \centering
     \begin{subfigure}[b]{\textwidth}
         \centering
         \includegraphics[width=0.7\textwidth]{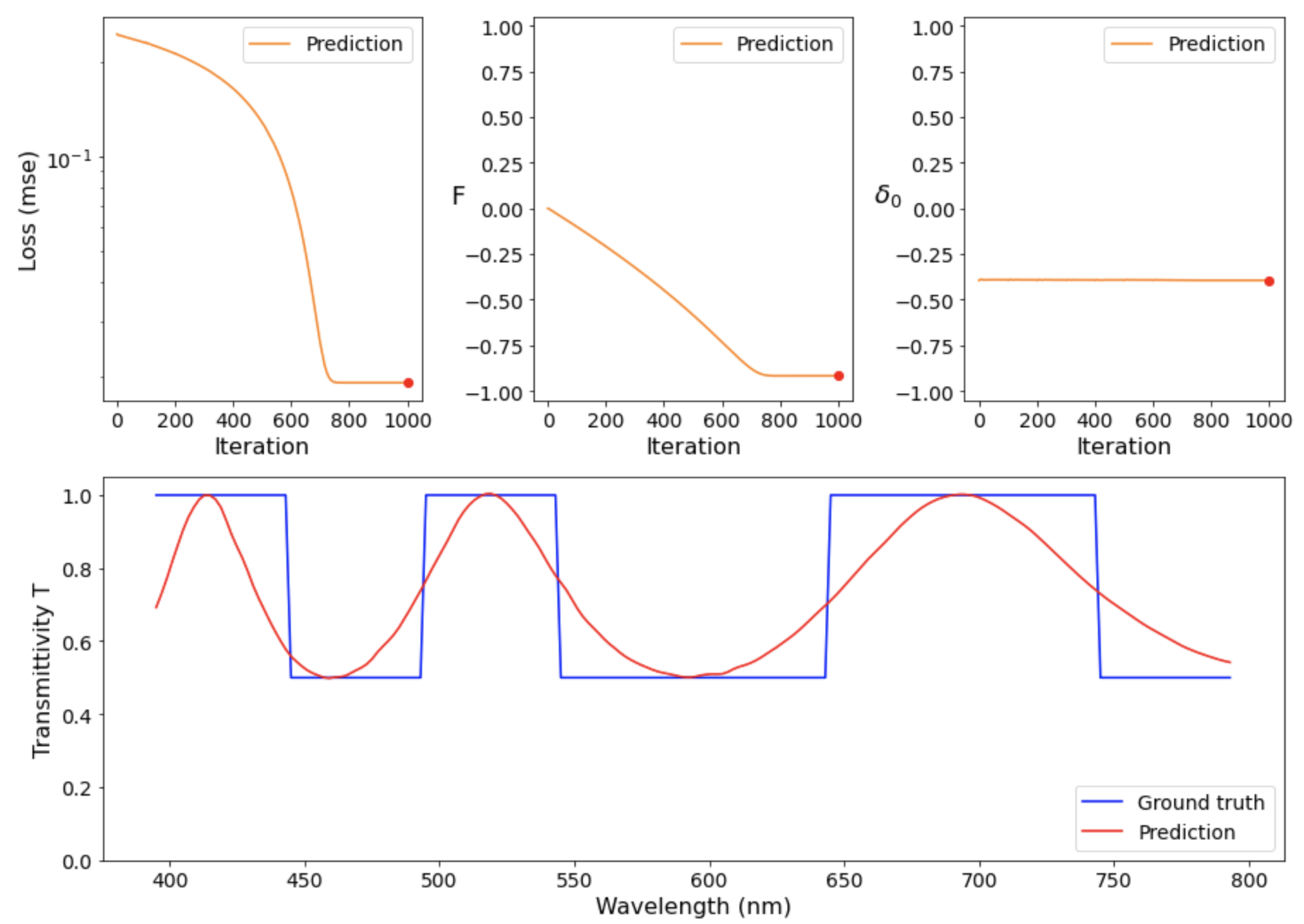}
         \caption{}
         \label{fig: block pulse FD}
     \end{subfigure}
     \hfill
     \begin{subfigure}[b]{\textwidth}
         \centering
         \includegraphics[width=0.7\textwidth]{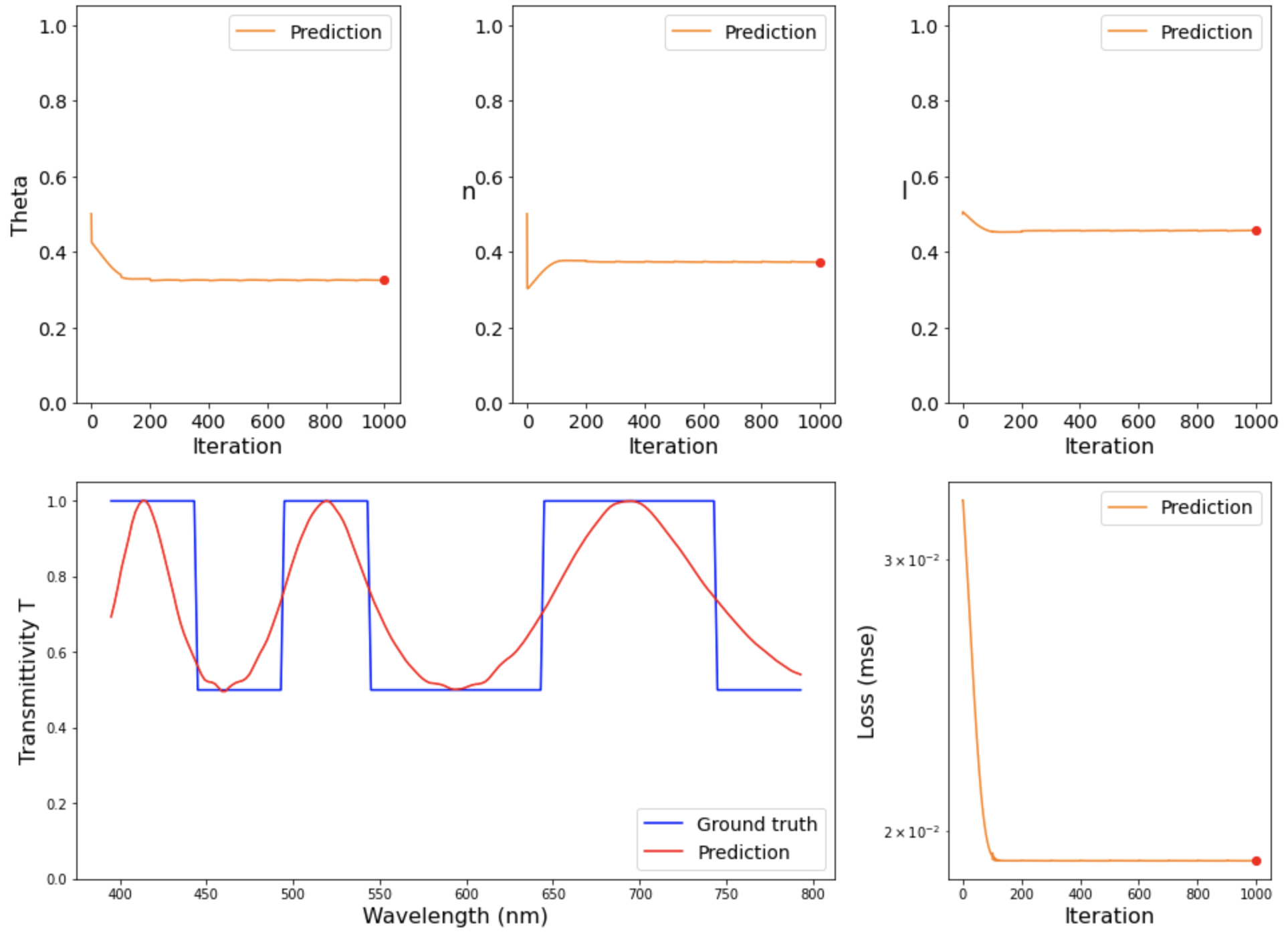}
         \caption{}
         \label{fig: block pulse MAT}
     \end{subfigure}
       \caption{Inverse design of a block pulse.  The lower panels show the desired block pulse in blue and the transmission predicted for the obtained design parameters in red. The upper panels show the  evolution of the design parameters. The MSE on the transmissions during training is also plotted. a) Inverse design on $F$ and $\delta_0$. We converge to a minimum. b) Inverse design on $\theta$, $n$ and $l$. The design parameters converge to a minimum.}
    \label{fig: block pulse}
\end{figure}

Let us now conclude with what we have promised at the very beginning, inverse design on an arbitrary transmission $T(\lambda)$. We start out with the case of a block pulse. We used the inverse design method where we made a guess every 100 epochs based on the combined MSE on the transmissions and the Fourier transform. We did this for both $F$ and $\delta_0$ and the original design parameters $\theta$, $n$ and $l$. The result is shown in figure \ref{fig: block pulse}. The results are as good as we can hope for. The transmission adapts itself to be near the block pulse, while still remaining a Fabry-Pérot transmission.

$\\$
To show that we can really approximate arbitrary transmissions, we use an example from the novel "Le petit prince" by Antoine de Saint-Exupéry. The book opens with a picture of an elephant that is being eaten by a snake. When this picture is shown to adults in the book, they see an ordinary hat. If only they knew. 

\begin{figure}[h!]
     \centering
     \begin{subfigure}[b]{\textwidth}
         \centering
         \includegraphics[width=0.75\textwidth]{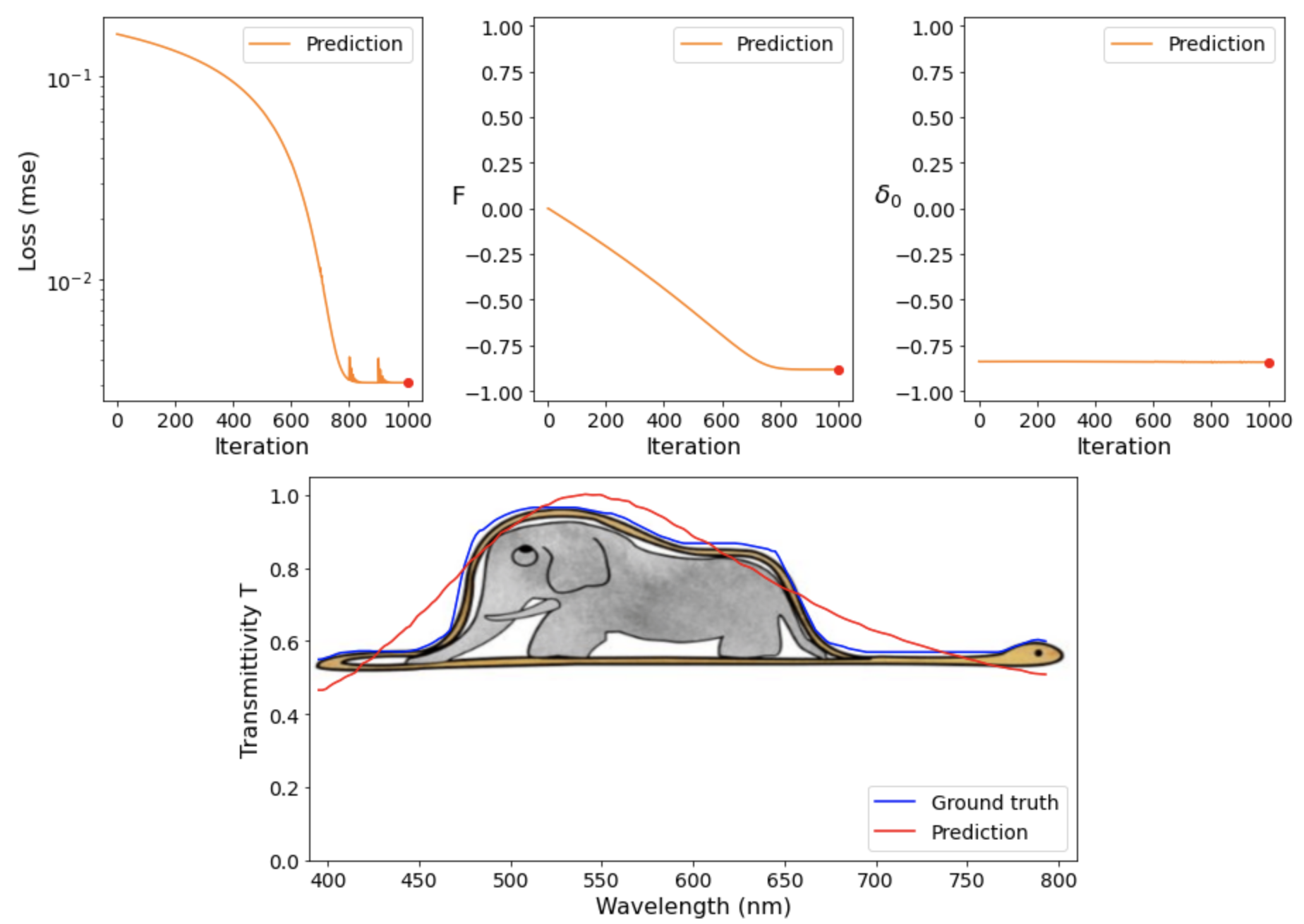}
         \caption{}
         \label{fig: elephant FD}
     \end{subfigure}
     \hfill
     \begin{subfigure}[b]{\textwidth}
         \centering
         \includegraphics[width=0.75\textwidth]{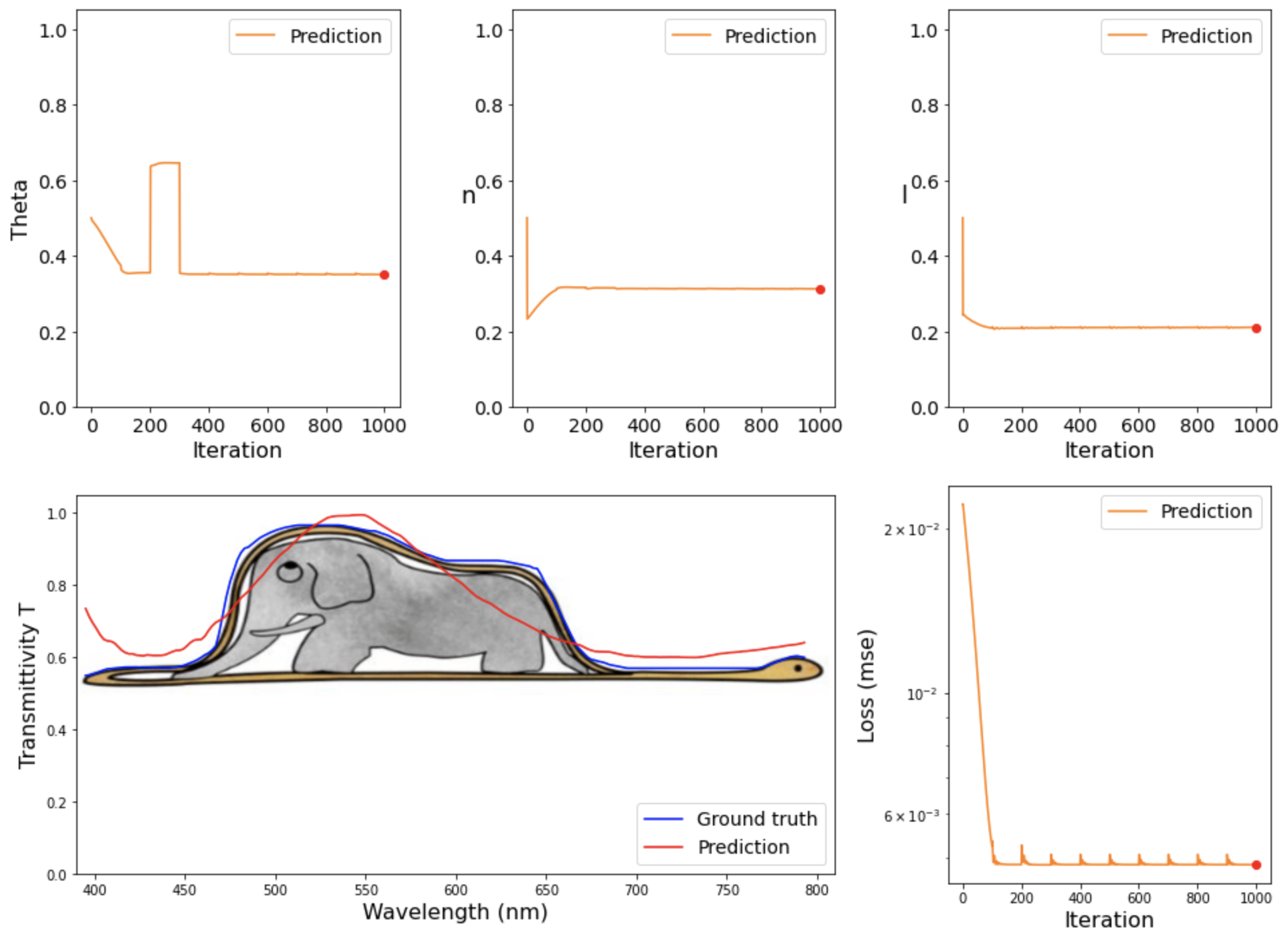}
         \caption{}
         \label{fig: elephant MAT}
     \end{subfigure}
       \caption{Inverse design of an elephant. The lower panels show the elephant with its approximation in blue and the transmission predicted for the obtained design parameters in red. The upper panels show the  evolution of the design parameters. The MSE on the transmissions during training is also plotted. a) Inverse design on $F$ and $\delta_0$. We converge to a minimum. b) Inverse design on $\theta$, $n$ and $l$. The design parameters converge to a minimum.}
    \label{fig: elephant}
\end{figure}

$\\$
We made an approximation of the elephant and performed inverse design on this 'transmission'. It was somewhat surprising to see that the algorithm was able to handle this and did converge to a solution. The result is shown in figure \ref{fig: elephant}. This shows that our inverse design method is capable of approximating really any transmission.

%% file: chapters/closingremarks.tex
\section{Main results}

We successfully trained different neural networks to learn the transmission of Fabry-Pérot. We showed that we can predict the wavelength-dependent transmission. We also succeeded in making a neural network predicting the angle-dependent transmission. This could be of use in the field of optical computing. Finally, we showed that the parameters $F$ and $\delta_0$ that appear in the analytical expression for the transmission can improve the predictions on the wave-length transmissions. The best networks we created for these three cases had a mean absolute error (MAE) of about 1\% or less. 

$\\$
Based on the observation that it is very useful to know $F$ and $\delta_0$, we wanted to recover them in a latent representation using a $\beta$-VAE. We created a $\beta$-VAE that was able to make good reconstructions of the transmissions $T(\lambda)$. Upon closer inspection, we saw that we could retrieve a transformation of $F$ in one of these dimensions. The parameter $\delta_0$ was however spread across multiple latent dimensions. The correlation between $\delta_0$ and the latent dimensions also showed a non-trivial pattern. 

$\\$
At last, we used the neural networks of chapter 4 to perform inverse design. We tried to recover the design parameters from which a transmission $T(\lambda)$ was made. We first tried gradient-descent with the MSE loss on the transmission. We got stuck in local minima because the MSE loss on the transmission is not a convex function of $\delta_0$. We then used the Fourier transform of the transmissions to get us to the global minimum. This allowed us to perform inverse design on all transmissions from the test set. We were also able to design both $F$ and $\delta_0$ and the original design parameters $\theta$, $n$ and $l$ for arbitrary transmissions $T(\lambda)$.

\section{Limitations of current research}

In chapter 4, we already experimented with different hyperparameters. There are however a lot of hyperparameters that we did not look at. We expect that bigger networks with more layers and more nodes could  perform even better than the networks we trained. Nonetheless, we did not believe it was very interesting to further improve our results. We were already quite satisfied with an MAE of about 1\%.

$\\$
The greatest shortcoming for the $\beta$-VAE was that $\delta_0$ was not stored in a single latent dimension. 

$\\$
For the inverse design, we came up with the loss function of the Fourier transform since we knew analytically that $\delta_0$ existed and was linked to the oscillation of the transmission. We might not have thought of this if we did not already know the properties of $\delta_0$.

\section{Future research}

The prediction of the transmission as well as the inverse design in this thesis were performed on Fabry-Pérot. This is a resonator with only one layer. It would be interesting to expand this to multilayer nanophotonic structures. These are more challenging and new results to design them could be of practical use. Moreover, it would be interesting to see if we can perform inverse design without knowing the underlying parameters.

$\\$
In future work on the $\beta$-VAE, we would like to understand what is happening to $\delta_0$. We see two interesting options to go forward. The first is to change the dimension of the latent representation and $\beta$. Changing these parameters has a similar effect, they both control how much information is stored in the latent representation. We could investigate over how much dimensions $\delta_0$ is spread and if this information is spread evenly among the dimensions.  

$\\$
The other option we see is to change the loss function. An MSE loss on the Fourier transform proved to be very useful for inverse design. Since we are again considering $\delta_0$ in the $\beta$-VAE, the Fourier transform might be useful again. 

$\\$
We see that this thesis only began to explore inverse design in Nanophotonics. There are still a lot of interesting directions to pursue that continue on this work. 

%% file: chapters/appendixA.tex
\begin{figure}
    \centering
    \includegraphics[width = 0.4\textwidth]{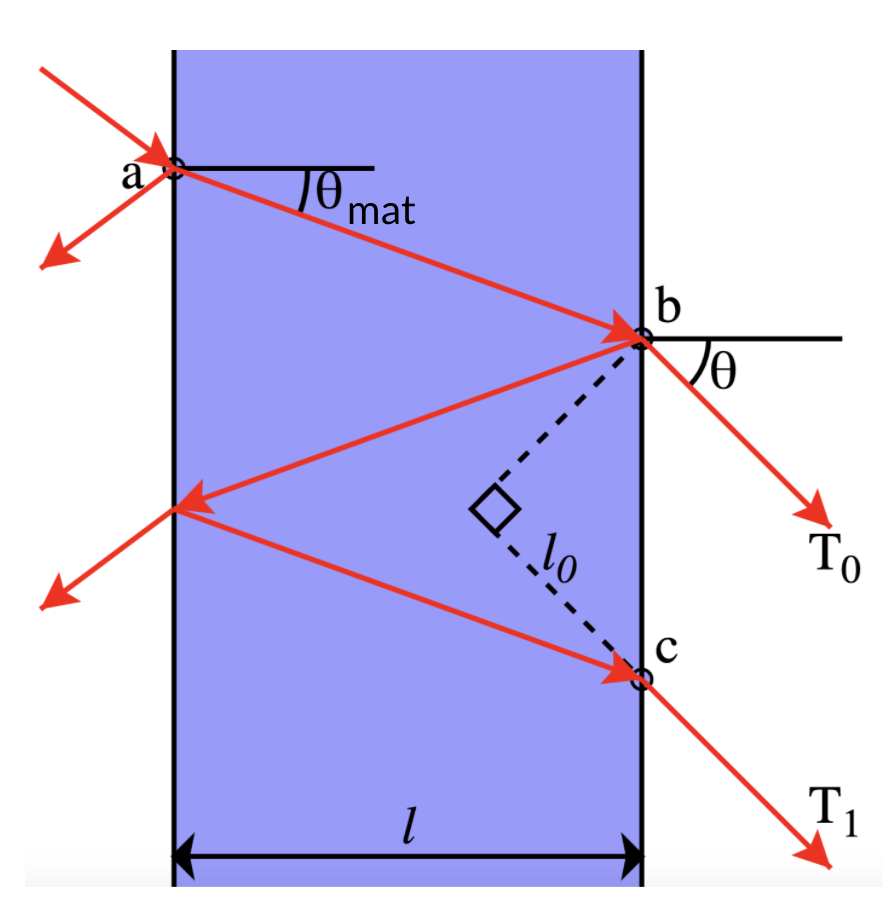}
    \caption{Fabry-Pérot resonator}
    \label{fig: appendix Fabry-Perot}
\end{figure}

The Fabry-Pérot system consists of a dielectric medium with index of refraction $n$ in which a light ray resonates several times. The different outgoing waves interfere, and this leads to a non-trivial total reflection $R$ and transmission $T$ of the system. If we assume that there is no loss in the system, conservation of energy implies $R + T$ = 1. We can calculate the resulting transmission $T$ by computing the sum of the different partial waves $T_i$. The set-up of our calculation is shown in figure \ref{fig: appendix Fabry-Perot}. The first outgoing wave is given by

\begin{equation}
\begin{aligned}
    t_{0} &= T e^{i \frac{kl}{\text{cos} \theta_{mat}}}, \\
    &= T e^{i \frac{2 \pi n}{\lambda} \frac{l}{\text{cos} \theta_{mat}}}.
\end{aligned}
\end{equation}

\noindent In this expression, $k$ is the wavenumber of the light inside the resonator. It can be written in function of the wavelength of the light and the angle of refraction inside the resonator. The angle $\theta_{mat}$ in the equation above is the angle inside the resonator. This angle is related to the angle $\theta$ of the incoming light by Snell's law

\begin{equation}
    n \text{sin} \theta_{mat} = n_0 \text{sin} \theta.
    \label{eq: Snell's law}
\end{equation}

\noindent We assume that the medium outside the resonator is air so that we have $n_0 = 1$. After a round trip in the resonator, a second outgoing wave emerges, given by

\begin{equation}
\begin{aligned}
    t_{1}' &= T R e^{i \frac{3kl}{\text{cos} \theta_{mat}}}.
\end{aligned}
\end{equation}

\noindent We can not simply add the two waves $t_0$ and $t_1'$ together, since the wave $t_0$ has traveled over a distance in the time the light made a round trip in the resonator. In order to compensate for this effect, we need to retard the second wave. We get

\begin{equation}
\begin{aligned}
    t_{1} &= T R e^{i \frac{3kl}{\text{cos} \theta_{mat}} - i k_0 l_0}.
\end{aligned}
\end{equation}

\noindent Looking at figure \ref{fig: appendix Fabry-Perot}, we see that we have 

\begin{equation}
\begin{aligned}
    k_0 = \frac{2 \pi n_0}{\lambda}, \hspace{0.5 cm} l_0 = 2l \text{tan} \theta_{mat} \ \text{sin} \theta. 
\end{aligned}
\end{equation}

\noindent In these equations, $k_0$ is the wavenumber of the light traveling in air with $n_0=1$. The length $l_0$ is the distance the first wave $t_0$ will have travelled in the time the light made a round trip in the resonator. To be able to add up the different outgoing partial waves, we calculate the phase difference between them

\begin{equation}
\begin{aligned}
    \delta &= \frac{2kl}{\text{cos} \theta_{mat}} - k_0 l_0, \\
    &= \frac{2kl}{\text{cos} \theta_{mat}} - \frac{2 \pi n_0}{\lambda} 2l \text{tan} \theta_{mat} \ \text{sin} \theta, \\
    &= \frac{2kl}{\text{cos} \theta_{mat}} - \frac{2 \pi n}{\lambda} 2l \text{tan} \theta_{mat} \ \text{sin} \theta_{mat}, \\
    &= \frac{2kl}{\text{cos} \theta_{mat}} - 2kl \frac{\text{sin}^2 \theta_{mat}}{\text{cos} \theta_{mat}}, \\
    &= \frac{2kl \text{cos}^2 \theta_{mat} }{\text{cos} \theta_{mat} }, \\
    &= 2kl \text{cos} \theta_{mat}.
\end{aligned}
\end{equation}

\noindent We used Snell's law \ref{eq: Snell's law} to go from the second to the third line. Then we used trigonometric identities to further simplify the expression. Using the phase difference between two outgoing waves, we can sum over all of them, we get

\begin{equation}
\begin{aligned}
    t &= \sum_{n=0}^{\infty} T R^n e^{i n \delta},\\
    &= \frac{T}{1 - R e^{i \delta}}.
\end{aligned}
\end{equation}

\noindent Now that we have an expression for the sum of the partial waves, we can calculate the total transmitted intensity

\begin{equation}
\begin{aligned}
    T &= t^{\ast} t = \frac{T^2}{(1 - R e^{i \delta})(1 - R e^{-i \delta})}, \\
    &= \frac{T^2}{1 - 2R cos \delta + R^2}, \\
    &= \frac{(1-R)^2}{1 - 2R + 4R sin^2 \frac{\delta}{2} + R^2}, \\
    &= \frac{1}{1 + \frac{4R}{(1-R)^2} sin^2 \frac{\delta}{2}}.
\end{aligned}
\end{equation}

\noindent We can introduce a new parameter $F$ to simplify this expression. We end up with

\begin{equation}
    T = \frac{1}{1 + F sin^2 (\frac{\delta}{2})},
    \label{eq: transmission}
\end{equation}

\noindent where the two parameters are given by

\begin{equation}
    F = \frac{4R}{(1-R)^2}, \hspace{1 cm} \delta = \frac{2 \pi n}{\lambda} 2l cos \theta_{mat}.
    \label{eq: F and delta}
\end{equation}

\noindent The parameter $F$ depends on the reflectance $R$ on the sides of the resonator. This can be calculated from the indices of refraction on both sides and the angle under which the light comes in. For TE polarized light we obtain

\begin{equation}
    R = \Bigg[\frac{n cos \theta_{mat} - cos \theta}{n cos \theta_{mat} + cos \theta} \Bigg]^2.
    \label{eq: reflectivity}
\end{equation}

\noindent We always work with TE polarized light in this work. Equations \ref{eq: transmission}, \ref{eq: F and delta} and \ref{eq: reflectivity} are what we need to analytically compute the transmission of the Fabry-Pérot resonator. 